\pdfoutput=1

\documentclass[11pt]{article}
\usepackage{CJKutf8}
\usepackage{acl}

\usepackage{times}
\usepackage{latexsym}

\usepackage[T1]{fontenc}

\usepackage[utf8]{inputenc}

\usepackage{microtype}

\usepackage{inconsolata}

\usepackage{times}
\usepackage{latexsym}
\usepackage{balance}
\usepackage[T1]{fontenc}

\usepackage[utf8]{inputenc}

\usepackage{microtype}

\usepackage{inconsolata}
\usepackage{CJKutf8}
\usepackage{booktabs}
\usepackage{graphicx}
\usepackage{array}
\usepackage{bm}
\usepackage{amsmath}
\usepackage{enumitem}
\usepackage{threeparttable}
\usepackage{multirow}
\usepackage{makecell}

\usepackage{latexsym} 

\usepackage[T1]{fontenc}

\usepackage[utf8]{inputenc}
\usepackage{float}

\usepackage{microtype}

\usepackage{xcolor}

\def\BibTeX{{\rm B\kern-.05em{\sc i\kern-.025em b}\kern-.08em
    T\kern-.1667em\lower.7ex\hbox{E}\kern-.125emX}}
\usepackage{cite}
\usepackage{amsmath,amssymb,amsfonts}
\usepackage{algorithmic}
\usepackage[ruled,linesnumbered,noline,noend]{algorithm2e}

\SetCommentSty{mycommfont}

\usepackage[medium,compact]{titlesec}

\SetNlSty{}{}{}

\let\oldnl\nl
\newcommand\nonl{%
  \renewcommand{\nl}{\let\nl\oldnl}}
  
\usepackage{hyperref}
\hypersetup{
    colorlinks=true,
    linkcolor=blue,}
\AtBeginDocument{%
  \providecommand\BibTeX{{%
    \normalfont B\kern-0.5em{\scshape i\kern-0.25em b}\kern-0.8em\TeX}}}

\usepackage{textcomp}
\usepackage{xcolor}
\def\BibTeX{{\rm B\kern-.05em{\sc i\kern-.025em b}\kern-.08em
    T\kern-.1667em\lower.7ex\hbox{E}\kern-.125emX}}
\usepackage{amsmath,amssymb,amsfonts}
\usepackage{algorithmic}
\usepackage[ruled,linesnumbered,noline,noend]{algorithm2e}
\usepackage{threeparttable}
\usepackage{booktabs}
\usepackage{graphicx}
\usepackage{textcomp}
\usepackage{array}
\usepackage{bm}
\usepackage{amsmath}
\usepackage{color, colortbl}
\SetCommentSty{mycommfont}

\usepackage[misc]{ifsym}
\usepackage{amsthm}
\newtheorem{Problem definition}{Problem definition}

\def\BibTeX{{\rm B\kern-.05em{\sc i\kern-.025em b}\kern-.08em
    T\kern-.1667em\lower.7ex\hbox{E}\kern-.125emX}}
\usepackage{hyperref}
\hypersetup{
    colorlinks=true,
    linkcolor=blue,}

\usepackage{color, colortbl}
\usepackage{array}
\usepackage{bm}
\usepackage{amsmath}
\usepackage{threeparttable}
\usepackage{booktabs}
\usepackage{graphicx}
\usepackage{hyperref}
\usepackage{multirow}
\usepackage[ruled,linesnumbered,noline,noend]{algorithm2e}
\usepackage{makecell}
\hypersetup{
    colorlinks=true,
    linkcolor=blue,}
\usepackage{CJKutf8}
\usepackage{booktabs}
\usepackage{graphicx}
\usepackage{textcomp}
\usepackage{array}
\usepackage{bm}
\usepackage{amsmath}

\usepackage{threeparttable}
\usepackage{multirow}
\usepackage{makecell}
\usepackage{times}
\usepackage{latexsym}
\usepackage[T1]{fontenc}

\usepackage[utf8]{inputenc}
\usepackage{float}

\usepackage{microtype}

\usepackage{xcolor}

\def\BibTeX{{\rm B\kern-.05em{\sc i\kern-.025em b}\kern-.08em
    T\kern-.1667em\lower.7ex\hbox{E}\kern-.125emX}}
\usepackage{cite}
\usepackage{amsmath,amssymb,amsfonts}
\usepackage{algorithmic}
\usepackage[ruled,linesnumbered,noline,noend]{algorithm2e}

\SetCommentSty{mycommfont}

\usepackage{fontawesome}

\SetNlSty{}{}{}

\let\oldnl\nl

\usepackage{hyperref}
\hypersetup{
    colorlinks=true,
    linkcolor=blue,}
\AtBeginDocument{%
  \providecommand\BibTeX{{%
    \normalfont B\kern-0.5em{\scshape i\kern-0.25em b}\kern-0.8em\TeX}}}
\usepackage[medium,compact]{titlesec}

\usepackage{times}
\usepackage{latexsym}

\usepackage[T1]{fontenc}

\usepackage[utf8]{inputenc}

\usepackage{microtype}
\usepackage{mathrsfs} 

\title{\textsc{HotVCom}: Generating Buzzworthy Comments for Videos}

\author{Yuyan Chen$^{1}$\thanks{\quad Work done during an internship at Ant Group.}, Yiwen Qian$^{2}$, Songzhou Yan$^{1}$, Jiyuan Jia$^{4}$, Zhixu Li$^{1}$, \\ \textbf{Yanghua Xiao}$^{1}$, \textbf{Xiaobo Li}$^{3}$, \textbf{Ming Yang}$^{3}$, \textbf{Qingpei Guo}$^{3}$ $^{\textrm{\Letter}}$ \thanks{\quad Qingpei Guo is the corresponding author.}\\
        $^1$Shanghai Key Laboratory of Data Science, School of Computer Science, Fudan University, \\
        $^2$Arizona State University,
        $^3$Ant Group,\\
        $^4$Southern University of Science and Technology\\
        \texttt{\{chenyuyan21@m., szyan21@m., zhixuli@, shawyh@\}fudan.edu.cn},\\
        \texttt{qianyiwenlyy@gmail.com},
        \texttt{\{xiaobo.lixb, m.yang, qingpei.gqp\}@antgroup.com},\\
        \texttt{jiajy2018@mail.sustech.edu.cn}\\
        }

\begin{document}
\maketitle
\begin{abstract}
In the era of social media video platforms, popular ``hot-comments'' play a crucial role in attracting user impressions of short-form videos, making them vital for marketing and branding purpose. However, existing research predominantly focuses on generating descriptive comments or ``danmaku'' in English, offering immediate reactions to specific video moments. Addressing this gap, our study introduces \textsc{HotVCom}, the largest Chinese video hot-comment dataset, comprising 94k diverse videos and 137 million comments. We also present the \texttt{ComHeat} framework, which synergistically integrates visual, auditory, and textual data to generate influential hot-comments on the Chinese video dataset. Empirical evaluations highlight the effectiveness of our framework, demonstrating its excellence on both the newly constructed and existing datasets.
\end{abstract}

\section{Introduction}
\begin{figure}[!t]
  \centering
  \includegraphics[width=0.9\linewidth]{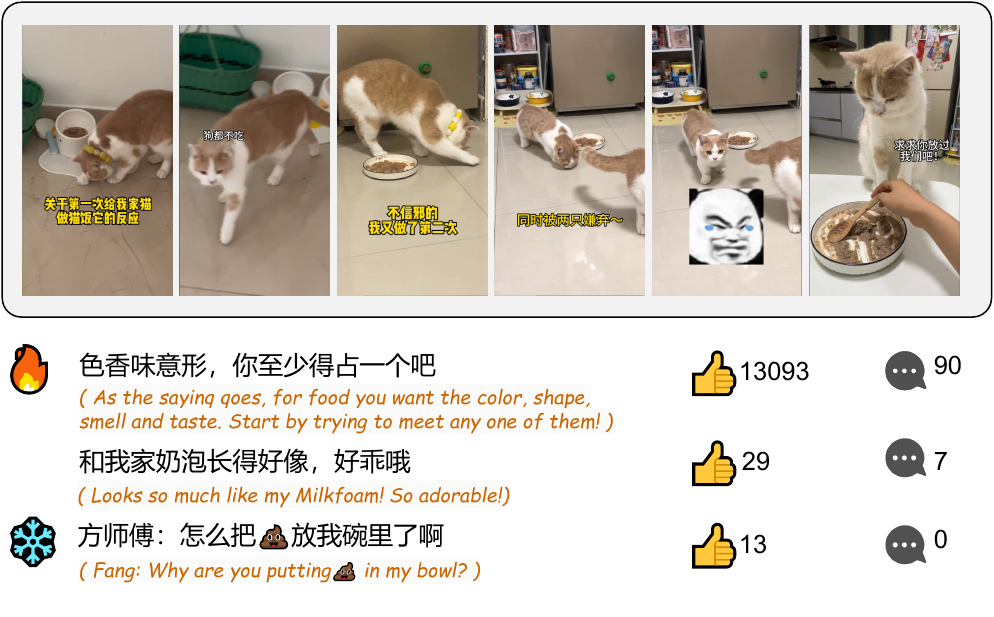}
  \caption{\small A hot comment attracts many likes and replies compared with a cold comment for a short video.}
  \label{fig:v-intro}
\end{figure}

With the increasing prevalence of video content on digital platforms, there is an evident significance of video comments in amplifying video reach~\citep{ren2024survey, chen2024xmecap}. Specifically, ``hot-comments'' have the potential to attract considerable user interaction, substantially increasing a video's user impressions, which is essential for product marketing and branding. A typical hot-comment often meets specific standards: receiving a larger number of likes and replies, being highly pertinent to the video content, and including the elements that resonate with viewers, as illustrated in Fig.~\ref{fig:v-intro}.

However, the prevailing literature, such as the work by ~\citet{ma2019livebot} and ~\citet{wang2020videoic}, largely concentrates on generating descriptive comments or ``danmaku'', which are closely tied to certain video moments, offering instantaneous reactions. While these comments do engage viewers to some extent, their potential to highlight the entire short-video content or foster deep user interactions is somewhat limited. When aiming to elevate a video's visibility, these real-time reactions might not be as influential as those hot-comments. Additionally, most research, including that by ~\citet{sun2023vico}, predominantly focuses on English comments, leaving a gap in the domain of Chinese hot-comment generation. Recently, large language models (LLMs) have played a significant role across various fields, making it feasible to use them for generating hot comments~\citep{xiong2024largelanguagemodelslearn,chen2023can,chen2024dolarge,chen2023mapo,chen2023hadamard,chen2024drAcademy}. Due to the lack of large-scale training datasets, we first construct a comprehensive Chinese video hot-comment generation dataset, \textsc{HotVCom}, which includes video titles, descriptions, captions, audio speeches, keyframes, and engagement records.
Compared to the English dataset from~\citet{sun2023vico}, our dataset \textsc{HotVCom} is more comprehensive, encompassing 93k videos with 137 million comments, making it the largest of its kind.
%
In addition, for the convenience of analysis and targeted generation of comments, videos in \textsc{HotVCom} are further categorized into different themes. Such a categorization helps language models to understand the context, therefore generating hot-comments that resonate with specific themes accordingly.

A key challenge in video hot-comment generation lies on assessing whether the generated comments are truly impressive to users to boost the interaction.
The existing work primarily utilizes ROUGE~\citep{ma2019livebot,wang2020videoic,chen2024temporalmed} or the number of likes~\citep{sun2023vico} as metrics. However, these metrics might not always reflect the genuine engagement of a comment~\citep{chen2023hallucination,chen2024emotionqueen}.
In this paper, we propose a novel comprehensive evaluation metric including the informativeness, relevance, creativity besides user engagement (i.e. likes and replies) of a comment, which thus reveals a well-rounded understanding of a comment's `hotness'.
With the guidance of our evaluation metrics, we then propose a novel video hot-comment generation framework named \texttt{ComHeat}. We implement the Supervised Fine-Tuning technique to generate preliminary comments and then enhance them with reinforcement learning. By incorporating knowledge-enhanced Tree-of-Thought method, the comments are further refined to improve the chance of their popularity.

In summary, our contributions are:
\begin{itemize}
    \item We work on a novel task namely video hot-comment generation. To achieve this, we construct the largest Chinese video comment dataset including 93k videos with 137 million comments, named \textsc{HotVCom}.
    
    \item 
    We introduce a novel comprehensive evaluation metric for video hot-comments generation, including informativeness, relevance, creativity, and user engagement, bridging the gap in qualitative comment analysis.

    \item We propose the \texttt{ComHeat} framework, which incorporates visual, auditory, and textual aspects, for generating engaging comments for Chinese short videos using reinforcement learning and Tree-of-Thought.

    \item Empirical results showcase that our \texttt{ComHeat} framework outperforms existing baselines on the newly-constructed dataset and also excels on other video comment generation and captioning tasks.
\end{itemize}

\section{Datasets}
\begin{figure*}[!t]
  \centering
  \includegraphics[width=0.8\linewidth]{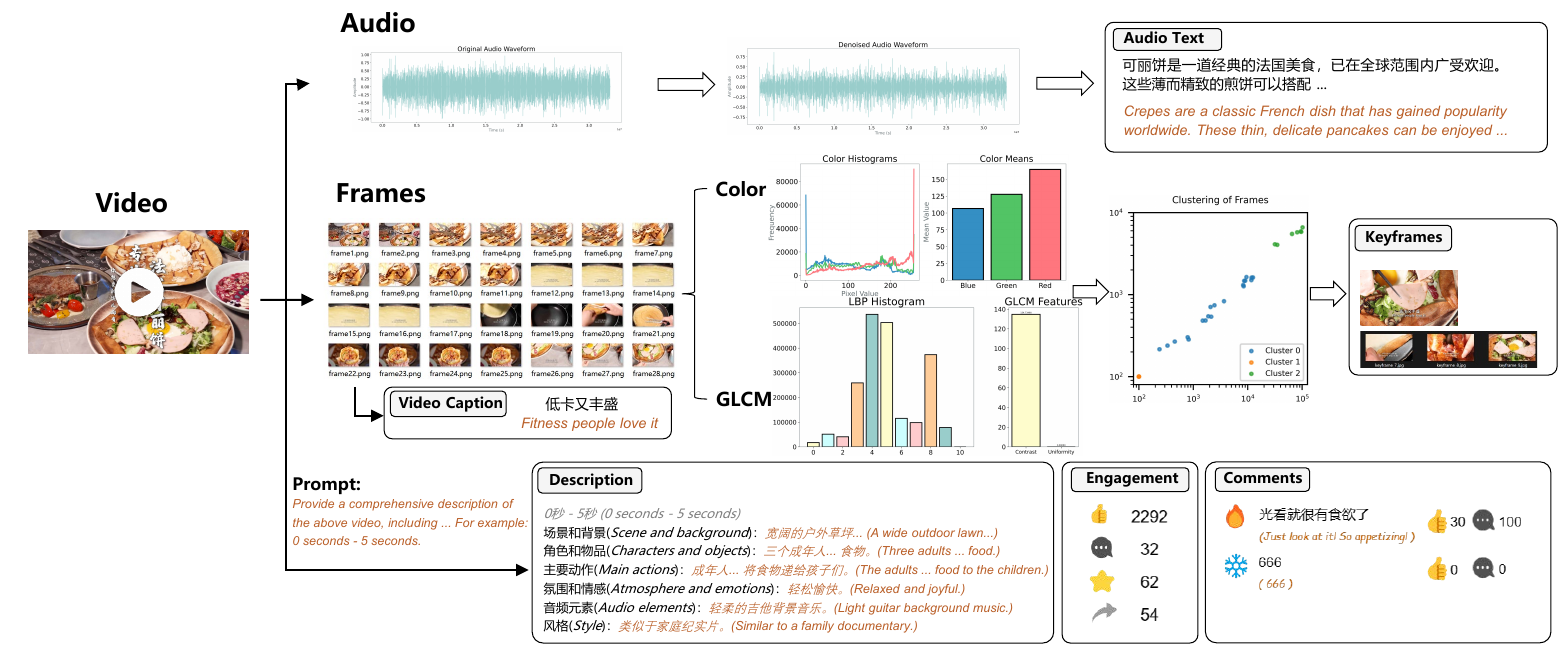}
  \caption{\small The process of constructing Chinese short video comment dataset \textsc{HotVCom}.}
  \label{fig:v-data-construct}
\end{figure*}

In this section, we construct a large-scale Chinese short video comment dataset named \textsc{HotVCom} including 94k short videos from Douyin with 137 million comments, where the entire process is shown in Fig.~\ref{fig:v-data-construct}. We also conduct an extensive exploratory data analysis as shown in Fig.~\ref{fig:v-eda}.

We initially collect Douyin videos from various themes in reverse chronological order up till 100k. Next, we conduct Optical Character Recognition (OCR) with PaddleOCR library~\footnote{https://github.com/PaddlePaddle/PaddleOCR} and Automatic Speech Recognition (ASR) with Xunfei open platform~\footnote{https://raasr.xfyun.cn/v2/api} on the videos to obtain their video captions and audio speech, respectively, and capture key frames of videos with the K-means clustering algorithm. After that, we also adopt PaddleOCR to extract the video's title, creation time, publishers' profile and engagement information, which includes the number of likes, comments, shares, and favorites, as well as each comment with its content, commenter's profile and engagement information, which includes the number of likes and replies. Furthermore, from the tags marked with ``\#'' in titles and themes provided by the Douyin platform, we categorize the videos with GPT-4~\footnote{https://chat.openai.com/} into 20 themes, including pets, food, etc. We also provide the descriptions for the video content and keyframes, encompassing scenes, objects, primary actions, atmosphere, and emotions with the help of video-ChatGPT~\citep{li2023videochat} and miniGPT4~\citep{zhu2023minigpt}, respectively. More details are shown in Appendix~\ref{sec:data}.

To maintain comment quality, we filter out comments with emojis, ASCII characters below 127, and those with less than 1 character or more than 50 characters. We also remove the comments with profanity, political content, negative tones, or promotional intent. The short videos with 0 or 1 comment are also discarded to avoid long-tail bias. In the end, we have 94k videos with 137 million comments, including video captions, audio texts, keyframes, engagement information, etc.
%
The statistics of \textsc{HotVCom} are shown in Table~\ref{tab:data}. On average, these videos are around 96.44 seconds long, indicating a preference for 1-2 minute content. The titles and descriptions average 43.62 and 419.22 characters, offering viewers a good context. Engagement information, such as the average number of likes of 25230.45 but a median of 3986, highlight the variability in video popularity. Similarly, while the mean number of comments is 1446.71, a median of 195 comments suggests the user engagement is concentrated on specific viral content. This dataset showcases the diverse content and engagement levels on the most popular short video platform in China. The more exploratory analysis on the constructed \textsc{HotVCom} is illustrated in Fig.~\ref{fig:v-eda} in Appendix~\ref{sec:eda}.

\begin{table}[]
\resizebox{0.45\textwidth}{!}{%
\begin{tabular}{llllll}
\toprule
Feature                          & Mean     & Median & Max     & Min & Std  \\ \midrule
\multicolumn{6}{c}{Video}     
\\
\midrule
Video Lengths                    & 96.44    & 60     & 584     & 5   & 1e4  \\
Video Keyframe Counts            & 65.65    & 35     & 300     & 1   & 1e4  \\
Video Title Lengths              & 43.62    & 41     & 1000    & 2   & 7e2  \\
Video Description Lengths        & 419.22   & 221    & 1654    & 37  & 2e5  \\
Video Caption Lengths                & 589.37   & 346    & 38032   & 2   & 8e5  \\
Audio Speech Lengths                & 395.76   & 224    & 12688   & 2   & 2e5  \\
\midrule
\multicolumn{6}{c}{Engagement} \\
\midrule
Video Likes Counts               & 25230.45 & 3986   & 1684000 & 12  & 1e10 \\
Video Comments Counts            & 1446.71  & 195    & 103000  & 6   & 4e7  \\
Video Favorites Counts           & 2883.61  & 395    & 127000  & 0   & 1e8  \\
Video Shares Counts              & 5243.27  & 255    & 717000  & 0   & 2e9  \\
Comment Likes Counts/per Video   & 210.5    & 1      & 1172929 & 0   & 2e7  \\
Comment Replies Counts/per Video & 10.67    & 0      & 67556   & 0   & 6e4  \\
Comment Lengths/per Video        & 16.47    & 12     & 500     & 3   & 3e2  \\ \bottomrule
\end{tabular}
}
\caption{\small The statistics of the constructed Chinese short video hot-comments dataset \textsc{HotVCom}.}
\label{tab:data}
\end{table}







\section{Evaluation}
We develop comprehensive evaluation metrics for hot-comments from four main aspects: informativeness, relevance, creativity, and user engagement. 

The informativeness score $I$ quantifies the utility of the information conveyed by a comment from lengths penalty and vocabulary diversity. Lengths penalty, denoted as $L_p$, quantify the lengths appropriateness of a comment. Vocabulary diversity, denoted as $V_d$, is calculated as the ratio of total bigrams to unique ones of this comment. The calculation process of informativeness score is as follows:
\begin{gather}
L_p = 
\begin{cases} 
\frac{L}{L_{\text{min}}} & \text{if } L < L_{\text{min}} \\
\frac{L}{L_{\text{max}}} & \text{if } L_{\text{min}} \leq L \leq L_{\text{max}} \\
1 - \alpha \times (L - L_{\text{max}}) & \text{if } L > L_{\text{max}} ,
\end{cases} \\
V_d = \frac{T_n}{U_n}, \quad I = w_1^I \times L_p + w_2^I \times V_d,
\end{gather}
where $L$ denotes the actual length of a given comment, $L_{\text{min}}$ and $L_{\text{max}}$ specify the optimal length boundaries of the comment length, which are set at 1 and 50, respectively, meaning that the comment length ranges from a minimum of 1 to a maximum of 50 characters. The constant $\alpha$, between 0 and 1, adjusts penalties for a comment that go beyond the optimal length. $T_n$ and $U_n$ signifying total and unique bigrams of a comment, respectively. $w_1^I$ and $w_2^I$ are trainable weights.

The relevance score $R$ quantifies the alignment of a comment to the video content through two primary dimensions: keyword and context matching degree. The keyword matching degree, denoted by $D_k$, is the proportion of words in a comment resonating with the keywords of video captions that are extracted with ChatGPT. The context matching degree, denoted by $D_c$, is derived from cosine similarity between video captions and a corresponding comment. The calculation process of relevance score is as follows:
\begin{gather}
D_k = \frac{N_x}{N_k}, \quad D_c = \frac{\mathbf{Com} \cdot \mathbf{Vid}}{||\mathbf{Com}||_2 \times ||\mathbf{Vid}||_2},\\
R = w_1^R \times D_k + w_2^R \times D_c,
\end{gather}
\noindent where $N_x$ denotes the number of words or phrases in a comment that matches the keywords in video captions, and $N_k$ is the total number of keywords extracted from the video captions with ChatGPT.
\( \mathbf{Com} \) and \( \mathbf{Vid} \) are the representation of a comment and the corresponding video, respectively.
\( w_1^R \) and \( w_2^R \) are trainable weights.
For the robustness of the evaluation by ChatGPT, we repeat the process five times to obtain the most common result. We find that the consistency rate across the five iterations reaches over 80\%. Additionally, we randomly sample 1000 cases and confirm that the results indeed align with our definition of keywords.

The creativity score $C$ offers a quantitative assessment of a comment's distinctiveness and novelty, which encompasses two primary metrics: the rhetorical technique score, denoted as $S_r$, such as metaphors and irony, and the trending term score, denoted as $S_t$. 
The calculation process of the creativity score as follows:
\begin{gather}
S_r = \frac{1}{1 + e^{-k_r(x_r - b_r)}}, \quad  S_t= \frac{1}{1 + e^{-k_t(x_t - b_t)}},\\
C= w_1^C \times S_r + w_2^C \times S_t,
\end{gather}
where $x_r$ and $x_t$ represent the occurrence of rhetorical techniques and trending terms in a comment, respectively, which are both counted by ChatGPT. \( w_1^C \) and \( w_2^C \) are also trainable weights.
Specifically, the sigmoid function is employed to modulate $S_r$ and $S_t$ within the [0,1] interval. Experimentally, the optimal values for $k_r$ and $k_t$ are 1, as well as $b_r$ and $b_t$ are -1, to ensure a score close approximate 0.1 when the count is zero, therefore achieving optimal score differentiation.

The user engagement score $U$ represents the level of users' interaction of a comment, which mainly includes likes and replies. The calculation process of user engagement score is as follows:
\begin{gather}
U'=w_1^U \times N_l + w_2^U \times N_r, \\
U = \frac{1}{1 + e^{-k_u(U' - b_u)}},
\end{gather}
where $N_l$ and $N_r$ represent the number of likes and replies in a comment, respectively. \( w_1^U \) and \( w_2^U \) are also trainable weights.
Similarly, we also adopt the sigmoid function to modulate $U'$ within the [0,1] interval. 

Finally, the comprehensive score $F$ of a comment is defined as:
\begin{gather}
F = w^I \times I + w^R \times R + w^C \times C + w^U \times U,
\end{gather}
where \(w^I\), \(w^R\), \(w^C\), and \(w^U\) are trainable weights assigned to the informativeness score, relevance score, creativity score, and user engagement score, respectively.

Furthermore, we manually score each metric (informativeness, relevance, creativity, and user engagement) with a binary 0 or 1. We set the threshold at 0.5 for each metric and use the AUC (Area Under Curve) to calculate the consistency between manual scoring and automatic scoring for each metric. We find that AUC of informativeness equals 0.83, AUC of relevance equals 0.86, AUC of creativeness equals 0.87, and AUC of user engagement equals 0.80. It suggests that the alignment of each metric with human judgment.

\section{Methods}
\begin{figure*}[!t]
  \centering
  \includegraphics[width=0.75\linewidth]{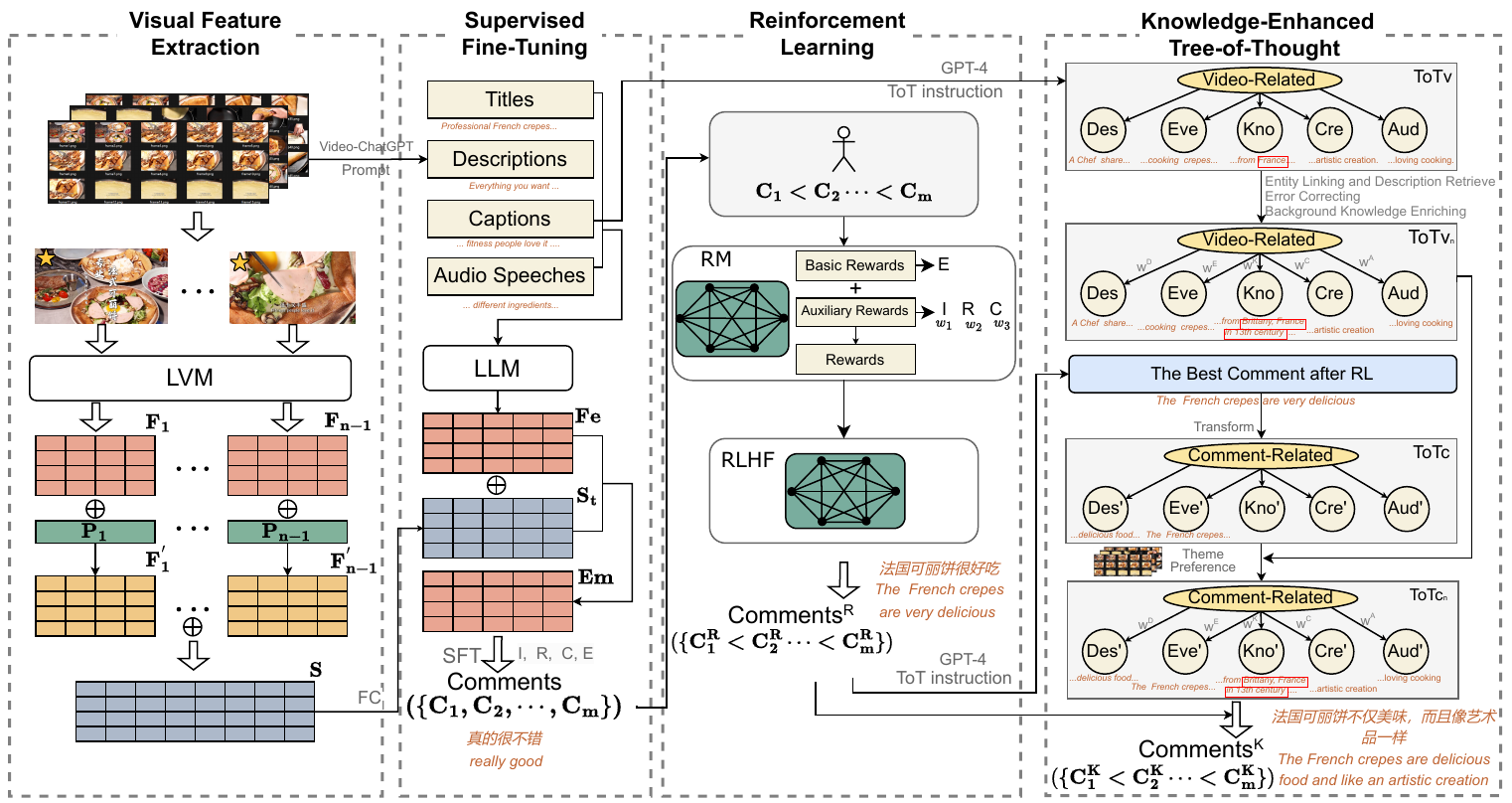}
  \caption{\small The overview of the proposed Chinese video hot-comments generation framework \texttt{ComHeat}.}
  \label{fig:v-framework}
\end{figure*}

We propose a Chinese video hot-comments generation framework named \texttt{ComHeat} as shown in Fig.~\ref{fig:v-framework}. Initial comments are first generated by the LLMs through supervised fine-tuning. Then we train a reward model based on the comprehensive score of comments and adopt reinforcement learning to refine the popularity of the generated comments. Finally, we utilize knowledge-enhanced Tree-of-Thought method for further optimization to generate hotter comments. 

\subsection{Visual Feature Extraction}
The aim of this step is to bridge the semantic gap between videos and comments. We first leverage embeddings from key video frames inspired by the previous work~\citep{alayrac2022flamingo,tao2024nevlp}. Next, recognizing the significance of event progression in videos, we preserve the keyframe sequence via positional embeddings. This ensures the sequence's essence and progression are encapsulated for subsequent processing. Mathematically:
\begin{gather}
F_k = LVM(k), \quad S = \sum_{k \in K} F_k \oplus P_k,
\end{gather}
Where \( F_k \) represents features from the k-th keyframe, extracted by LVM, \( P_k \) represents the positional embedding for the k-th keyframe, \( S \) represents the serialized sequence of keyframe embeddings, and \( K \) represents a video's keyframe set.

\subsection{Supervised Fine-Tuning}
We leverage an LLM, such as baichuan2-13B~\citep{yang2023baichuan}, for Supervised Fine-Tuning (SFT) with the aim of generating hot-comments that resonate with the ground truths. The input encompasses both textual information (titles, video captions, audio speeches, video descriptions) and visual features extracted in the last step.
Firstly, textual information, denoted by $T$, is encoded to a dense vector $T_e$. Concurrently, the visual features, represented by $S$, are transformed via a fully connected layer, ensuring compatibility with text dimensions. These modalities are then weighted fused linearly as follows:
\begin{gather}
T_e = \text{Enc}(T), S_t = \text{FC}(S), E_m = \alpha \cdot T_e + \beta \cdot S_t,
\end{gather}
where \( \alpha \) and \( \beta \) are adjustable weights.
This integrated feature, \( E_m \), is decoded to generate hot-comments of videos.
Specifically, during the training phase, the loss function, denoted as $L_{\text{SFT}}$, integrates two components, including the inherent cross entropy of the SFT process, denoted as \( L_{\text{CE}} \), and the Mean Squared Error (MSE) that quantifies the difference between the comprehensive scores of a predicted hot-comment and the ground truth, denoted as \( L_{\text{F}} \), as follows:
\begin{gather}
L_{\text{SFT}} = w_1^S \cdot L_{\text{CE}} + w_2^S \cdot L_{\text{F}},
\end{gather}
where \( w_1^S \) and \( w_2^S \) are trainable weights.

\subsection{Reinforcement Learning}
The primary goal of this step is to align the generated comments with human preferences based on Reinforcement Learning (RL) using a reward model.
The reward model associates the predicted hot-comment $x_i$ with the ground truth comment $y_i$ to compute a reward $r = R(x_i, y_i)$. 

Initially, we utilize comprehensive scores to assess 1\% of the comments generated in the SFT process. This yields a a ranked sequence for an LLM, represented as $\{c_1, c_2, \ldots, c_{n-1}, c_m\}$.
Subsequently, using this sequence, we train a reward model. We adopt a powerful LLM, named baichuan2, and replace its softmax layer with a linear one. The reward model takes a comment as the input and returns a score that indicates the comment's quality. For training, we collate responses from the ranking sequence and apply the Pairwise Ranking Loss, depicted as follows:
\begin{gather}
L_r=-\frac{1}{\binom{k}{2}}E_{(x,y_w,y_l)\sim D}
[log(\sigma (r_\theta (x,y_w)-r_\theta (x,y_l)))],
\end{gather}
where $x$ denotes the original comment, $y_w$ and $y_l$ represent the comments with higher and lower scores in the given ranking pair, respectively. $r_\theta$ is the scalar output from the reward model, $D$ is the set of ranking pairs, and $K$ is the number of comments generated during SFT.
Through this process, the reward model assigns higher scores (rewards) to high-quality comments and lower scores (penalties) to inferior comments, effectively imitating human's preferences.

Specifically, rewards comprise basic rewards, determined by user engagement scores, and auxiliary rewards determined by informativeness, relevance, and creativity scores. The basic rewards are sourced from evaluations by human annotators. 
Correlation coefficients are then computed between user engagement and one of the scores in informativeness, relevance, and creativity for each video. After that, the reward model adopts the basic and auxiliary rewards as the final rewards of a comment as follows:
\begin{gather}
w_i = \frac{ \text{corr}(x_i,u_i) }{ \sum_{j=1}^{n} \text{corr}(x_j, u_j) },\quad
r = \text{BR} + \sum_{i=1}^{n} w_i \cdot \text{AR}_i,
\end{gather}
where \( w_i \) represents the weight for one of the scores in informativeness, relevance, and creativity for a comment. 
\text{BR} and \text{AR} represent the basic rewards and the auxiliary rewards, respectively.

After that, we feed comment $x$ generated by the SFT model into the RL model $\pi_{\phi}^{RL}$ to obtain a more human-preferred comment $y$.
We first input $(x, y)$ into the reward model $r_{\theta}$ and calculate a score (i.e., reward), which represents the real-time feedback from the reward model. Next, we aims to maintain similarity between the RL model and the SFT model with Kullback-Leibler (KL) divergence. Finally, we combine the two loss functions as follows:
\begin{gather}
L_{\phi}^{r_{\theta}} = E(x, y) \sim D_{\pi_{\phi}^{RL}}[r_{\theta}(x, y)],\\
L_\phi^{\text{SFT}}=- log(\pi_{\phi}^{RL}(y | x)/\pi^{SFT}(y | x)),\\
L_{\text{RL}} = w_1^{RL} \cdot L_{\phi}^{r_{\theta}} + w_2^{RL} \cdot L_\phi^{\text{SFT}},
\end{gather}  
where $\pi_{\phi}^{RL}(y | x)$ and $\pi^{SFT}(y | x)$ represent comments generated by RL model and the SFT model, respectively, \( w_1^{RL} \) and \( w_2^{RL} \) are trainable weights.

\subsection{Tree-of-Thought Refining}
To enhance the appeal of generated comments, we propose the knowledge-enhanced Tree-of-Thought (TOT) method which is inspired by ~\citet{yao2023tree}, \citet{chen2024talk}, and \citet{chen2022grow} as shown in Fig.~\ref{fig:v-framework}.

We first construct TOT instructions to generate TOT content related to the video with GPT-4. TOT instructions in Fig.~\ref{fig:v-framework} (denoted as ``$ToTv$'') include video description (denoted as ``Des'', such as ``\emph{A chef shares...}''), key events (denoted as ``Eve'', such as ``\emph{Selecting ingredients, Making sauce, Cooking crepes.}''), background knowledge (denoted as ``Kno'', such as ``\emph{French crepes are a traditional...}''), creative associations (denoted as ``Cre'', such as ``\emph{Crepes are like a piece of art...}''), and target audience (denoted as ``Aud'', such as ``\emph{People who love...}''). 
    
Next, we optimize the TOT content generated by GPT-4 with external knowledge bases like Wikidata in Fig.~\ref{fig:v-framework} (denoted as ``$ToTv_n$'') inspired by \citet{chen2023xmqas}.
Specifically, i) we first link entities from the background knowledge in the TOT content to corresponding entities in knowledge graphs with TagMe~\citep{ferragina2010tagme}. For example, the description from Wikidata of the ``crepes'' is ``\emph{a European recipe originating from France, ...}''.
ii) Second, we detect if there are errors in the description of the linked entities. 
iii) Third, we enrich the background knowledge with GPT-4 that is to ask GPT-4 to expand the content with an instruction. 
Afterwards, we obtain the optimized video-related TOT content including ``\emph{French crepes originated in the 13th century ...}'' as background knowledge and other four aspects maintain unchanged.
iv) Fourth, we determinate weights for the optimized video-related TOT content.
Knowledge-enhanced TOT is a tree where each content (i.e. each node) has a weight. We determine each weight's order with GPT-4 based on video titles, captions, descriptions, and audio content. 
We introduce a utility function \( U(c, W^T) \) for calculating the utility of a comment as follows:
\begin{gather}
U(c, W^T) = \sum_{i=1}^{5} w_j^T \cdot f_i^T(c),
\end{gather}
where \( c \) is a comment, \( W^T \) represents the weight set for each TOT dimension, \( f_i^T(c) \) denotes the utility of comment \( c \) concerning the TOT's \( i^{th} \) dimension. Weights are optimized through gradient descent and iterated until convergence: 
\begin{gather}
\frac{\partial U(c, W^T)}{\partial w_i^T} = f_i^T(c), \quad
w_i^{T'} = w_i^T - \alpha \times \frac{\partial U(c, W^T)}{\partial w_i^T},
\end{gather}
where \( \alpha \) as the learning rate.
In this video, the weights for video description and creative association are much higher than others.

Then, we generate comment-related TOT content in Fig.~\ref{fig:v-framework} (denoted as ``$ToTc$'') including comment description (denoted as ``Des''', such as ``\emph{French crepes are delicious.}''), key events (denoted as ``Eve''', such as ``\emph{The taste of French crepes.}''), background knowledge (denoted as ``Kno'''), creative associations (denoted as ``Cre''', and target audience (denoted as ``Aud''').

After that, we optimize comment-related TOT content with optimized video-related TOT content  in Fig.~\ref{fig:v-framework} (denoted as ``$ToTc_n$''). The optimized comment-related TOT content contains 
``\emph{French crepes originated in the 13th ...}'' as background knowledge, ``\emph{Crepes are like a piece of art ...}'' as creative association, ``\emph{People who love ...}'' as target audience. Other two aspects maintain unchanged.

Finally, we regenerate the comments with prompt engineering. The input is the best comment after RL (i.e.``\emph{The  French crepes are very delicious}'') and the optimized comment-related TOT content. The output is the best comment after TOT (i.e.``\emph{The French crepes are delicious food and like an artistic creation.}'').

\begin{table}[]
\centering
\resizebox{0.48\textwidth}{!}{%
\begin{tabular}{@{}l|cccc|cccccc@{}}
\toprule
                            & \multicolumn{4}{c|}{Popularity metrics} & \multicolumn{6}{c}{General metrics} \\ 
                            & Info    & Rele    & Crea     & Enga    & BLEU    & ROUGE & BLEURT	&COSMic& CIDEr  & METEOR  \\\midrule
UT                          & 62.13   & 44.55   & 28.62    & 38.23   & 11.02   & 23.21 & -0.45	&43.54& 45.14  & 6.31    \\
MML-CG                      & 68.52   & 47.32   & 34.74    & 43.55   & 19.45   & 34.89 & -0.31	&50.28& 54.55  & 12.34   \\
KLVCG                       & 71.51   & 51.03   & 39.12    & 52.54   & 24.76   & 40.39 & -0.27	&62.34& 62.65  & 16.65   \\
\texttt{ComHeat}                     & 93.54   & 84.78   & 83.93    & 88.32   & 59.21   & 79.36 & -0.13	&78.85& 92.47  & 49.55   \\ \midrule
$\uparrow$     & 22.03   & 33.75   & 44.81    & 35.78   & 34.45   & 38.97 & 0.14 	&16.51 & 29.82  & 32.90   \\
$\uparrow$(\%) & 30.81   & 66.14   & 114.54   & 68.10   & 139.14  & 96.48 & 107.69 	&26.48 & 47.60  & 197.60  \\ \bottomrule
\end{tabular}
}
\caption{\small The performance of \texttt{ComHeat} in comparison to other baselines in Chinese video comment datasets on our proposed \textsc{HotVCom}. Results of baselines are derived from being trained with our dataset.}
\label{tab:exp1}
\end{table}

\begin{table}[]
\centering
\resizebox{0.45\textwidth}{!}{%
\begin{tabular}{@{}l|ccccc|ccc@{}}
\toprule
                     & \multicolumn{5}{c|}{General  metrics}  & \multicolumn{3}{c}{Human  metrics} \\ 
Model                       & R@1  & R@5  & R@10 & MR$\downarrow$   & MRR  & Flue      & Rele      & Corr      \\\midrule
S2S-IC                      & 12.89 & 33.78 & 50.29 & 17.05 & 0.25 & 4.07      & 2.23       & 2.91      \\
FRNN                & 17.25 & 37.96 & 56.1  & 16.14 & 0.27 & 4.45      & 2.95       & 3.34      \\
UT         & 18.01 & 38.12 & 55.78 & 16.01 & 0.28 & 4.31      & 3.07       & 3.45      \\
MML-CG                      & 10.42 & 36.43 & 54.81 & 15.64 & 0.24 & -         & -          & -         \\
KLVCG                       & 13.49 & 41.43 & 59.31 & 13.09 & 0.28 & -         & -          & -         \\
KLVCG+                      & 14.88 & 44.81 & 62.5  & 11.91 & 0.30 & -         & -          & -         \\
\texttt{ComHeat}                     & 20.34 & 47.31 & 66.31 & 9.36  & 0.32 & 4.97      & 4.29       & 4.58      \\
Human                       & -     & -     & -     & -     & -     & 4.82      & 3.31       & 4.11      \\\midrule
$\uparrow$     & 2.33  & 2.50  & 3.81  & 2.55  & 0.02  & 0.15      & 0.98       & 0.47      \\
$\uparrow$(\%) & 12.94 & 5.58  & 6.10  & 21.41 & 6.67  & 3.11      & 29.61      & 11.44     \\ \bottomrule
\end{tabular}
}
\caption{\small The performance of \texttt{ComHeat} in comparison to other baselines in Chinese video comment datasets on the Livebot dataset. Results of baselines are derived from their published paper. MR$\downarrow$ means the lower of the value, the better of the method.}
\label{tab:exp2-1}
\end{table}

\section{Experiments}
In this section, we conduct extensive experiments to evaluate the performance of our proposed \texttt{ComHeat} framework in comparison to other baselines on generating hot-comments for \textsc{HotVCom}.

\subsection{Experimental Setups}
Our experiments are conducted on four Nvidia A100 GPUs, each with 80GB of memory, using PyTorch~\footnote{https://pytorch.org/} in Python~\footnote{https://www.python.org/}. For enhanced training efficiency, we utilize DeepSpeed. We set the maximum sequence length for both input and output sequences to maximum 1024 tokens. The training process is set to 10 epochs. We list detailed training hyperparameters in Table~\ref{tab:para} in the Appendix.

\subsection{Datasets, Baselines and Metrics}
We utilize four datasets and eleven baselines for comparison with details shown in Appendix~\ref{sec:base}. All results are reported
on the corresponding test sets or 20\% subset split from the original dataset. For public datasets, including VideoIC, Livebot, and MovieLC, the tests are conducted directly on these public datasets without training. For our self-collected datasets, which include \textsc{HotVCom} and the TikTok dataset, other baseline models have undergone training on our datasets. 

We classify our metrics into two categories: popularity metrics and general metrics.
Popularity metrics encompass informativeness, relevance, creativity, and user engagement. User engagement is essentially assessed through manual ratings from 1 to 5, where 1 means the worst and 5 means the best. The final scores will be scaled to 1-100.
We enroll three volunteers, and each of them is required to give scores for the randomly selected 500 videos with generated comments. We also calculate Inter-rater agreement of Krippendorff’s Alpha (IRA) to ensure the confidence of human ratings. For the controversial ratings which have low agreements ($<$0.7), we discard this comment.
General metrics encompass BLEU~\citep{BLEU}, ROUGE~\citep{ROUGE}, BLEURT~\citep{sellam2020bleurt}, COSMic~\citep{inan2021cosmic}, METEOR~\citep{banerjee2005meteor} and self-CIDEr~\citep{wang2019describing}, which assess the relevance and the diversity of generated comments.

\begin{table}[]
\centering
\resizebox{0.4\textwidth}{!}{%
\begin{tabular}{@{}l|ccccc|ccc@{}}
\toprule
                            & \multicolumn{5}{c|}{General  metrics} & \multicolumn{3}{c}{Human  metrics} \\ 
Model                       & R@1  & R@5  & R@10 & MR$\downarrow$   & MRR & Flue  & Rele & Corr \\\midrule
FRNN                        & 22.32 & 48.03 & 57.11 & 14.70 & 0.34 & 4.26     & 2.80      & 3.13        \\
UT                          & 26.34 & 54.66 & 64.37 & 12.66 & 0.39 & 4.18     & 3.49      & 3.93        \\
MML-CG                      & 27.50 & 56.12 & 65.68 & 12.21 & 0.40 & 4.44     & 3.84      & 4.15        \\
KLVCG                       & 34.08 & 57.22 & 71.37 & 9.51  & 0.46 & -        & -         & -           \\
KLVCG+                      & 34.11 & 57.33 & 71.32 & 9.43  & 0.46 & -        & -         & -           \\
\texttt{ComHeat}                     & 37.24 & 60.34 & 73.57 & 8.22  & 0.50 & 5.13     & 4.68      & 5.25        \\
Human                       & -     & -     & -     & -     & -    & 4.92     & 4.24      & 4.80        \\\midrule
$\uparrow$     & 3.13  & 3.01  & 2.20  & 1.21  & 0.04 & 0.21     & 0.44      & 0.45        \\
$\uparrow$(\%) & 9.18  & 5.25  & 3.08  & 12.83 & 9.37 & 4.27     & 10.38     & 9.38        \\ \bottomrule
\end{tabular}
}
\caption{\small The performance of \texttt{ComHeat} in comparison to other baselines on the VideoIC dataset. Results of baselines are derived from their published paper. MR$\downarrow$ means the lower of the value, the better of the method.}
\label{tab:exp2-2}
\end{table}

\subsection{Main Results}
In the domain of Chinese video comment generation, \texttt{ComHeat} consistently outperforms prior methods, as shown in Tables \ref{tab:exp1}, \ref{tab:exp2-1}, \ref{tab:exp2-2}, and \ref{tab:exp2-3}. These evaluations span general, popularity, and human-centric metrics on the \textsc{HotVCom} dataset, as well as on the established datasets: Livebot, VideoIC, and MovieLC.
Following the research by \citet{sun2023vico}, we collect 1000 English videos from TikTok. Results, presented in Table~\ref{tab:exp3}, show that \texttt{ComHeat} maintains its effectiveness in an English video comment generation task, illustrating its cross-linguistic capabilities.

\begin{table}[]
\centering
\resizebox{0.27\textwidth}{!}{%
\begin{tabular}{@{}lccccc@{}}
\toprule
                     & \multicolumn{5}{c}{General  metrics}  \\ 
                            & R@1  & R@5  & R@10 & MR$\downarrow$   & MRR  \\\midrule
UT                          & 6.24  & 18.98 & 31.98 & 23.76 & 0.15  \\
MML-CG                      & 6.25  & 17.81 & 30.64 & 24.7  & 0.14  \\
KLVCG                       & 7.36  & 19.26 & 29.92 & 24.87 & 0.15  \\
KLVCG+                      & 8.01  & 20.51 & 31.68 & 23.71 & 0.16  \\
\texttt{ComHeat}                     & 10.34 & 23.12 & 32.55 & 20.16 & 0.19  \\\midrule
$\uparrow$     & 2.33  & 2.61  & 0.57  & 3.55  & 0.03  \\
$\uparrow$(\%) & 29.09 & 12.73 & 1.78  & 14.97 & 18.75 \\ \bottomrule
\end{tabular}
}
\caption{\small The performance of \texttt{ComHeat} in comparison to other baselines in Chinese video comment datasets on the MovieLC dataset. Results of baselines are derived from their published paper. MR$\downarrow$ means the lower of the value, the better of the method.}
\label{tab:exp2-3}
\end{table}

\begin{table}[]
\centering
\resizebox{0.5\textwidth}{!}{%
\begin{tabular}{@{}l|cccc|cccccc@{}}
\toprule
                            & \multicolumn{4}{c|}{Popularity metrics} & \multicolumn{6}{c}{General metrics} \\ 
                            & Info     & Rele    & Crea    & Enga    & BLEU    & ROUGE &BLEURT	&COSMic & CIDEr  & METEOR  \\\midrule
UT-ResNet                   & 52.35    & 27.38   & 10.56   & 12.56   & 4.63    & 7.23 & -1.84	&46.98 & 30.55  & 1.22    \\
UT-CLIP                     & 55.31    & 31.58   & 17.57   & 20.33   & 6.25    & 10.5 &-1.57	&55.35 & 32.57  & 4.43    \\
CLIP4C                      & 60.31    & 37.82   & 20.39   & 25.7    & 6.31    & 10.57 & -1.22	&61.66& 33.68  & 4.02    \\
GIT-2                       & 63.59    & 41.33   & 24.69   & 32.11   & 11.67   & 15.32 &-0.98	&68.13 & 35.92  & 8.57    \\
ViCo-r                      & 65.17    & 46.76   & 28.33   & 38.55   & 11.33   & 15.58 & -0.79	&72.44& 37.82  & 8.34    \\
ViCo-u                      & 72.03    & 52.77   & 34.34   & 47.53   & 14.36   & 20.33 &-0.83	&73.25 & 44.37  & 10.25   \\
ViCo-f                      & 80.35    & 60.32   & 52.12   & 63.47   & 23.78   & 29.66 &-0.78	&73.99 & 50.41  & 18.9    \\
\texttt{ComHeat}                     & 83.46    & 66.14   & 63.53   & 71.76   & 27.55   & 30.82& -0.55	&77.27 & 53.89  & 20.31   \\\midrule
$\uparrow$     & 3.11     & 5.82    & 11.41   & 8.29    & 3.77    & 1.16  &0.23 	&3.28  & 3.48   & 1.41    \\
$\uparrow$(\%) & 3.87     & 9.65    & 21.89   & 13.06   & 15.85   & 3.91  &41.82& 	4.43  & 6.90   & 7.46    \\ \bottomrule
\end{tabular}
}
\caption{\small The performance of \texttt{ComHeat} in comparison to other baselines in English video comment datasets on the randomly collected TikTok dataset.}
\label{tab:exp3}
\end{table}

\subsection{Ablation Study}
\begin{figure}[!t]
  \centering
  \includegraphics[width=0.95\linewidth]{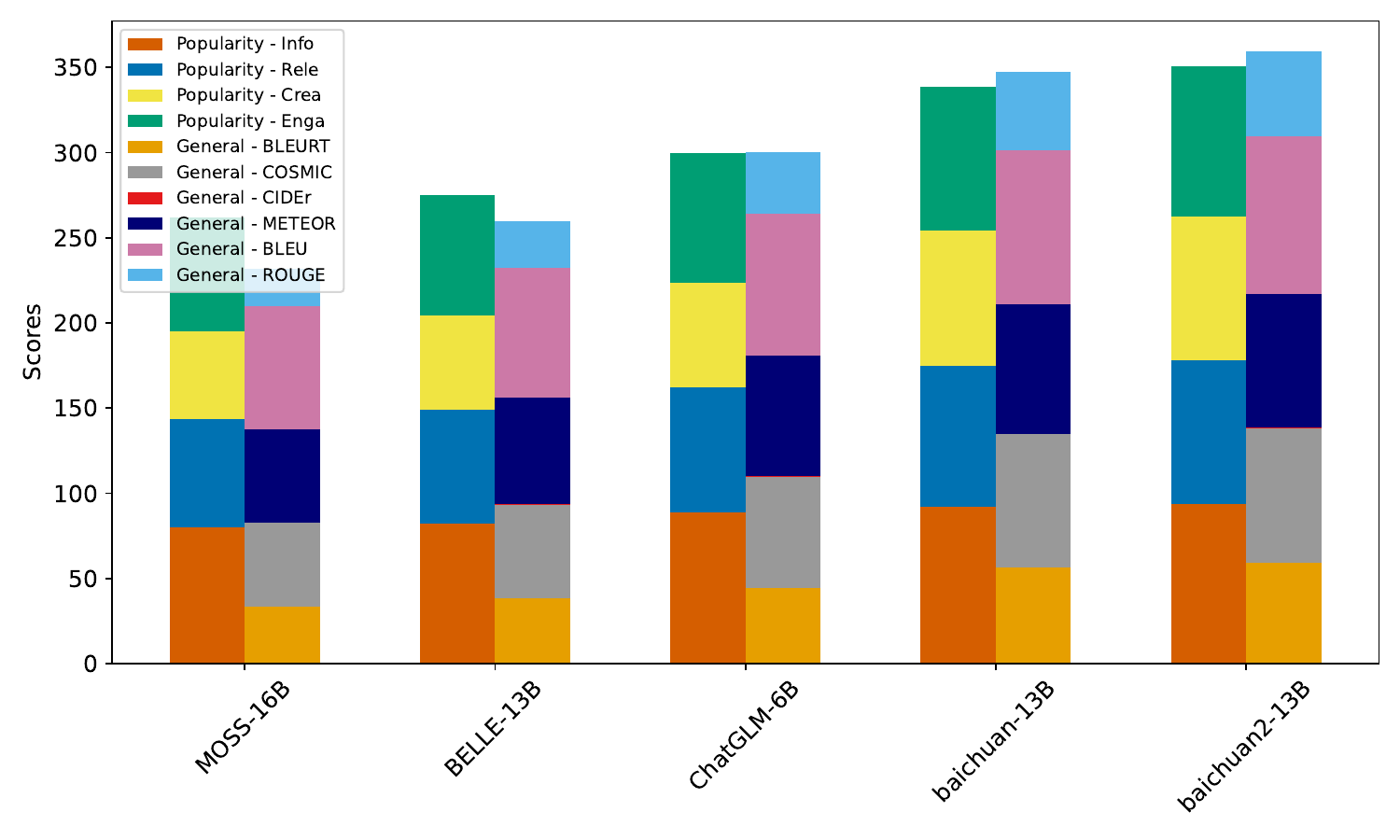}
  \caption{\small The performance of adopting other LLMs as backbones in Chinese video comment generation.}
  \label{fig:v-exp7}
\end{figure}

In Table~\ref{tab:exp6}, we analyze the impact of each \texttt{ComHeat} component. Excluding SFT (``w/o SFT''), RL (``w/o RL''), and TOT (``w/o TOT'') leads to the biggest drops in results, showing their importance in video comment generation. TOT is more effective than COT, as COT tends to produce longer comments. Basic and auxiliary rewards (``RW w/ br'', ``RW w/ ar'') have similar effects. The results also show a preference for informativeness over creativity or knowledge (``w/o K'', ``w/o I'', ``w/o C''). The knowledge-based correction (``w/o corr'') has limited influence. Overall, \texttt{ComHeat} performs better than all other setups, highlighting the combined strength of its parts.
We also tested other backbones like MOSS-16B~\citep{sun2023moss}, BELLE-13B~\citep{belle2023exploring}, ChatGLM-6B~\citep{du2022glm}, and baichuan-13B, as seen in Fig.~\ref{fig:v-exp7}. In summary, backed by the baichuan2-13B model, \texttt{ComHeat} stands out, showing its potential in video comment generation.

\begin{table}[]
\centering
\resizebox{0.5\textwidth}{!}{%
\begin{tabular}{@{}l|cccc|cccccc@{}}
\toprule
         & \multicolumn{4}{c|}{Popularity metrics} & \multicolumn{6}{c}{General metrics} \\ 
         & Info     & Rele    & Crea    & Enga    & BLEU    & ROUGE  & BLEURT	&COSMic&CIDEr  & METEOR  \\\midrule
\multicolumn{11}{c}{Training-related}                                                    \\\midrule
w/o V    & 74.47    & 72.28   & 66.34   & 74.69   & 46.12   & 65.36  & -0.29 	&60.18 &81.09  & 35.76   \\
w/o SFT  & 40.37    & 37.52   & 32.13   & 38.96   & 22.23   & 34.76  & -0.60 &	40.45 &58.37  & 14.24   \\
RW w/ br & 81.24    & 77.20   & 73.34   & 81.03   & 50.22   & 69.56  & -0.25 	&66.99 &84.12  & 41.99   \\
RW w/ ar & 82.05    & 78.24   & 75.37   & 81.29   & 51.66   & 71.32  & -0.22 	&68.83& 85.06  & 43.89   \\
w/o RL   & 67.56    & 62.53   & 56.87   & 63.22   & 38.17   & 59.46  & -0.38 	&53.17 &75.71  & 30.68   \\\midrule
\multicolumn{11}{c}{TOT-related}                                                         \\\midrule
w/o TOT  & 60.22    & 56.34   & 51.39   & 54.45   & 33.35   & 53.52  & -0.42 	&48.10& 70.33  & 26.28   \\
w/o corr & 89.66    & 82.00   & 80.37   & 85.93   & 56.78   & 76.13  & -0.15 	&75.84& 88.14  & 47.01   \\
w/o K    & 85.23    & 80.86   & 78.83   & 83.55   & 54.84   & 73.38  & -0.19 	&71.60 &86.26  & 46.03   \\
w/o I    & 78.30    & 75.45   & 69.25   & 77.31   & 48.21   & 68.36  & -0.26 	&63.11 &82.17  & 38.69   \\
w/o C    & 92.81    & 83.27   & 82.76   & 87.35   & 58.24   & 78.44  & -0.14 	&77.32 &91.58  & 48.03   \\
TOT2COT  & 71.35    & 70.78   & 63.13   & 70.52   & 43.48   & 63.66  & -0.33 	&56.46 &78.53  & 33.69   \\
\midrule
\texttt{ComHeat}  & 93.54    & 84.78   & 83.93   & 88.32   & 59.21   & -0.13	&78.85& 79.36  & 92.47  & 49.55   \\ \bottomrule
\end{tabular}
}
\caption{\small The contributions of each component of our proposed \texttt{ComHeat}.}
\label{tab:exp6}
\end{table}

\subsection{Case Study}
Examples of our proposed \texttt{ComHeat}'s performance are in the Appendix from Table~\ref{tab:case-more1} to Table~\ref{tab:case-more10} across various themes. We find that Unified Transformer, MML-CG, and KLVCG give basic comments, such as mentioning ``cat food'' in a query-like form, lacking directness. Our proposed \texttt{ComHeat} provides more detailed and emotional content. When compared to human comments, \texttt{ComHeat} has a different but engaging perspective, showing areas to improve.
Comparing \texttt{ComHeat} and human generated comments as in Table~\ref{tab:error}, \texttt{ComHeat} is off-topic or too formal at times. Yet, its comments can be popular and resonate with some audiences due to their unique or funny nature, as shown in Table~\ref{tab:humanhigh}. This shows the great potential of \texttt{ComHeat} in adding variety to online comments.

\section{Related Work}
\subsection{Video Comment Generation}
Recent research predominantly emphasizes live or synchronized video comments like danmaku. Systems like GraspSnooker by ~\citet{sun2019graspsnooker} and the rap-style generator by ~\citet{jumneanbun2020rap} exemplify this trend. Major datasets like VideoIC by ~\citet{wang2020videoic} and innovations like the open-domain approach by ~\citet{marrese2022open} have been introduced. While ~\citet{chen2023knowledge} targets long videos, ~\citet{ma2019livebot} merges visual and textual contexts. 
Our focus diverges towards generating alluring comments for full videos. Despite ~\citet{sun2023vico}'s contributions with the ViCo-20k dataset, they lack comprehensive engagement metrics. Notably, datasets like LiveBot~\citep{ma2019livebot} are live-comment centric, and ~\citet{sun2023vico}'s dataset, being English, might be less fitting for Chinese scenarios.

\subsection{Video Caption Generation}
Research on video caption generation has seen diverse approaches, which is widely used in various scenarios~\citep{202407.0981, 202407.2102,li2024vqa}. ~\citet{qi2023goal} launched GOAL, emphasizing Knowledge grounded Video Captioning (KGVC). Techniques like attention-based learning have been explored by ~\citet{ji2022attention}, while ~\citet{song2022contextual} introduce the Contextual Attention Network (CANet) for context-rich learning. Meanwhile, ~\citet{yan2022gl} and ~\citet{babavalian2023learning} offer unique frameworks for improved caption relevance and diversity. ~\citet{yang2023vmsg} propose a weighted semantic model, VMSG. 
Among multi-modal language model advancements~\citep{chen2024recent}, our work generate richer video descriptions, aiding in effective comment generation.

\section{Conclusions and Future Work}
In conclusion, this study underscores the crucial role of ``hot-comments'' in enhancing video visibility, setting them apart from the prevalent ``danmaku'' comments. Through the introduction of the extensive Chinese video hot-comment dataset \textsc{HotVCom} and the innovative \texttt{ComHeat} framework, we offer a novel approach to generate relevant and engaging video comments. 
In the future, we plan to ensure the ethics and fairness of the \texttt{ComHeat} framework from
equal access and fair algorithms. We will provide low-cost or free tools and services, enabling smaller brands to enhance their video visibility.
We also intend to include even more categories, especially those that are more niche. 
Moreover, we will explore cross-lingual perspectives for video comment generation.

\section*{Limitations}
While our research offers promising advancements in video hot-comment generation, it also presents certain limitations. First, our approach predominantly caters to Chinese short videos, which may constrain its applicability in diverse linguistic and cultural contexts. Second, the reliance on the \texttt{ComHeat} framework assumes that visual, auditory, and textual data are always present and of high quality, which might not always be the case in real-world scenarios. Furthermore, the optimization techniques employed, though effective, may not capture the full depth of human creativity. Lastly, while the dataset we introduce is comprehensive, it is inherently subject to the biases and characteristics of its source, potentially affecting the generalizability of our findings.

\section*{Ethic Statement}
Throughout the course of our research, we have maintained unwavering commitment to the highest ethical standards. Our foremost priorities include ensuring transparency, fairness, and the utmost respect for all participants involved in this study. We have taken extensive measures to safeguard user identities and protect privacy through a meticulous anonymization process applied to all data within our dataset.
Our overarching objective is to enrich the user experience and interactions on video platforms while simultaneously upholding the principles of individual rights and human dignity. In a world where the influence of AI and technology continues to expand, we remain acutely aware of the profound impact these innovations can have on society as a whole.
It is essential to acknowledge that in the realm of AI-driven comment generation, there exists the potential for harmful comments to emerge. Thus, we remain vigilant and resolute in our commitment to responsible research practices, with a strong emphasis on ethical considerations and societal well-being.

\section*{Acknowledgements}
This work is supported by Ant Group Research Intern Program,
Science and Technology Commission of Shanghai Municipality Grant (No. 22511105902), Shanghai Municipal Science and Technology Major Project (No.2021SHZDZX0103), the National Natural Science Foundation of China (No.62072323), Shanghai Science and Technology Innovation Action Plan (No. 22511104700), and the Zhejiang Lab Open Research Project (NO. K2022NB0AB04).

\bibliography{anthology,main}

\appendix

\begin{figure*}[!h]
  \centering
  \includegraphics[width=\linewidth]{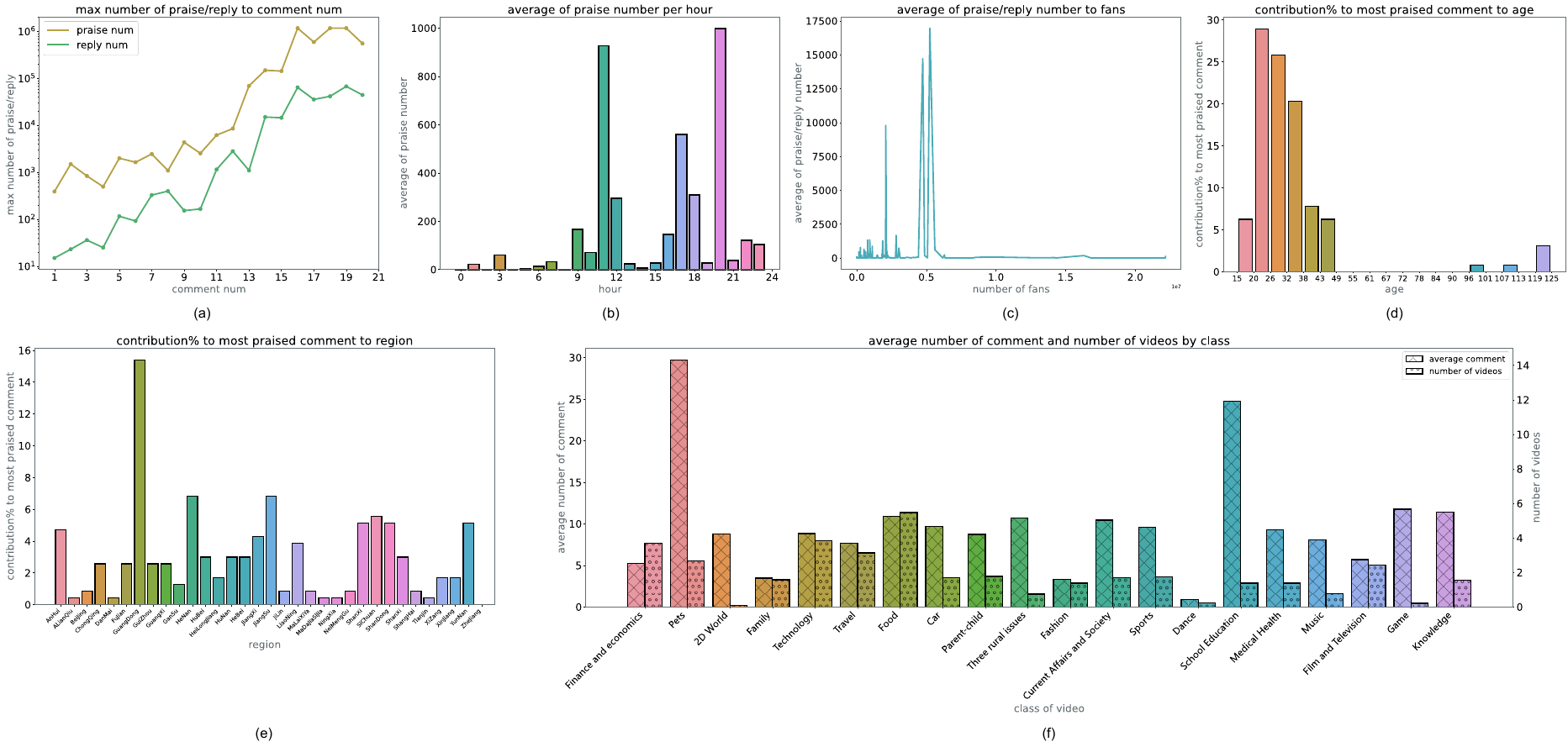}
  \caption{\small The exploratory data analysis on \textsc{HotVCom}.}
  \label{fig:v-eda}
\end{figure*}

\section{Details of data construction}
\label{sec:data}
Specifically, in the OCR process, we use FFmpeg~\footnote{https://ffmpeg.org/} to transform videos into uniform frame sequences. For enhanced caption accuracy, the OpenCV library~\footnote{https://opencv.org/} is employed, allowing dynamic adjustment of RGB thresholds, initially set at 127. Gaussian blurring, with parameters tailored to each video, is applied for noise mitigation.
During ASR, FFmpeg again comes into play, isolating audio streams and converting them to linear 16 PCM format. To heighten speech extraction accuracy, the noisereduce library~\footnote{https://github.com/timsainb/noisereduce} is used, designating the first audio second as the noise benchmark.
For keyframe extraction, OpenCV is utilized to derive color histograms and contours, and the Local Binary Pattern (LBP) algorithm~\citep{pietikainen2010local} extracts texture features. After feature normalization, K-means clustering categorizes frames, with each cluster's central frame chosen as the keyframe. The extraction's precision is validated using SSIM~\citep{wang2004image} and PSNR measurements against the original video content, and 1\% of videos undergo manual validation.
For theme categorization, Video-ChatGPT discerns video content, while ChatGPT offers text insights. 
We randomly select 1\% of videos for manually checking for accurate tagging.

\section{Exploratory Data Analysis}
\label{sec:eda}
We make detailed exploratory data analysis on the constructed \textsc{HotVCom} as illustrated in Fig.~\ref{fig:v-eda}.

Fig.~\ref{fig:v-eda}(a) shows the relationship between the maximum number of likes/replies and the number of comments. The X-axis represents the number of comments, and the Y-axis (logarithmic scale) represents the maximum number of likes and replies. The graph shows a trend of increasing maximum likes and replies as the number of comments increases.

Fig.~\ref{fig:v-eda}(b) shows the average number of likes during different hours of the day. There are noticeable peaks in average likes during specific times, such as at 3 PM and 9 PM.

Fig.~\ref{fig:v-eda}(c) shows the relationship between the number of fans and the average number of likes/replies. The X-axis represents the fan number ratio, and the Y-axis represents the average number of likes/replies. The chart shows that the average number of likes/replies initially increases with the number of fans but then decreases.

Fig.~\ref{fig:v-eda}(d) shows the contribution percentage of users of different age groups to the most praised comment. It can be observed that certain age groups, such as those between 25 to 29 years old, have a particularly high contribution rate.

Fig.~\ref{fig:v-eda}(e) shows the contribution percentage of different regions to the most praised comment. There is a significant variance in contributions by region, with some regions contributing much more than others.

Fig.~\ref{fig:v-eda}(e) shows the average number of comments and the number of videos across different video categories. The X-axis represents the category of the video, the left Y-axis represents the average number of comments, and the right Y-axis represents the number of videos. It is evident that some categories, like ``Finance and Economics'', have a high average number of comments, while others have a larger quantity of videos, such as ``Knowledge''.

Combining the above insights, we can conclude that:
i) Popular comments tend to increase in likes and replies with an increasing number of comments.    
ii) Users are more active during certain periods, indicating peak times for liking videos.
iii) Users with more fans tend to have comments that receive more likes and replies, although this trend is not linear.
iv) Contributors to the most popular comments are more concentrated within certain age brackets.
v) There is a significant regional difference in contributions to the most popular comments, suggesting that regional culture may affect comment interaction.
vi) Different video categories perform differently in terms of attracting comments and the production of videos.

\section{More experiments}
Moreover, we introduce a dedicated dataset using video descriptions as labels to augment the information content of the generated captions. \texttt{ComHeat}'s performance, when compared to baselines in Chinese video comment generation, is evident in Table~\ref{tab:exp4}. This underlines \texttt{ComHeat}'s prowess in delivering contextually relevant Chinese captions. Furthermore, on the English video caption generation datasets MSVD~\citep{chen2011collecting} and MSR-VTT~\citep{xu2016msr}, as presented in Table~\ref{tab:exp5}, we list the results of the baselines based on the corresponding paper, and find \texttt{ComHeat} consistently excels, marking its robustness in bilingual caption generation tasks.

\begin{table}[]
\centering
\resizebox{0.35\textwidth}{!}{%
\begin{tabular}{@{}lcccccc@{}}
\toprule
                            & BLEU  & ROUGE &BLEURT	&COSMIC& CIDEr & METEOR \\ \midrule
UT                          & 15.67 & 28.65&  -0.47 &	50.12 & 44.69   & 8.33  \\
MML-CG                      & 22.70 & 35.08  &-0.39 	&56.34 & 56.67   & 17.83 \\
KLVCG                       & 28.00 & 39.64  &-0.37 &	58.92 & 62.83   & 22.55 \\
\texttt{ComHeat}                     & 41.21 & 58.19  &-0.26 	&66.31 & 85.36   & 30.47 \\\midrule
$\uparrow$     & 13.21 & 18.55 &0.11 	&7.39  & 22.53   & 7.92  \\
$\uparrow$(\%) & 47.18 & 46.80& 42.31 	&12.54  & 35.86   & 35.12 \\ \bottomrule
\end{tabular}
}
\caption{\small The performance of \texttt{ComHeat} in comparison to other baselines in Chinese video comment datasets on the Chinese video caption dataset sourced from our proposed \textsc{HotVCom}.}
\label{tab:exp4}
\end{table}

\begin{table}[]
\resizebox{0.5\textwidth}{!}{%
\begin{tabular}{@{}l|cccc|cccc@{}}
\toprule
                            & \multicolumn{4}{c|}{MSVD}        & \multicolumn{4}{c}{MSR-VTT}    \\ 
                            & BLEU  & ROUGE & CIDEr  & METEOR & BLEU  & ROUGE & CIDEr & METEOR \\\midrule
POSRL                       & 53.90 & 72.10 & 91.00  & 34.90  & 41.30 & 62.10 & 53.40 & 28.70  \\
ORG-TRL                     & 54.30 & 73.90 & 95.20  & 36.40  & 43.60 & 62.10 & 50.90 & 29.70  \\
O2NA                        & 55.40 & 74.50 & 96.40  & 37.40  & 41.60 & 62.40 & 51.10 & 28.50  \\
OA-BTG                      & 56.90 & -     & 90.60  & 36.20  & 41.40 & -     & 46.90 & 28.20  \\
GL-RG+IT                    & 60.50 & 76.40 & 101.00 & 38.90  & 46.90 & 65.70 & 60.60 & 31.20  \\
\texttt{ComHeat}                     & 63.90 & 80.30 & 104.20 & 42.70  & 51.30 & 69.90 & 66.30 & 34.70  \\\midrule
$\uparrow$    & 3.40  & 3.90  & 3.20   & 3.80   & 4.40  & 4.20  & 5.70  & 3.50   \\
$\uparrow$(\%) & 5.62  & 5.10  & 3.17   & 9.77   & 9.38  & 6.39  & 9.41  & 11.22  \\ \bottomrule
\end{tabular}
}
\caption{\small The performance of \texttt{ComHeat} in comparison to other baselines in English caption datasets including MSVD and MSR-VTT.}
\label{tab:exp5}
\end{table}


\begin{table*}[]
\centering
\resizebox{0.8\textwidth}{!}{%
\begin{tabular}{@{}lll@{}}
\toprule
Parameter Name & Parameter Value & Parameter Meaning                                                 \\ \midrule
$L_{min}$      & 1               & Minimum optimal comment length boundary                           \\
$L_{max}$      & 50              & Maximum optimal comment length boundary                           \\
$\alpha$       & 0.05            & Penalty adjustment coefficient for exceeding optimal length       \\
$w_1^I$        & 0.6             & Weight for the length penalty in informativeness score            \\
$w_2^I$        & 0.6             & Weight for the vocabulary diversity in informativeness score      \\
$w_1^R$        & 0.6             & Weight for the keyword matching degree in relevance score         \\
$w_2^R$        & 0.6             & Weight for the context matching degree in relevance score         \\
$w_1^C$        & 0.6             & Weight for the rhetorical technique score in creativity score     \\
$w_2^C$        & 0.6             & Weight for the trending term score in creativity score            \\
$k_r, k_t$     & 1               & Slope of the sigmoid function for rhetorical and trending scores  \\
$b_r, b_t$     & -1              & Offset of the sigmoid function for rhetorical and trending scores \\
$w_1^U$        & 0.5             & Weight for the number of likes in user engagement score           \\
$w_2^U$        & 0.5             & Weight for the number of replies in user engagement score         \\
$k_u$          & 1               & Slope of the sigmoid function for user engagement score           \\
$b_u$          & -1              & Offset of the sigmoid function for user engagement score          \\
$w^I$          & 0.2             & Weight for the informativeness score in comprehensive score       \\
$w^R$          & 0.2             & Weight for the relevance score in comprehensive score             \\
$w^C$          & 0.2             & Weight for the creativity score in comprehensive score            \\
$w^U$          & 0.4             & Weight for the user engagement score in comprehensive score       \\
$\alpha$       & 0.5             & Adjustable weight for linear fusion of text and visual features   \\
$\beta$        & 0.5             & Adjustable weight for linear fusion of text and visual features   \\
$w_1^S$        & 0.8             & Weight for the cross-entropy loss in SFT process                  \\
$w_2^S$        & 0.2             & Weight for the mean squared error loss in SFT process             \\
$w_1^{RL}$     & 0.3             & Weight for the reward model score in RL model                     \\
$w_2^{RL}$     & 0.7             & Weight for the KL divergence loss in RL model                     \\
$w_j^T$        & [0,1]      & Weight for the Tree-of-Thought (TOT) dimensions                   \\ \bottomrule
\end{tabular}%
}
\caption{Detailed training hyperparameters.}
\label{tab:para}
\end{table*}

\section{Baselines}
\label{sec:base}
We elaborate baselines as follows:

S2S-IC~\citep{ma2019livebot} integrates both visual and textual information. It utilizes two separate encoders to process images and comments. The outputs of these encoders are then concatenated and fed into an LSTM decoder to generate output comments.

FRNN~\citep{ma2019livebot}, the Fusional RNN model, comprises three components: a video encoder, a text encoder, and a comment decoder. The video encoder processes a sequence of consecutive frames using an LSTM layer on top of a CNN layer, while the text encoder encodes surrounding live comments into vectors also using an LSTM layer. The comment decoder then generates the live comment based on these encodings.

UT/UT-ResNet~\citep{ma2019livebot}, the unified transformer model, uses a linear structure to capture the dependencies between comments and videos. It consists of a video encoder, a text encoder, and a comment decoder.

UT-CLIP~\citep{sun2023vico} uses CLIP instead of ResNet for feature extraction based on the unified transformer model framework.

MML-CG~\citep{wang2020videoic}, the Multimodal Multitask Learning framework for Comments Generation, first extracts different modality features using various encoders. These features are then integrated by a multimodal encoder, which jointly optimizes two tasks: temporal relation prediction and comment generation, in an end-to-end manner.

KLVCG~\citep{chen2023knowledge} comprises independent modality encoders for video context, comment context, and external knowledge, along with a cross encoder and a decoder.

KLVCG+~\citep{chen2023knowledge} represents the results of KLVCG pre-trained on a mixture of Livebot, VideoIC, MovieLC datasets.

CLIP4C~\citep{tang2021clip4caption}, the CLIP4Caption framework for video captioning, involves two stages of training. Initially, a video-text matching network is pre-trained on the MSR-VTT dataset for better visual representation. Then, this pre-trained network is used as a video feature extractor in the fine-tuning stage. The system inputs a sequence of frame embeddings to a video encoder, linked to a decoder that generates text. For ensemble purposes, multiple caption models with different layers of encoder and decoder are trained, and their outputs are combined for a robust final result.

GIT-2~\citep{wang2022git} comprises a single image encoder and a text decoder, pre-trained on 0.8 billion image-text pairs with a language modeling task. It first uses a contrastive task to pre-train the image encoder, followed by a generation task to pre-train both the image encoder and text decoder.

ViCo-r, ViCo-u, and ViCo-f~\citep{sun2023vico} are sourced from ViCo. ViCo-r uses a generator trained with randomly sampled comments, ViCo-u employs a generator trained with uniqueness-guided sampling, and ViCo-f proposes a model called ViCo with three novel designs focusing on engagement quantification, automatic engagement evaluation, and alleviating scarcity of high-quality comments using reward feedback.

POSRL~\citep{wang2019controllable} consists of a gated fusion network, a POS sequence generator, and a description generator. It utilizes a self-critical sequence training method for reinforcement learning, focusing on exploiting relationships among different video features and POS tags of descriptions for generating comprehensive and accurate captions.

ORG-TRL~\citep{zhang2020object} follows an encoder-decoder framework, with an object relational graph at its core. It dynamically learns the interaction among different objects and attentively aggregates visual features to generate descriptions. The learning process includes both teacher-enforced and teacher-recommended strategies.

O2NA~\citep{liu2021o2na} is based on the Transformer decoder, comprises an object predictor, a length predictor, and two Transformer decoders, focusing on generating objects in parallel and then linking them to form fluent captions.

OA-BTG~\citep{zhang2019object} is an advanced video captioning model that follows an encoder-decoder framework. It begins by extracting frames and object regions from a video, constructing a bidirectional temporal graph to capture complex temporal dynamics. The model then aggregates these features into discriminative representations using learnable VLAD models, focusing on both local object details and global frame context. Finally, in the decoding stage, it integrates these representations and employs GRU units with hierarchical attention to generate detailed video descriptions, effectively balancing the contributions of multiple objects.

GL-RG+IT~\citep{yan2022gl} is with an encoder-decoder architecture, includes a global-local encoder and a captioning decoder. It emphasizes aggregating different frame features to enrich global-local vision representations and translates these into natural language sentences using an incremental training strategy.

\begin{table}[]
\centering
\resizebox{0.3\textwidth}{!}{%
\begin{tabular}{@{}lll@{}}
\toprule
Model   & Engagement & Agreement \\ \midrule
UT      & 38.23      & 0.82      \\
MML-CG  & 43.55      & 0.87      \\
KLVCG   & 52.54      & 0.84      \\
ComHeat & 88.32      & 0.91      \\ \bottomrule
\end{tabular}%
}
\caption{The agreement scores for comments generated by baseline models and \texttt{ComHeat} on the proposed \textsc{HotVCom}.}
\label{tab:agree-h}
\end{table}

\begin{table}[]
\centering
\resizebox{0.3\textwidth}{!}{%
\begin{tabular}{@{}lll@{}}
\toprule
Model     & Engagement & Agreement \\ \midrule
UT-ResNet & 12.56      & 0.80      \\
UT-CLIP   & 20.33      & 0.83      \\
CLIP4C    & 25.70      & 0.82      \\
GIT-2     & 32.11      & 0.81      \\
ViCo-r    & 38.55      & 0.85      \\
ViCo-u    & 47.53      & 0.84      \\
ViCo-f    & 63.47      & 0.89      \\
ComHeat   & 71.76      & 0.90      \\ \bottomrule
\end{tabular}%
}
\caption{The agreement scores for comments generated by baseline models and \texttt{ComHeat} on Tiktok.}
\label{tab:agree-t}
\end{table}

\begin{table*}[]
\centering
\resizebox{0.8\textwidth}{!}{%
\begin{tabular}{@{}lllllll@{}}
\toprule
Model   & Fluency & Fluency-Agreement & Relevance & Relevance-Agreement & Correctness & Correctness-Agreement \\ \midrule
ComHeat & 4.97    & 0.89              & 4.29      & 0.86                & 4.58        & 0.84                  \\ \bottomrule
\end{tabular}%
}
\caption{The agreement scores for comments generated by \texttt{ComHeat} on Livebot.}
\label{tab:agree-l}
\end{table*}

\begin{table*}[]
\centering
\resizebox{0.8\textwidth}{!}{%
\begin{tabular}{@{}lllllll@{}}
\toprule
Model   & Fluency & Fluency-Agreement & Relevance & Relevance-Agreement & Correctness & Correctness-Agreement \\ \midrule
ComHeat & 5.13    & 0.88              & 4.68      & 0.85                & 5.25        & 0.82                  \\ \bottomrule
\end{tabular}%
}
\caption{The agreement scores for comments generated by \texttt{ComHeat} on VideoIC.}
\label{tab:agree-v}
\end{table*}

\begin{table*}[]
\centering
\resizebox{0.9\textwidth}{!}{%
\begin{tabular}{@{}ll@{}}
\toprule
Score & \multicolumn{1}{l}{Description}                                                                                                     \\ \midrule
Score 1                   & Disengaging: The comment is disengaging and irrelevant, failing to stimulate any noticeable user interaction.                       \\
Score 2                   & Marginally Engaging: The comment is marginally engaging, eliciting only minimal user interaction.                                   \\
Score 3                   & Moderately Engaging: The comment is moderately engaging, sparking some interest and interaction.                                    \\
Score 4                   & Highly Engaging: The comment is highly engaging, resulting in significant user interaction.                                         \\
Score 5                   & Exceptionally Engaging: The comment is exceptionally engaging, generating substantial interest and high levels of user interaction. \\ \bottomrule
\end{tabular}%
}
\caption{Annotation protocol of human evaluation.}
\label{tab:criteria}
\end{table*}

\section{Metrics}

\subsection{Annotation protocol}
The manual metrics Fluency (short as Flue), Relevance (short as Rele), and Correctness (short as Corr) derive from ~\citet{ma2019livebot}. In our study, we do not re-evaluate the Fluency (short as Flue), Relevance (short as Rele), and Correctness (short as Corr) for each model in ~\citet{wang2020videoic} and ~\citet{ma2019livebot}; these scores are taken directly from the values published in their respective articles. We also list the corresponding content of three metrics as follows:

Fluency is designed to measure whether the generated live comments are fluent, setting aside their relevance to videos. Relevance is designed to measure the relevance between the generated live comments and the videos. Correctness is designed to synthetically measure the confidence that the generated live comments are made by humans in the context of the video. For all three aspects, we stipulate that the score should be an integer in \{1, 2, 3, 4, 5\}, with higher scores being better. Scores are evaluated by three human annotators, and we take the average of the three raters as the final result.

It is important to note that the human metrics defined in our paper include only user engagement. Informativeness, relevance, and creativity are automatic metrics. We have explained that user engagement represents the level of users' interaction with a comment. It is essentially assessed through manual ratings from 1 to 5, where 1 means the worst and 5 means the best. The final scores will be scaled to 1-100. Below, we also list the specific protocol as shown in Table~\ref{tab:criteria}.

\subsection{Agreement scores among annotators}
We list the agreement scores for comments generated by each model and our method, ComHeat, on the proposed HOTVCOM (see Table~\ref{tab:agree-h}), Livebot (see Table~\ref{tab:agree-l}), VideoIC (see Table~\ref{tab:agree-v}), and Tiktok (see Table~\ref{tab:agree-t}) datasets. To facilitate easy reference, we retain the human evaluation scores and provide the corresponding agreement scores next to them.

\section{More cases}
\label{sec:cases}
We list more cases as shown from Table~\ref{tab:case-more1} to Table~\ref{tab:case-more10}.

\begin{CJK}{UTF8}{gkai}
\begin{table*}[htbp]
\centering
\resizebox{\textwidth}{!}{%
\begin{tabular} {p{2cm}|p{1.5cm}p{3cm}|p{1.5cm}p{2.5cm}|p{2cm}p{2cm}|p{2cm}p{3cm}}
\toprule
Video&\multicolumn{2}{c|}{\raisebox{-0.22\height}{\includegraphics[width=80pt]{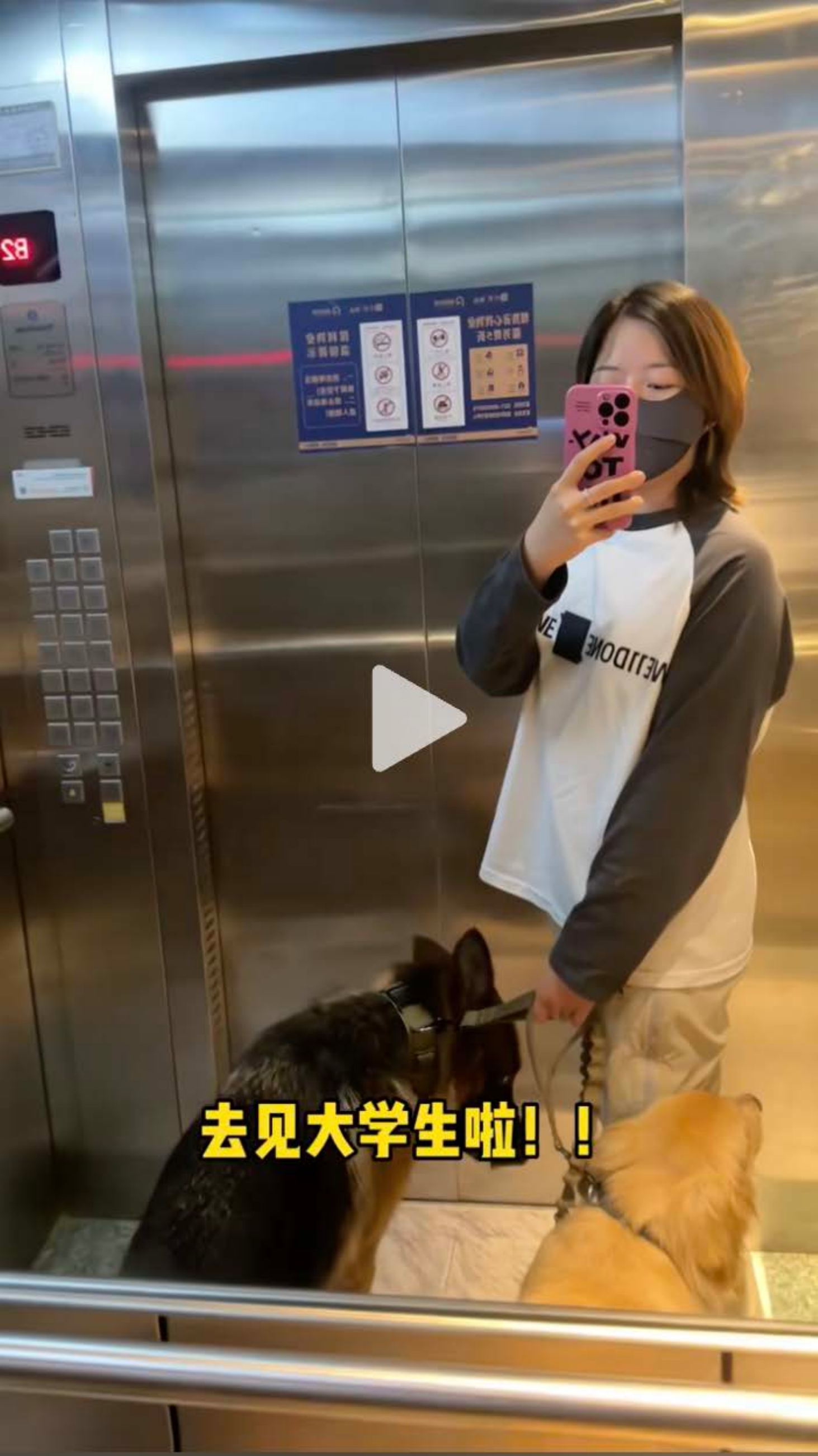}}}	&\multicolumn{2}{c|}{\raisebox{-0.22\height}{\includegraphics[width=80pt]{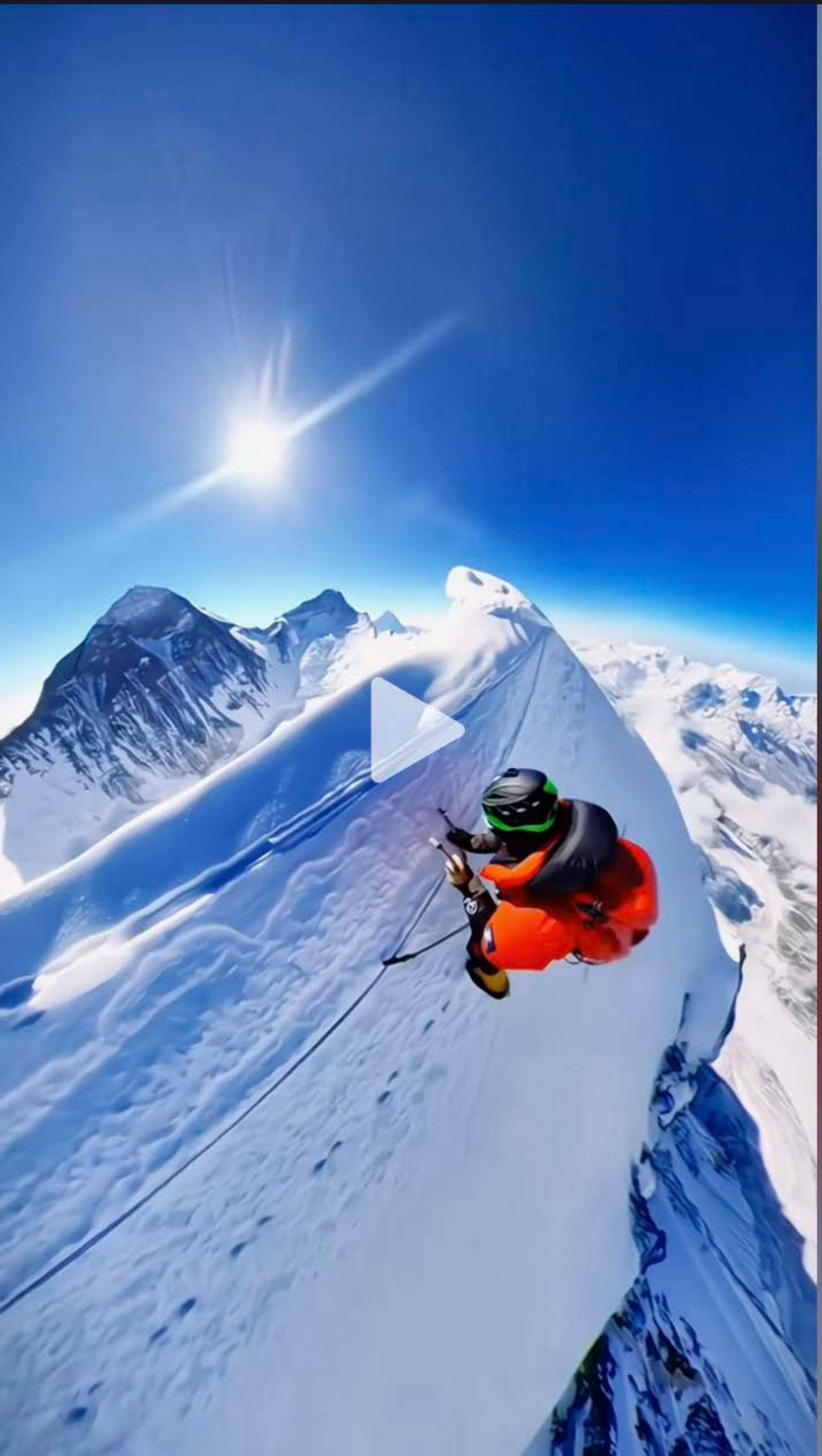}}}	&\multicolumn{2}{c|}{\raisebox{-0.22\height}{\includegraphics[width=80pt]{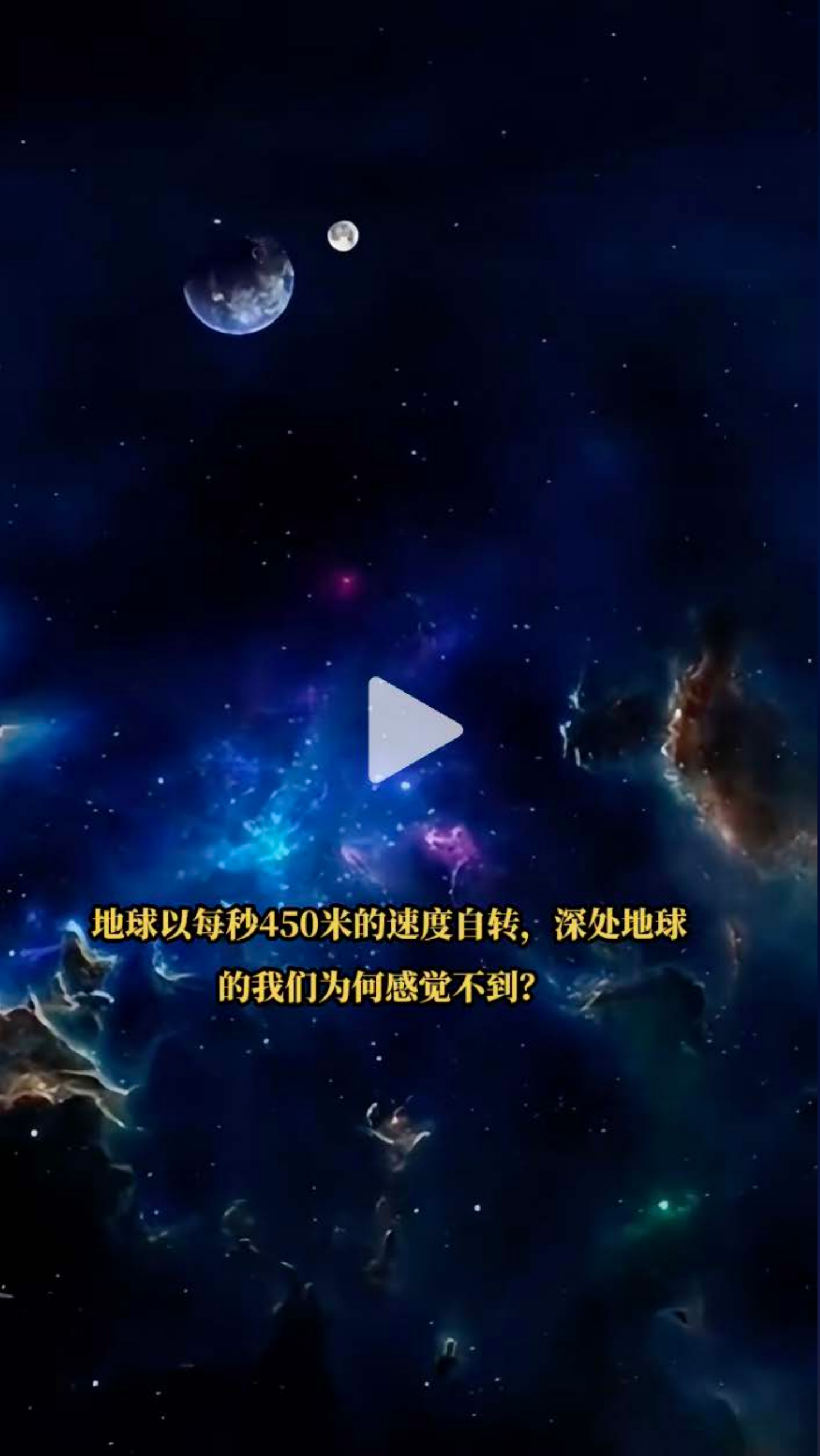}}}&	\multicolumn{2}{c}{\raisebox{-0.22\height}{\includegraphics[width=80pt]{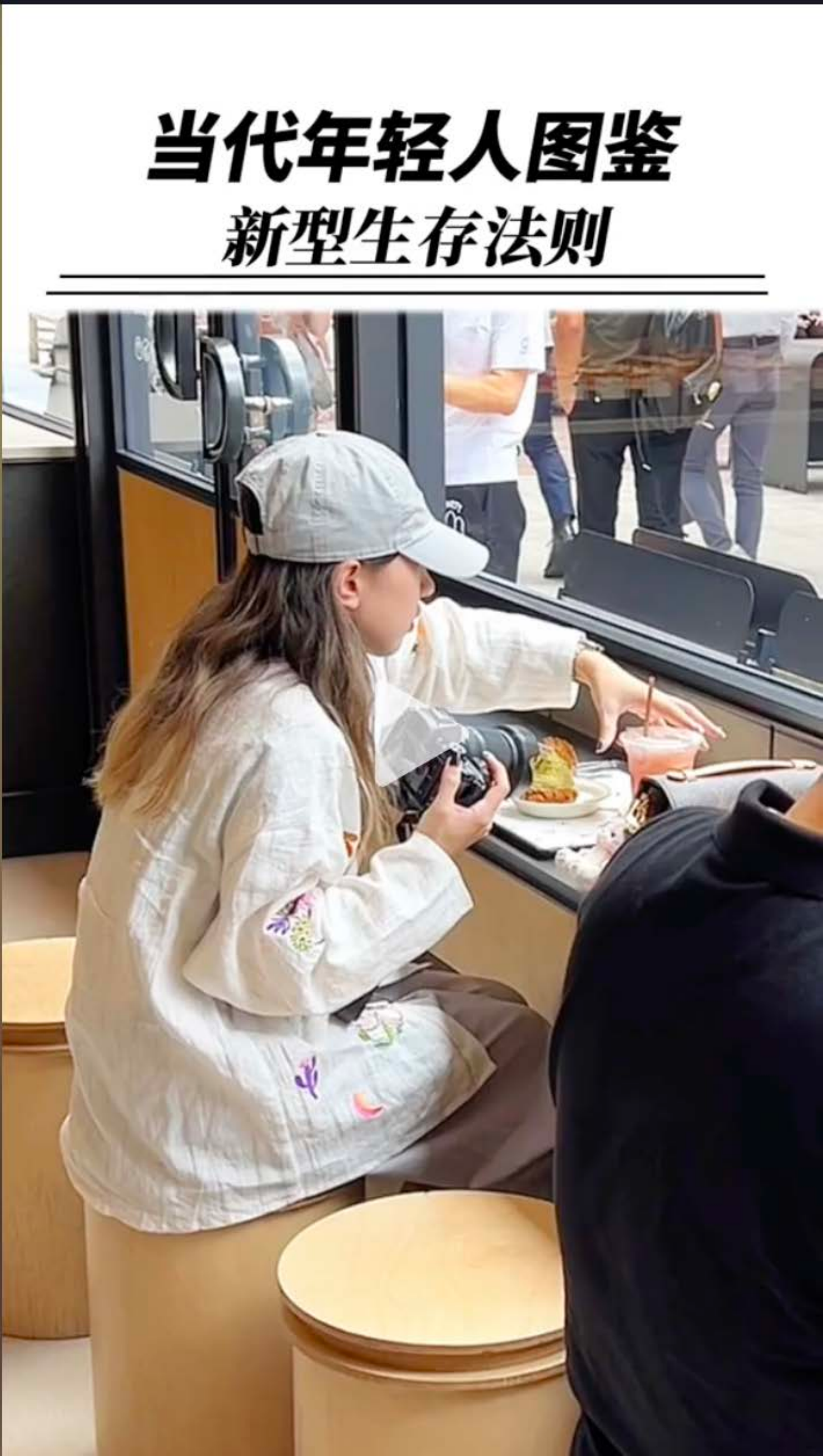}}}\\ \midrule
Theme   & 宠物                                                         & Pets                                                                                                                                                                           & 旅游                                                                   & Travel  
& 科技                                                                   &Technology  
& 时尚                                                                   & Fashion  
\\ \midrule
Title   & 感谢大学生！！！！就是怎么好像是在被菠萝包遛啊\#德牧 \#金毛 \#大学生在抖音遛到狗了 & Thanks to college students! It feels like the dog is taking us for a walk, just like the "Pineapple Bun". \#GermanShepherd \#GoldenRetriever \#StudentsWalkingDogOnTikTok & 8848米，珠穆朗玛峰登顶视角，最接近天空的地方，神奇的大自然\#场面壮观 \#换个角度看世界 \#极限运动 \#惊险刺激 & 8848 meters, the view from the top of Mount Everest, the closest place to the sky, the magic of nature. \#MagnificentView \#DifferentPerspective \#ExtremeSports \#Thrilling & 地球以每秒450米的速度自转，深处地球的我们为何感觉不到？知道的评论区… \#探索宇宙未解之谜 & The Earth rotates at a speed of 450 meters per second. Why can't we, who are deep within the Earth, feel it? Comment if you know... & 这样的当代年轻人新型生存法则，你们觉得怎么样呢？\#消费降级 \#年轻人 \#上海                                & What do you think of the new survival rules for modern young people? \#ConsumerDowngrade \#YoungPeople \#Shanghai                                                                                                                             \\ \midrule
\texttt{ComHeat} & 大学生遛狗，遛出了孙悟空遛猴子的感觉！                           & College students walking the dog, feels like Sun Wukong walking the monkey!                                                                                               & 珠穆朗玛峰，一座高耸入云的巨人，吸引着人们去攀登和探索。                                  & Mount Everest, a giant towering into the clouds, attracts people to climb and explore.                                                                                       & 我以每小时16公里的速度开车，为什么我感觉不到                           & I drive at a speed of 16 kilometers per hour, why can't I feel it?                                                                  & 在消费降级的背景下，上海年轻人在降低消费的同时保持生活质量，他们运用智慧和创意找到了享受生活的方法。这场智慧的探险让人敬佩，也给我们带来了启示。 & In the context of consumption downgrade, young people in Shanghai maintain their quality of life while reducing consumption. They use wisdom and creativity to find ways to enjoy life. This journey of wisdom is admirable and enlightening. \\\midrule
Human   & 大学生遛完回来就这样 打都打不醒                              & After the college students walk, they can't be woken up, even by hitting.                                                                      & 哇，地球好小                                                        & Wow, the Earth is so small.                                                                                                                           & 我感觉不到 怪哉                                          & {I can't feel it. How strange.}                                                                                & 上海精致年轻人会喝瑞幸                                                              & Young refined Shanghai folks like to drink Luckin Coffee.                                                                                                                                                              \\ \bottomrule
\end{tabular}
}
\caption{\small Some cold comments generated by the proposed \texttt{ComHeat}.}
\label{tab:error}
\end{table*}
\end{CJK}

\begin{CJK}{UTF8}{gkai}
\begin{table*}[htbp]
\centering
\resizebox{\textwidth}{!}{%
\begin{tabular} {p{2cm}|p{2.5cm}p{2.5cm}|p{2.5cm}p{2.5cm}|p{2.5cm}p{2.5cm}}
\toprule
Video&\multicolumn{2}{c|}{\raisebox{-0.22\height}{\includegraphics[width=80pt]{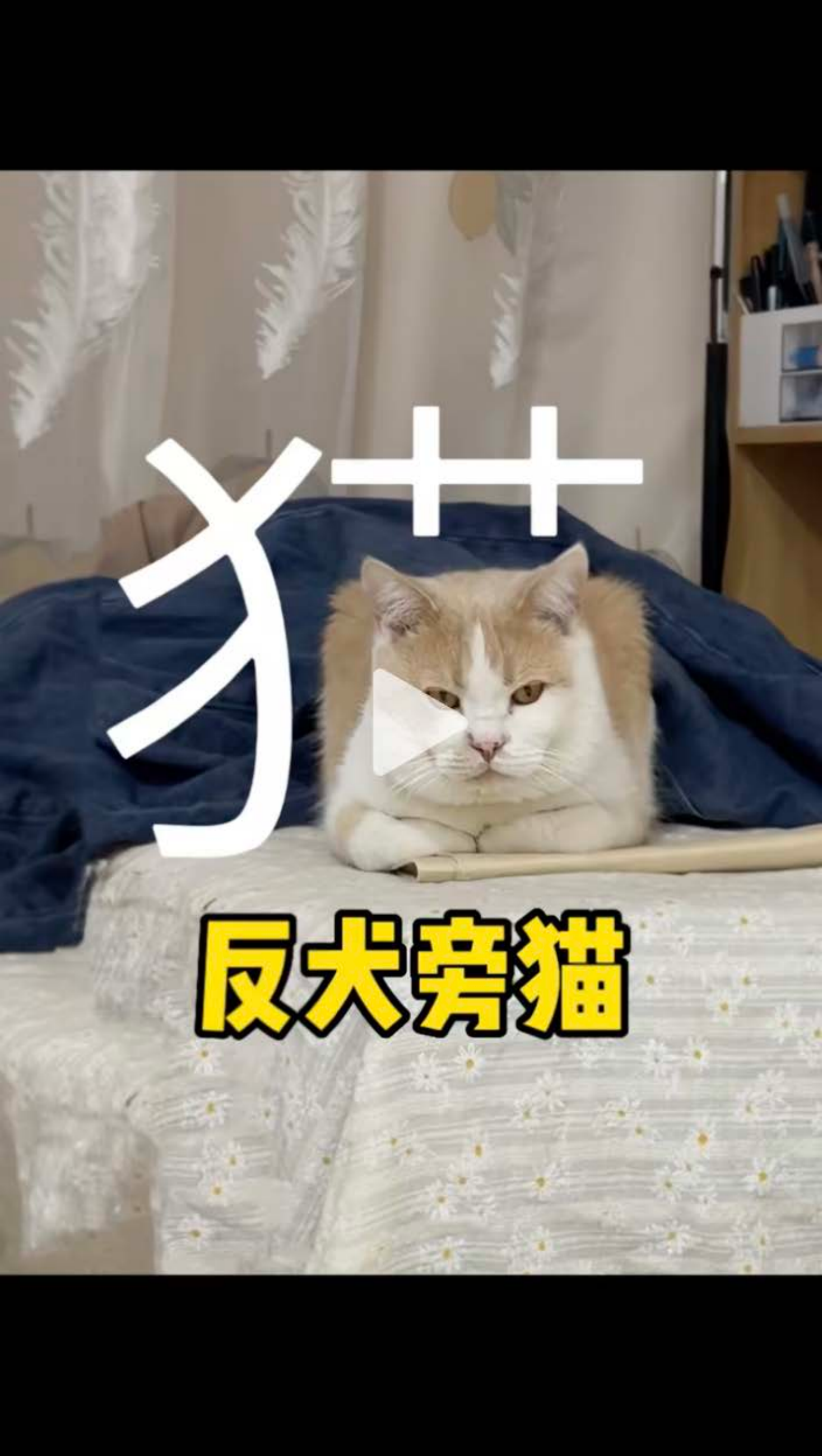}}}	&\multicolumn{2}{c|}{\raisebox{-0.22\height}{\includegraphics[width=80pt]{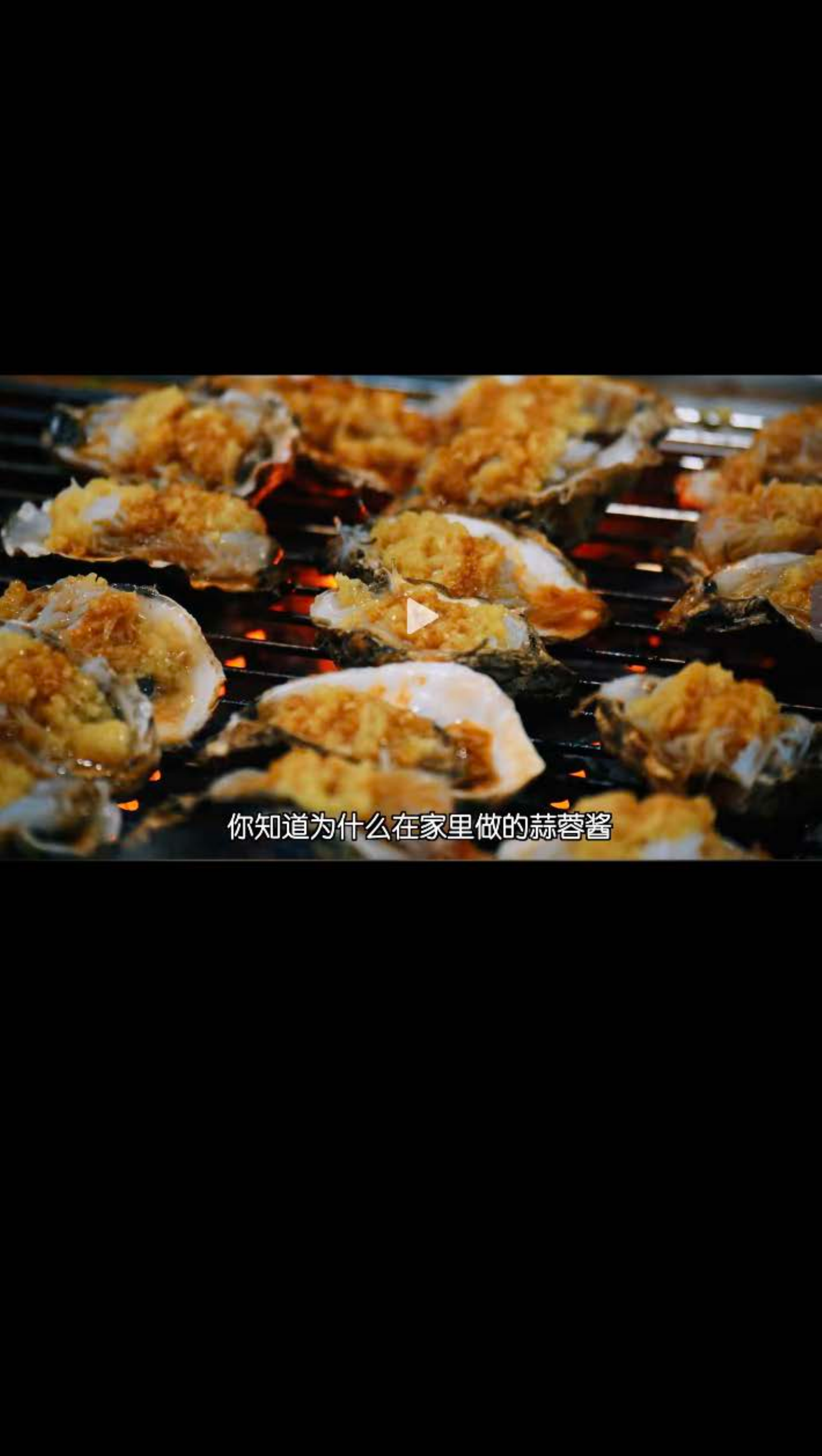}}}	&\multicolumn{2}{c}{\raisebox{-0.22\height}{\includegraphics[width=80pt]{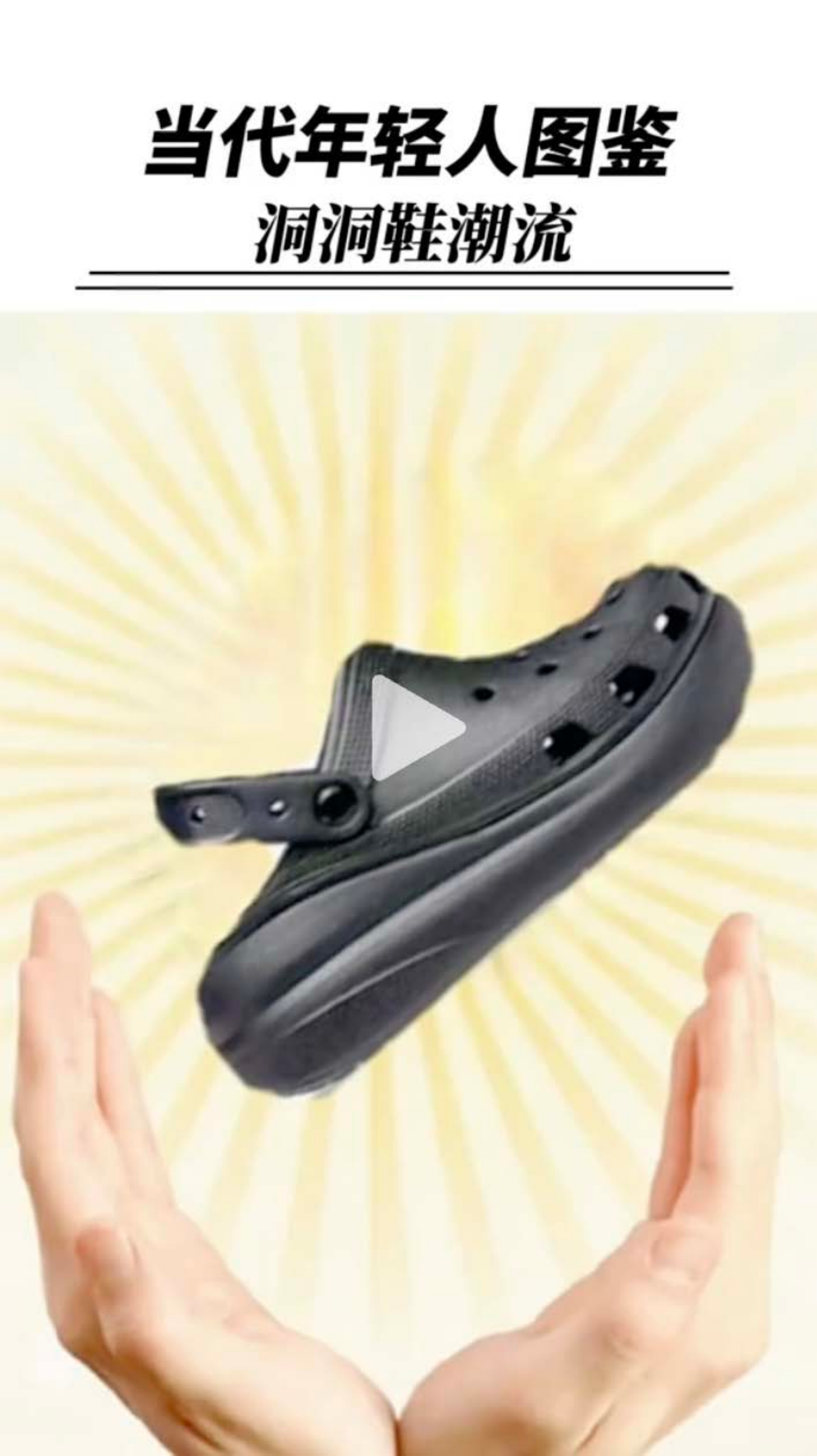}}}\\ \midrule
Theme   & 宠物                                                         & Pets                                                                                                                                                                           & 美食                                                                   & Food  
& 时尚                                                                   & Fashion  
\\ \midrule
Title   & 我的评价是你们的“猫”字都没写对！\#记录猫咪日常 \#快来补全反犬旁 & My comment is that none of you wrote the character "cat" correctly! \#DocumentingCatDailyLife \#ComeAndCorrectTheDogPart & 一个视频教你学会烧烤摊的《蒜蓉酱》到底为什么那么香！\#抖音美食推荐官 \#美食vlog & A video teaches you why the barbecue stall's "garlic sauce" is so fragrant! \#TikTokFoodRecommendation \#FoodVlog & 这样的时髦新门派，你觉得怎么样？\#洞洞鞋 \#洞门 \#年轻人 & What do you think of this trendy new faction? \#HoleyShoes \#HoleGate \#YoungPeople \\ \midrule
\texttt{ComHeat} & 猫：我不是，我没有，别瞎说                       & Cat: I'm not, I don't have, don't talk nonsense.                                                                         & 我怀疑你是在教我做菜                                   & I suspect you're teaching me how to cook.                                                                         & 洞门永存，洞洞鞋才是潮流！                    & The hole door is eternal, holey shoes are the trend!                                \\\midrule
Human   & 所以，师傅姓田？                            & So, is the master's surname Tian?                                                                                        & 嘻嘻我是烧烤摊主…..因为放了超多的鸡精味精鸡粉蚝                    & Hehe, I'm the barbecue stall owner... because I added a lot of chicken essence, MSG, and oyster sauce.            & 这个鞋有gap hole，很适合citywalk         & These shoes have a gap hole, very suitable for citywalk.                            \\ \bottomrule
\end{tabular}
}
\caption{\small Some hot-comments generated by the proposed \texttt{ComHeat} with high score in popular metrics but low score in general metrics.}
\label{tab:humanhigh}
\end{table*}
\end{CJK}

\begin{CJK}{UTF8}{gkai}
\begin{table*}[htbp]
\centering
\resizebox{\textwidth}{!}{%
\begin{tabular} {p{2cm}|p{3cm}p{5cm}|p{4cm}p{4cm}}
\toprule
Video&\multicolumn{2}{c|}{\raisebox{-0.22\height}{\includegraphics[width=80pt]{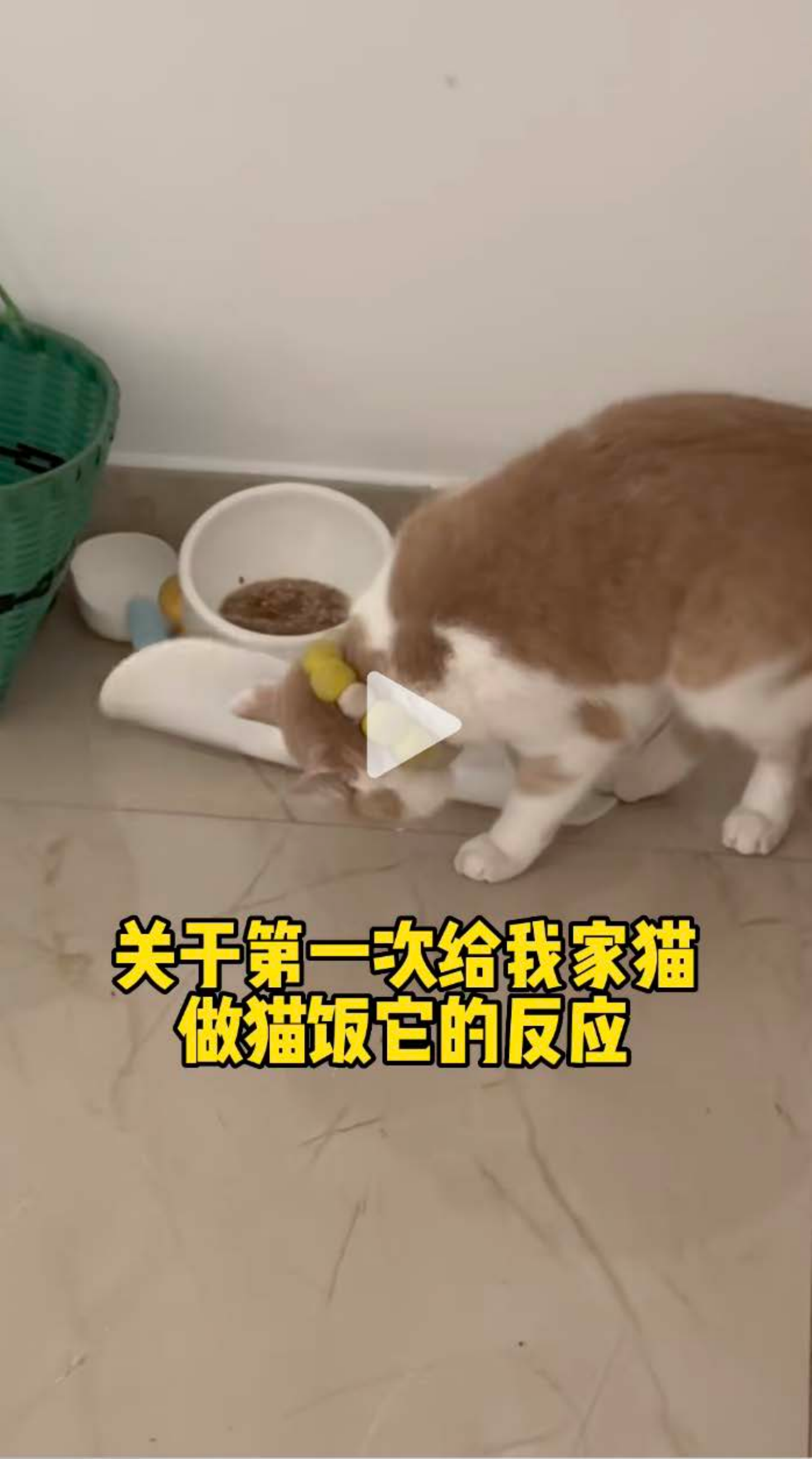}}}	&\multicolumn{2}{c}{\raisebox{-0.22\height}{\includegraphics[width=80pt]{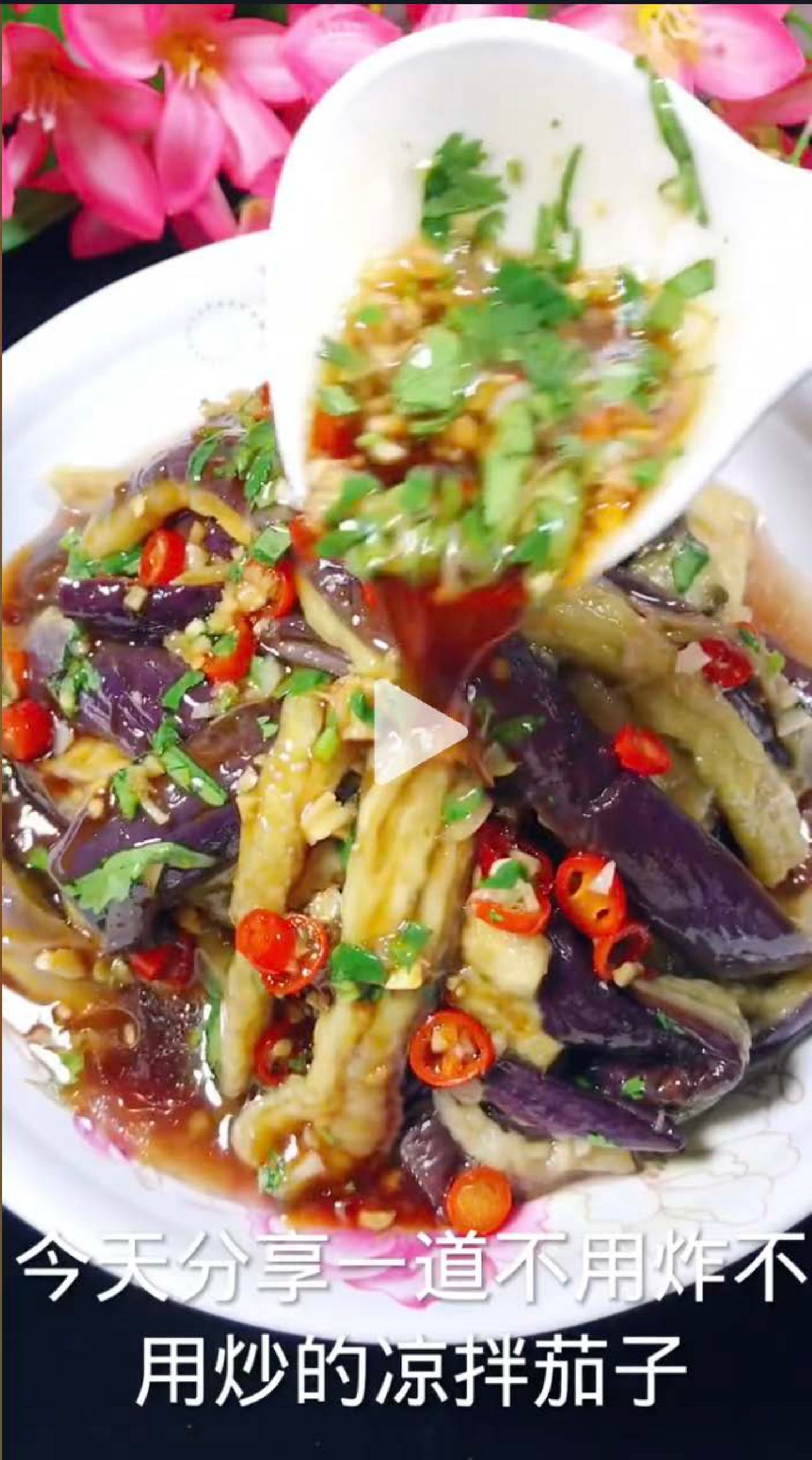}}}\\ \midrule
Theme               & 宠物                                                         & Pets                                                                                                                                                                           & 美食                                                                   & Food                                                                                                                                       \\ \midrule
Title               & 关于第一次给我家猫做猫饭，它的反应！ \#记录猫咪日常 \#方师傅 \#网易天成猫粮 \#兔肉猫粮 \#网易严选猫粮 & About the first time I made cat food for my cat, its reaction! \#RecordingCatDailyLife \#MasterFang \#NetEaseTianchengCatFood \#RabbitCatFood \#NetEaseStrictlySelectedCatFood & 茄子买回家不要炒着吃拉，像我这样做一道凉拌茄子，香辣开胃，好吃又简单\#凉拌茄子 \#下饭菜                       & Don't fry the eggplants you buy. Make a cold dish like I do, spicy and appetizing, delicious and simple \#ColdEggplantDish \#RiceCompanion \\ \midrule
Unified Transformer & 关于第一次给我家猫做猫饭，它的反应什么意思呀？                                    & What does my cat's reaction mean when I made cat food for it for the first time?                                                                                               & 凉拌茄子，香辣开胃，好吃又简单                                                      & Cold mixed eggplant, spicy and appetizing, delicious and simple.                                                                           \\ \midrule
MML-CG              & 关于第一次给我家猫做猫饭，它的反应！                                         & About the first time I made cat food for my cat, its reaction!                                                                                                                 & 很多人都不知道茄子还能这样做。不用煎也不用炸。好吃又开胃。                                        & Many people don't know eggplant can be prepared this way. No need to fry. Delicious and appetizing.                                        \\ \midrule
KLVCG               & 第一次给猫做猫饭，它们的反应是怎样的？                                        & How did the cats react the first time I made cat food for them?                                                                                                                & 茄子买回家不要炒着吃拉，像我这样做一道凉拌茄子，香辣开胃，好吃又简单                                   & Don't just fry the eggplant when you bring it home. Make it like I do: a cold mixed eggplant, spicy and appetizing, delicious and simple.  \\ \midrule
baichuan2           & 方师傅家的猫咪也太挑食了吧，哈哈！                                          & Master Fang's cat is really picky, haha!                                                                                                                                       & 我不喜欢吃茄子，怎么办？                                                         & I don't like eating eggplant, what should I do?                                                                                            \\ \midrule
baichuan2sft        & 方师傅太棒了！我家猫也超爱吃冻干兔肉猫粮，毛色都变得更亮啦～                             & Master Fang is amazing! My cat also loves freeze-dried rabbit cat food, its fur has become shinier$\sim$                                                                       & 凉拌茄子也太好吃了吧！                                                          & Cold mixed eggplant is so delicious!                                                                                                       \\ \midrule
baichuan2sft-rl     & 猫咪：我还是喜欢我的专属猫粮，谢谢你的好意！                                     & Cat: I still prefer my special cat food, thank you for the kind gesture!                                                                                                       & 跟着视频学做凉拌茄子，酸辣开胃，夏日必备！                                                & Follow the video to learn how to make cold mixed eggplant, sour and spicy, a summer essential!                                             \\ \midrule
\texttt{ComHeat}             & 猫：你自己尝尝啥味。。我吃饱了，你吃吧                                        & Cat: You taste it yourself... I'm full, you eat it.                                                                                                                            & 凉拌茄子也太好吃了吧！在家尝试了一下，味道真的很棒！这道菜一定要试试，简单又美味，绝对下饭！茄子爱好者们快试试看吧！ & The cold eggplant dish is really delicious! I tried it at home and the taste is amazing!                                                   \\ \midrule
Human               & 色香味意形 你至少得占一个吧                                             & Appearance, aroma, flavor, and texture - you should at least excel in one.                                                                                                     & 为啥不是大绿茄子                                                             & Why isn't it a large green eggplant?                                                                                                       \\ \bottomrule
\end{tabular}
}
\caption{\small Some hot-comments derived from several themes generated by the proposed \texttt{ComHeat} in comparison to other baselines of Chinese video comment generation.}
\label{tab:case-more1}
\end{table*}
\end{CJK}

\begin{CJK}{UTF8}{gkai}
\begin{table*}[htbp]
\centering
\resizebox{\textwidth}{!}{%
\begin{tabular} {p{2cm}|p{3cm}p{5cm}|p{4cm}p{4cm}}
\toprule
Video&\multicolumn{2}{c|}{\raisebox{-0.22\height}{\includegraphics[width=80pt]{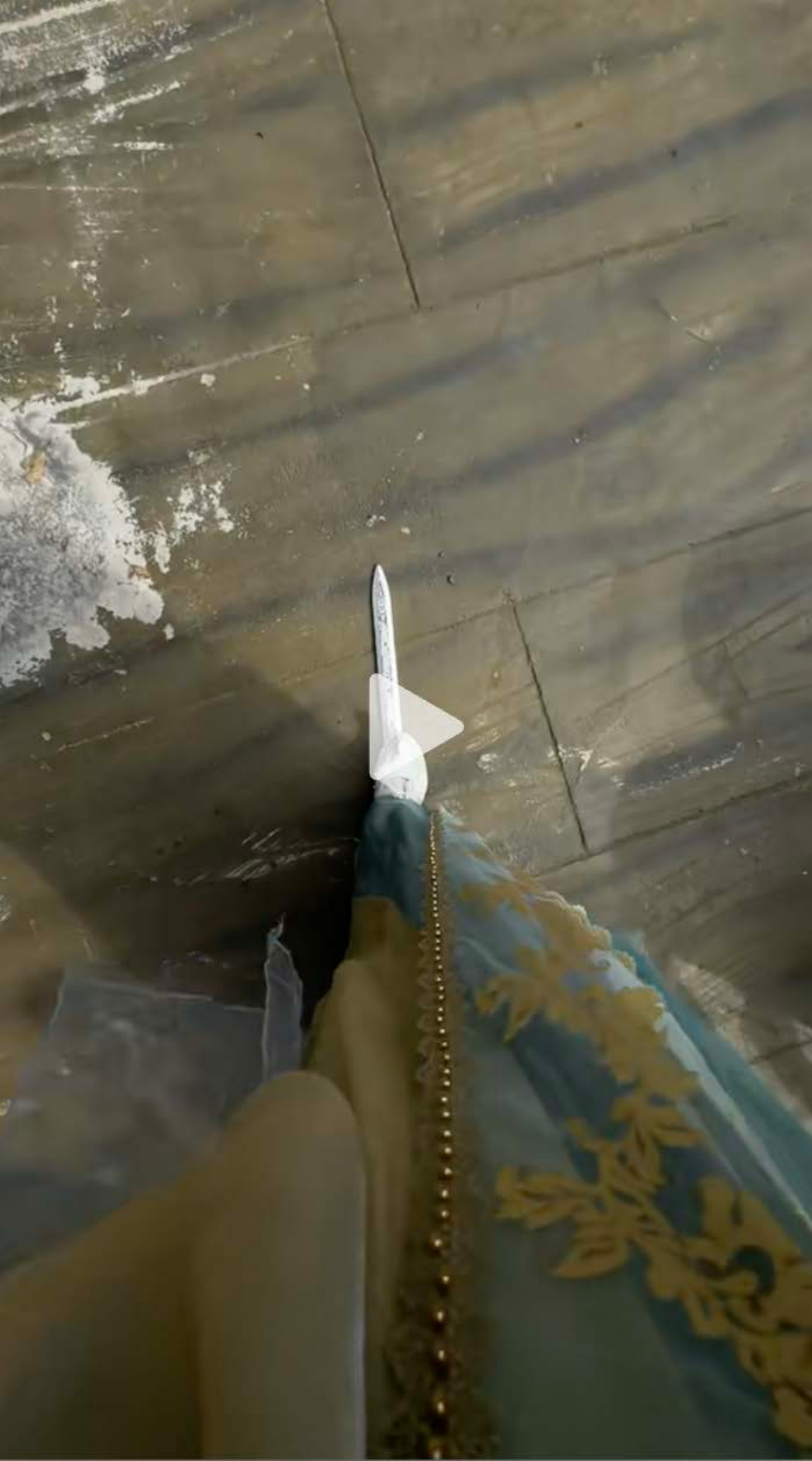}}}	&\multicolumn{2}{c}{\raisebox{-0.22\height}{\includegraphics[width=80pt]{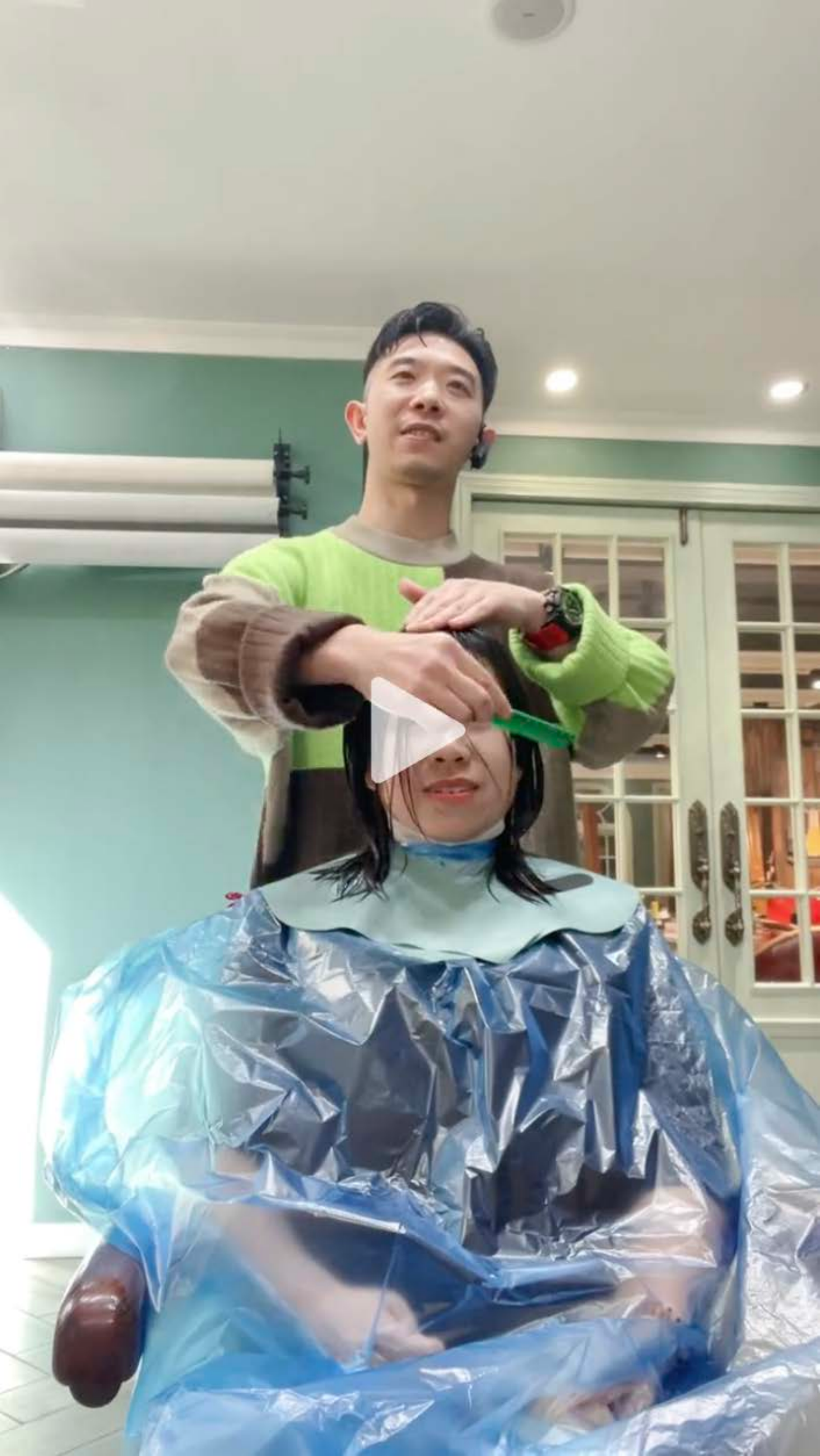}}}\\ \midrule
Theme               & 旅游                                                                        & Travel                                                                                                                                                                                        & 时尚                                 & Fashion                                                                                                    \\ \midrule
Title               & "走啊少年，去见一见命运里的风\#去见一见命运里的风 \#光雾山 \#毕业旅行来四川耍就对了 \#避暑胜地 \#持高考准考证可免费游玩哪些景点", & "Let's go, young man, to feel the wind in fate \#ToFeelTheWindInFate \#GuangwuMountain \#ComeToSichuanForGraduationTrip \#SummerRetreat \#FreeAttractionsWithHighSchoolEntranceExamAdmitCard" & 我这头剪完了就跟换个头似的，阿林这老小子今天整的挺好\#记录真实生活 & After I got this haircut, I felt like I had a new head. Ah Lin did a great job today. \#RecordRealLife     \\ \midrule
Unified Transformer & 毕业旅行来四川耍就对了                                                               & Taking a graduation trip to Sichuan is the right choice.                                                                                                                                      & 这头剪完了就跟换个头似的，阿林这老小子今天整的挺好。         & After the haircut, it feels like a whole new head, Alin did a great job today.                             \\ \midrule
MML-CG              & 持高考准考证可免费游玩哪些景点                                                           & Which attractions can be visited for free with the college entrance examination admission ticket?                                                                                             & 阿林这老小子今天整的挺好                       & Alin did a great job today.                                                                                \\ \midrule
KLVCG               & 走啊，少年去见一见命运里的风。                                                           & Let's go, young man, to feel the wind in your destiny.                                                                                                                                        & 剪头发的老小子，你这头剪的真好，我这头剪完就跟换个头似的，哈哈哈哈。 & You did a great job with the haircut, it feels like a whole new head after getting it done, hahaha.        \\ \midrule
baichuan2           & 去见一见命运里的风，感受大自然的美好，让心灵得到净化。                                               & Go feel the wind of destiny, experience the beauty of nature, and purify your soul.                                                                                                           & 剪完头发就像换了个人，阿林的手艺真是太棒了！             & It feels like a different person after the haircut, Alin's craftsmanship is truly amazing!                 \\ \midrule
baichuan2sft        & 光雾山门票多少钱                                                                  & How much is the ticket for Guangwu Mountain?                                                                                                                                                  & 我怀疑你是故意的                           & I suspect you did it on purpose.                                                                           \\ \midrule
baichuan2sft-rl     & 四川的风景真美，有机会一定要去看看。                                                        & The scenery in Sichuan is truly beautiful; I must visit if I get the chance.                                                                                                                  & 这家理发店真是神奇，剪完头发就像换了张脸，阿林今天的造型真是太棒了！ & This barbershop is magical, it feels like a new face after the haircut, Alin's styling today is fantastic! \\ \midrule
\texttt{ComHeat}             & 人生得意须尽欢，莫使金樽空对月。走啊，少年，去见一见命运里的风！                                          & One should make the most of the joys in life and not let the golden goblet face the moon in vain. Go on, young man, and experience the winds of fate!                                         & 剪刀手爱德华！托尼老师阿林这老小子今天整的挺好            & Edward Scissorhands! Teacher Tony and Ah Lin did a great job today.                                        \\ \midrule
Human               & 南充怎么去呀，导航都没坐车的路线                                                          & How do I get to Nanchong? The navigation doesn't show any driving route.                                                                                                                      & 不错，托尼老师技术不错，小迟年轻帅气十足               & Not bad, Teacher Tony's skills are good, and Xiao Chi is young and handsome.                               \\ \bottomrule
\end{tabular}}
\caption{\small Some hot-comments derived from several themes generated by the proposed \texttt{ComHeat} in comparison to other baselines of Chinese video comment generation.}
\label{tab:case-more2}
\end{table*}
\end{CJK}

\begin{CJK}{UTF8}{gkai}
\begin{table*}[htbp]
\centering
\resizebox{\textwidth}{!}{%
\begin{tabular} {p{2cm}|p{3cm}p{5cm}|p{4cm}p{4cm}}
\toprule
Video&\multicolumn{2}{c|}{\raisebox{-0.22\height}{\includegraphics[width=80pt]{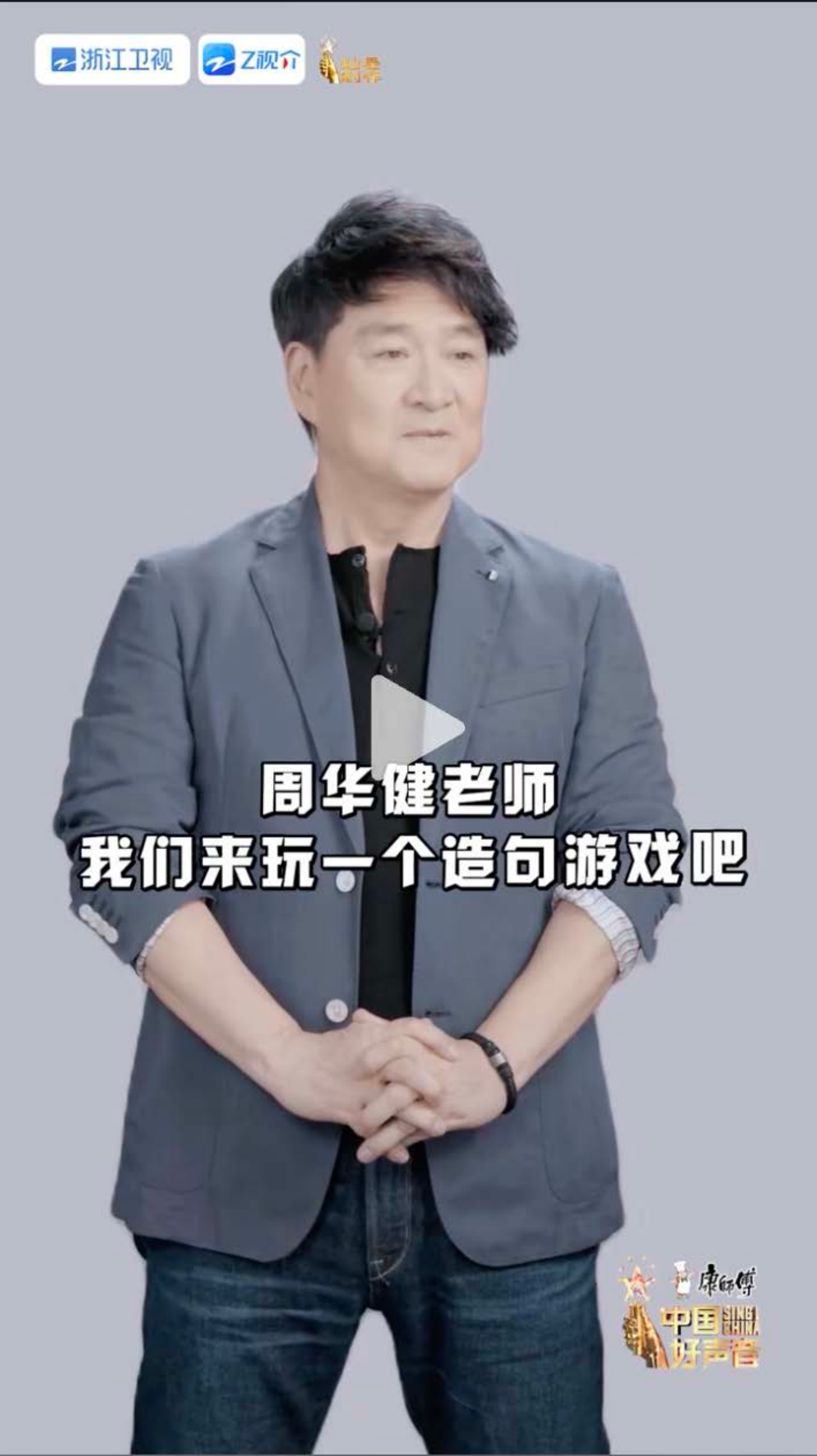}}}	&\multicolumn{2}{c}{\raisebox{-0.22\height}{\includegraphics[width=80pt]{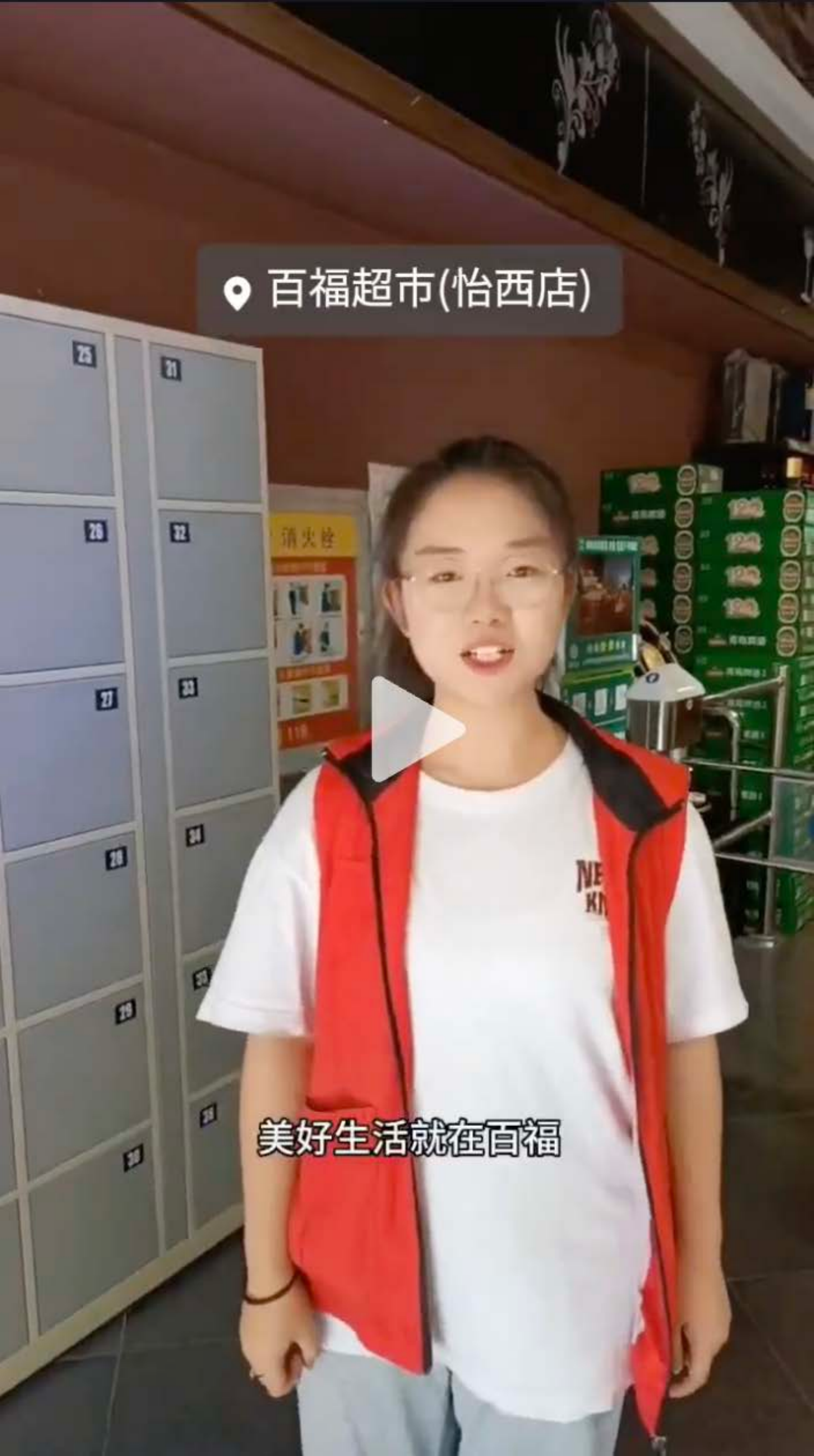}}}\\ \midrule
Theme               & 音乐                                                       & Music                                                                                                                                                                                       & 家居                                                                               & Family                                                                                                                                                                                                                                                            \\ \midrule
Title               & \#2023好声音导师官宣  最后一位导师定了！华语乐坛常青树周华健坐镇！期待7月14日晚八点档一起寻找好声音！ & \#2023TheVoice OfficialAnnouncement The last mentor is decided! The evergreen of Chinese music, Zhou Huajian, is here! Looking forward to 8 PM on July 14th to find the good voice together! & 美好生活就在百福\#家门口的生活超市 \#一站式购物 \#超市 \#购物 @DOU+小助手                                    & A wonderful life is right at BaiFu. \#LocalLifeSupermarket \#OneStopShopping \#Supermarket \#Shopping @DOU+Assistant                                                                                                                                              \\ \midrule
Unified Transformer & 周华健                                                      & Wakin Chau                                                                                                                                                                                  & 美好生活就在百福                                                                         & A wonderful life is at BaiFu.                                                                                                                                                                                                                                     \\ \midrule
MML-CG              & 周华健老师。好。                                                 & Teacher Wakin Chau. Good.                                                                                                                                                                   & 美好生活就在百福，家门口的生活超市                                                                & A wonderful life is at BaiFu, the supermarket right at your doorstep.                                                                                                                                                                                             \\ \midrule
KLVCG               & 周华健老师，我喜欢你的音乐                                            & Teacher Wakin Chau, I like your music.                                                                                                                                                      & 听说开发区管委会对面。开了一家百福精品生活超市。可是百福超市的升级版哦。                                             & I heard across the Development Zone Committee, there's a new BaiFu Boutique Life Supermarket. It's like an upgraded version of BaiFu.                                                                                                                             \\ \midrule
baichuan2           & 期待周华健老师                                                  & Looking forward to Teacher Wakin Chau.                                                                                                                                                      & 我昨天去了一趟，感觉一般般                                                                    & I went there yesterday, felt it was just average.                                                                                                                                                                                                                 \\ \midrule
baichuan2sft        & 周华健老师，我也要报名参加中国好声音！                                      & Teacher Wakin Chau, I also want to sign up for "The Voice of China"!                                                                                                                        & 超市的装修风格真不错，下次去逛逛。                                                                & The supermarket's decor style is really nice, I'll visit again next time.                                                                                                                                                                                         \\ \midrule
baichuan2sft-rl     & 周华健老师，你是我心中的音乐之神，期待你的转椅。                                 & Teacher Wakin Chau, you are the god of music in my heart, looking forward to your chair turn.                                                                                               & 百福精品生活超市，购物者的天堂！新鲜水果、美食、饮料、零食应有尽有，价格还非常良心。6月16号开业还有精彩演出和抽奖活动，快来和家人一起逛逛吧！         & BaiFu Boutique Life Supermarket, a shopper's paradise! Fresh fruits, delicacies, drinks, snacks, all at very reasonable prices. Grand opening on June 16 with amazing performances and lottery events. Come and visit with your family!                           \\ \midrule
\texttt{ComHeat}             & 我是2023中国好声音导师周华健，今年夏天我来寻找和我一样热爱音乐的朋友，是你吗？                & I am Zhou Huajian, a mentor for The Voice of China 2023. This summer, I am looking for friends who share my passion for music. Is it you?                                                   & 百福超市升级，购物者的天堂！食品新鲜便宜，奥利奥1.28，桃汁1.99，纯牛奶3.9，大西瓜9.9……6月16号开业，还有演出和抽奖活动，快来和家人一起逛逛吧！ & BaiFu Supermarket upgraded, a shopper's paradise! Fresh and affordable food: Oreo at 1.28, peach juice at 1.99, pure milk at 3.9, large watermelon at 9.9... Grand opening on June 16th with performances and lottery activities. Come and shop with your family! \\ \midrule
Human               & 应该是好声音历史咖位最高的了                                           & It should be the highest historical star ranking in The Voice.                                                                                                                              & 希望不会有过期食品存在                                                                      & Hope there are no expired products available.                                                                                                                                                                                                                     \\ \bottomrule
\end{tabular}}
\caption{\small Some hot-comments derived from several themes generated by the proposed \texttt{ComHeat} in comparison to other baselines of Chinese video comment generation.}
\label{tab:case-more3}
\end{table*}
\end{CJK}

\begin{CJK}{UTF8}{gkai}
\begin{table*}[htbp]
\centering
\resizebox{\textwidth}{!}{%
\begin{tabular} {p{2cm}|p{3cm}p{5cm}|p{5cm}p{7cm}}
\toprule
Video&\multicolumn{2}{c|}{\raisebox{-0.22\height}{\includegraphics[width=80pt]{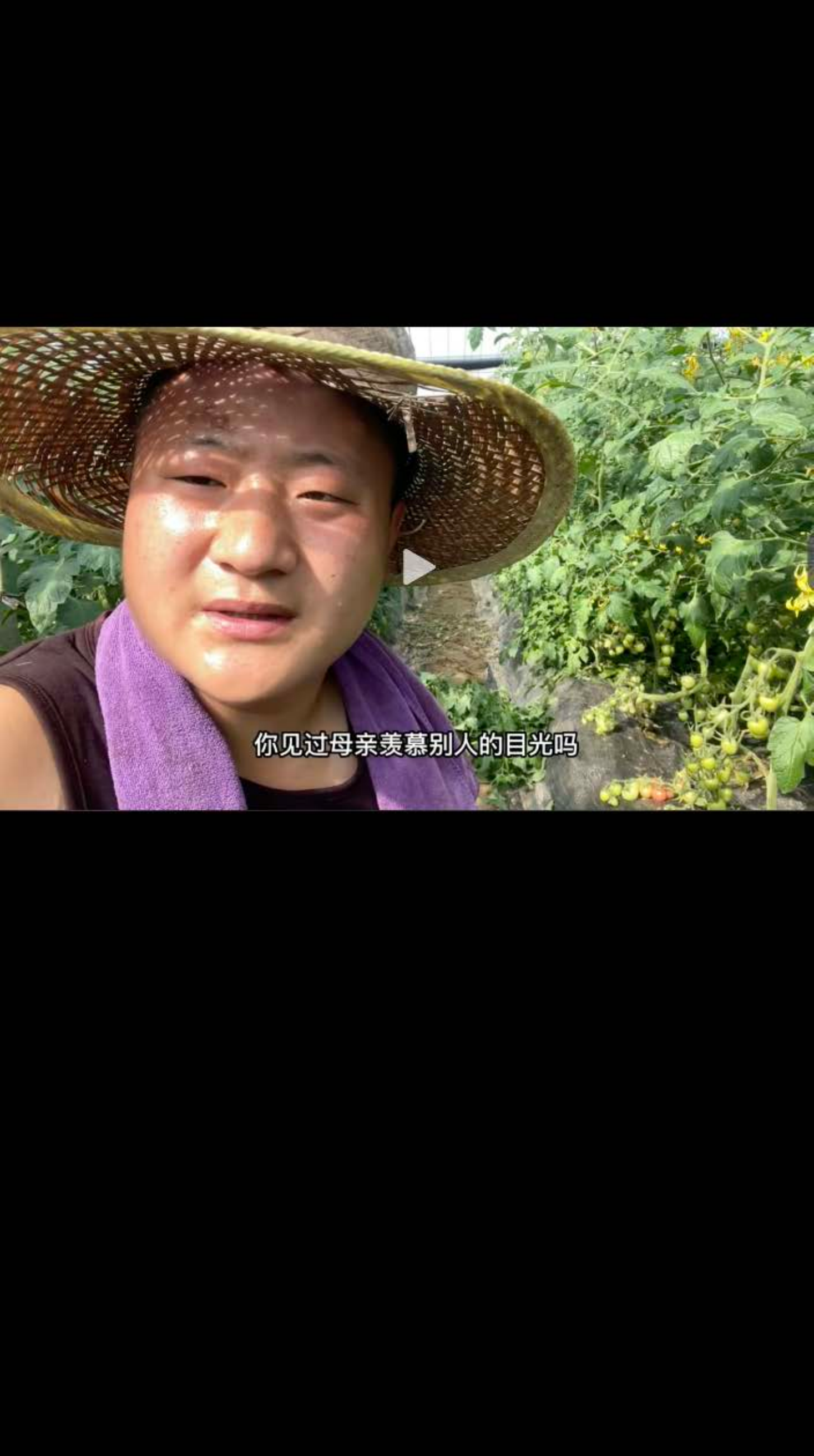}}}&	\multicolumn{2}{c}{\raisebox{-0.22\height}{\includegraphics[width=80pt]{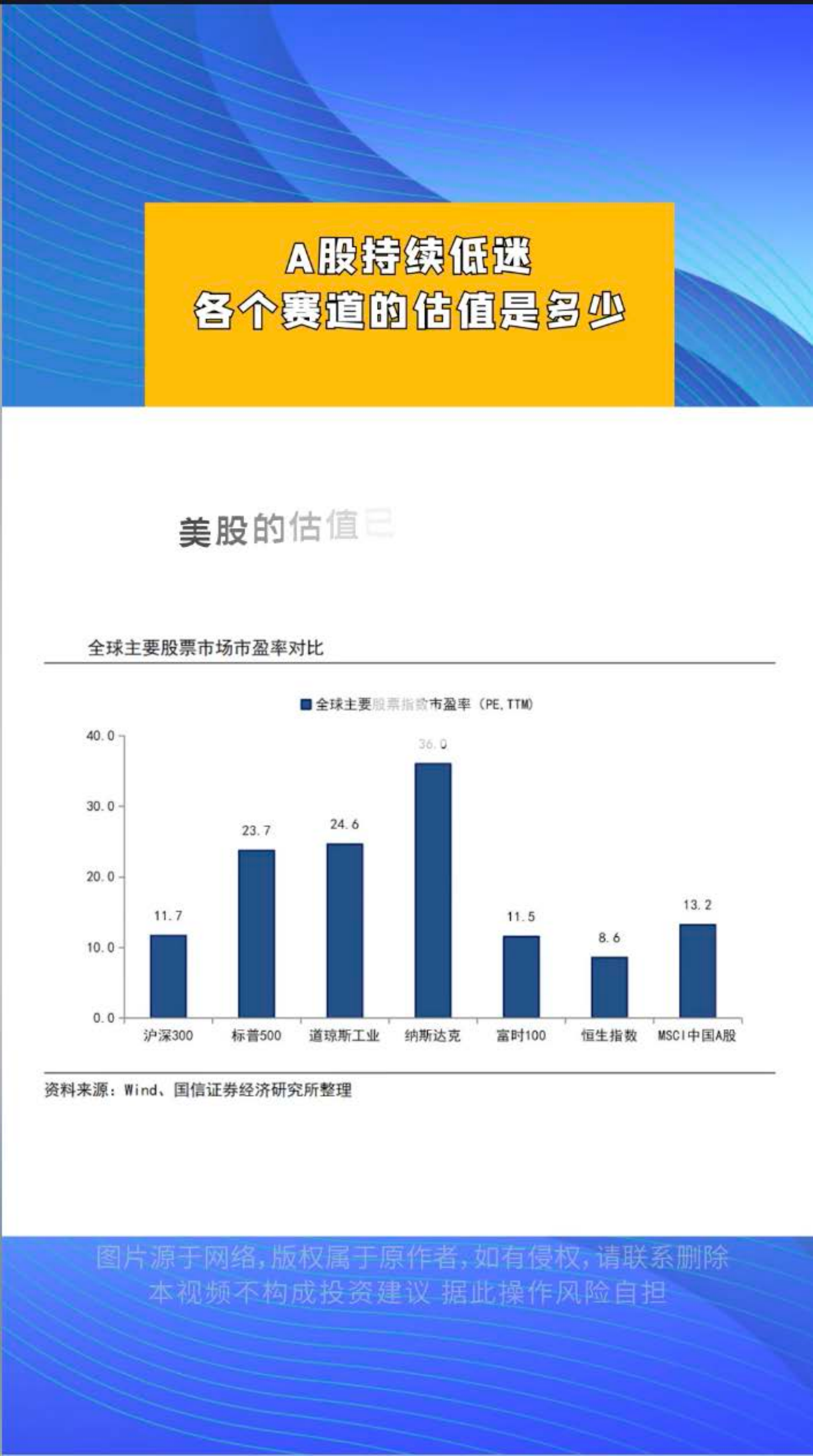}}}\\ \midrule
Theme               & 三农                                                    & Three rural issues                                                                                                                                & 财经                                                                                                  & Finance and economics                                                                                                                                                                                                                                                                                                                                                                                 \\ \midrule
Title               & \#新农人计划2023 \#农村生活 没到收获的季节，硕果累累总是让人感到开心和喜悦@山里人老谭 @小跟班 & \#NewFarmerPlan2023 \#RuralLife Before the harvest season, abundant fruits always bring happiness and joy@MountainManOldTan @LittleFollower       & A股持续低迷，各个赛道的估值是多少？\#股票 \#投资 \#医药 \#贵州茅台 \#中国中免 \#创业板 \#隆基绿能 \#宁德时代 \#药明康德                           & A-shares continue to be sluggish. What are the valuations for each sector? \#Stocks \#Investment \#Pharmacy \#KweichowMoutai \#ChinaDutyFreeGroup \#GEM \#LongiGreenTech \#CATL \#WuXiAppTec                                                                                                                                                                                                          \\ \midrule
Unified Transformer & 硕果累累总是让人感到开心和喜悦                                       & Abundant fruits always bring happiness and joy.                                                                                                   & 新能源光伏半导体                                                                                            & New energy photovoltaic semiconductor                                                                                                                                                                                                                                                                                                                                                                 \\ \midrule
MML-CG              & 农民伯伯们真辛苦，为我们的美食付出很多汗水。                                & The farmer uncles work really hard, sweating a lot for our delicious food.                                                                        & 白酒，医疗，新能源，半导体，军工，光伏，中药，半导体，光伏                                                                       & Liquor, medical, new energy, semiconductor, military industry, photovoltaic, traditional Chinese medicine, semiconductor, photovoltaic                                                                                                                                                                                                                                                                \\ \midrule
KLVCG               & 这个葡萄看起来好甜，期待品尝一下！                                     & These grapes look sweet, looking forward to tasting them!                                                                                         & 美股的估值已经远远高于A股                                                                                       & The valuation of US stocks is far higher than A-shares.                                                                                                                                                                                                                                                                                                                                               \\ \midrule
baichuan2           & 巫山那个酸椒火锅好吃                                            & The pickled pepper hot pot in Wushan is delicious.                                                                                                & 在全球视角下，A股市场估值合理，各大赛道投资价值凸显！                                                                         & From a global perspective, the valuation of the A-share market is reasonable, and the investment value of each major sector stands out!                                                                                                                                                                                                                                                               \\ \midrule
baichuan2sft        & 老谭，你这个瓜好吃吗？                                           & Old Tan, is this melon tasty?                                                                                                                     & A股持续低迷，各个赛道的估值是多少？看来投资需要更深入的了解和分析。                                                                  & A-shares continue to be sluggish. How are the valuations for each sector? It seems that investment requires a deeper understanding and analysis.                                                                                                                                                                                                                                                      \\ \midrule
baichuan2sft-rl     & 这个瓜好吃，我也要去摘。                                          & This melon is delicious, I also want to pick some.                                                                                                & 在A股持续低迷的情况下，各个赛道的估值确实值得关注。从市盈率的角度来看，许多赛道已经回到了历史低位，这或许是一个值得关注的投资时机。不过，投资前请务必做好充分的研究，了解风险并谨慎决策。祝投资顺利！ & 2Given the ongoing slump in A-shares, the valuations of each sector are indeed worth paying attention to. From a price-to-earnings ratio perspective, many sectors have returned to historical lows, which may be an investment opportunity worth noting. However, please make sure to conduct thorough research, understand the risks, and make cautious decisions. Best wishes for your investment! \\ \midrule
\texttt{ComHeat}             & 农人勤劳致富，丰收的喜悦满满！祝你们明年收获更多李子，生活更美满！                     & Farmers work hard to prosper, the joy of harvest is overwhelming! Wishing you to harvest more plums next year, leading to a more fulfilling life! & A股持续低迷，各个赛道的估值是多少？我不知道，市盈率多少有大佬知道么？                                                                 & A-shares continue to be sluggish. What are the valuations for each sector? I don't know, does anyone know the price-to-earnings ratio?                                                                                                                                                                                                                                                                \\ \midrule
Human               & 国人屋里这么多李子不吃，非要吃巫山曲尺的李子也不嫌贵                            & People have so many plums at home but prefer to eat the expensive ones from Wushan.                                                               & 人家的龙头公司市盈率还是很低啊                                                                                     & The leading companies still have a very low price-to-earnings ratio.                                                                                                                                                                                                                                                                                                                                  \\ \bottomrule
\end{tabular}}
\caption{\small Some hot-comments derived from several themes generated by the proposed \texttt{ComHeat} in comparison to other baselines of Chinese video comment generation.}
\label{tab:case-more4}
\end{table*}
\end{CJK}

\begin{CJK}{UTF8}{gkai}
\begin{table*}[htbp]
\centering
\resizebox{\textwidth}{!}{%
\begin{tabular} {p{2cm}|p{3cm}p{5cm}|p{4cm}p{4cm}}
\toprule
Video&\multicolumn{2}{c|}{\raisebox{-0.22\height}{\includegraphics[width=80pt]{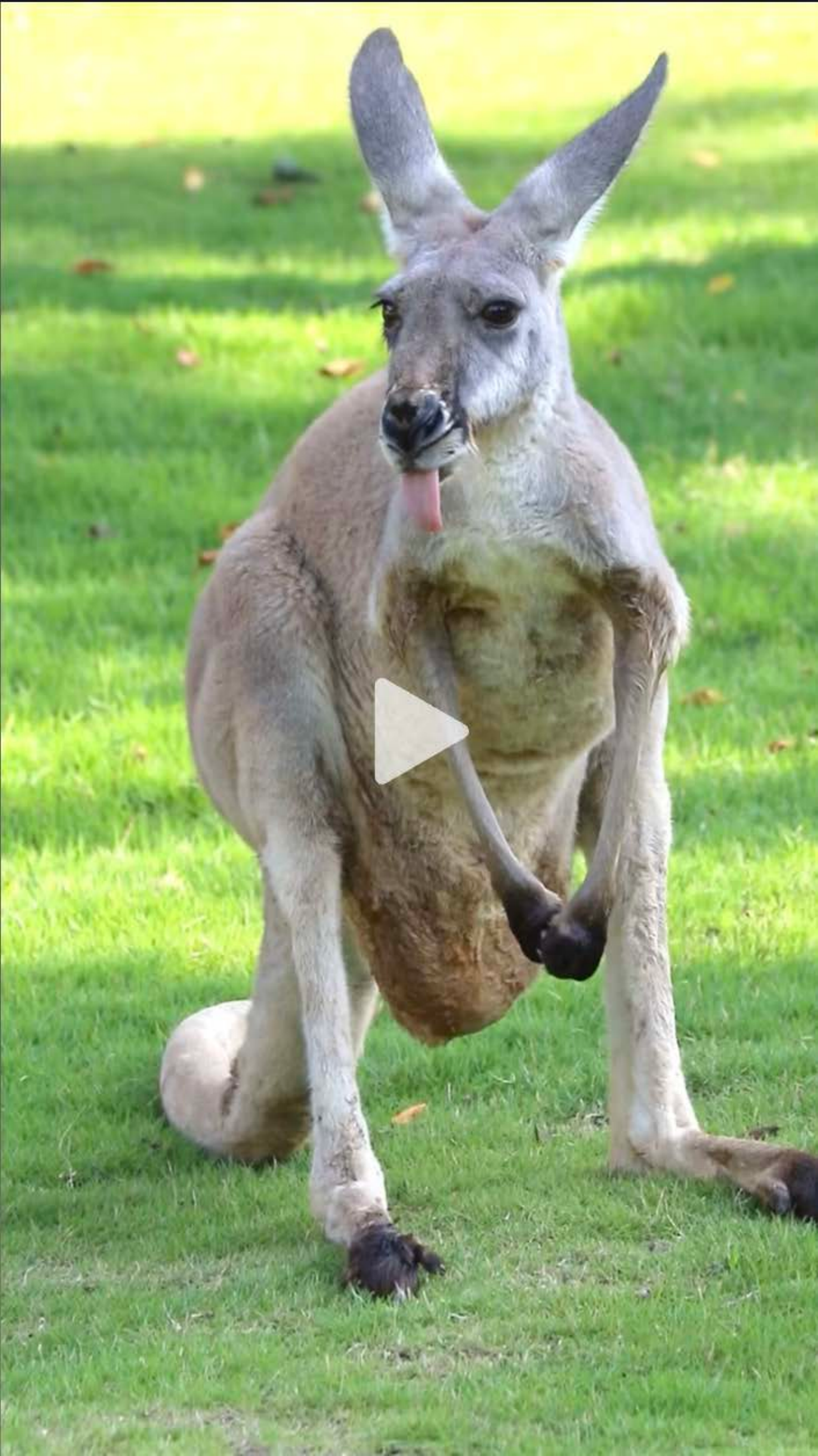}}}&	\multicolumn{2}{c}{\raisebox{-0.22\height}{\includegraphics[width=80pt]{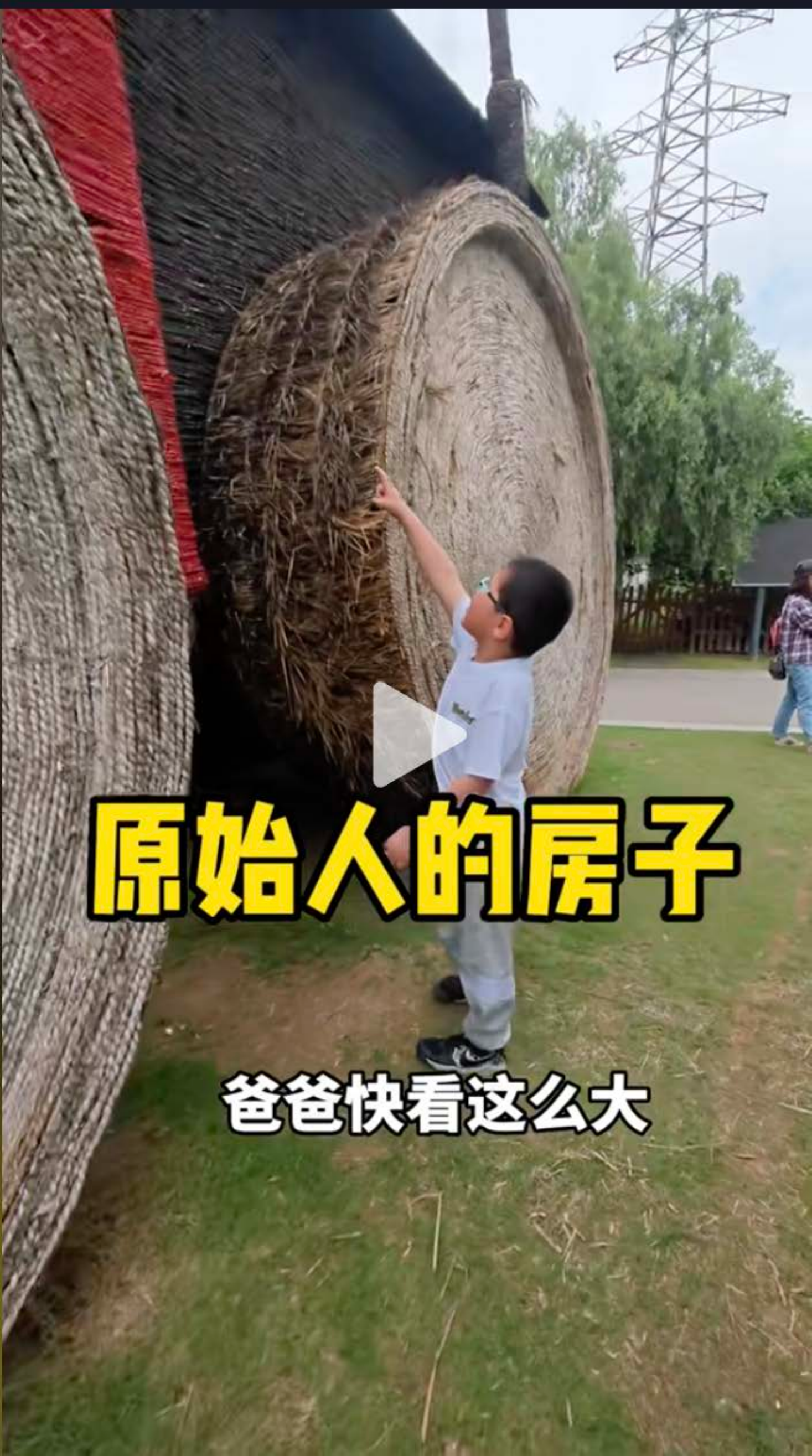}}}\\ \midrule
Theme               & 知识                                                       & Knowledge                                                                                                                                                                  & 亲子                                     & Parent-child                                                                                                                 \\ \midrule
Title               & 它的唾液不仅是清洁剂，更加是降温神器！\#袋鼠 \#动物的迷惑行为                        & Its saliva is not only a cleaning agent, but also a cooling tool! \#Kangaroo \#AnimalsMysteriousBehavior                                                                   & 这难道是原始人的房子\#人类幼崽观察团 \#金牌育言家 \#萌娃搞笑日常   & Is this the house of primitive people? \#HumanCubObservationGroup \#TopParentingExpert \#CuteKidDailyLaughs                  \\ \midrule
Unified Transformer & 它的唾液不仅是清洁剂，更加是降温神器！\#袋鼠 \#动物的迷惑行为                        & Its saliva is not only a cleaning agent, but also a cooling tool! \#Kangaroo \#AnimalsMysteriousBehavior                                                                   & 这难道是原始人的房子吗？                           & Is this the house of primitive people?                                                                                       \\ \midrule
MML-CG              & 动物的迷惑行为                                                  & Animals' puzzling behaviors                                                                                                                                                & 哇，这个原始人的房子真的好有趣！                       & Wow, this primitive person's house is really interesting!                                                                    \\ \midrule
KLVCG               & 袋鼠的唾液真是神奇，                                               & Kangaroo's saliva is truly amazing,                                                                                                                                        & 原始人的房子，我也要去看看。                         & The primitive person's house, I want to see it too.                                                                          \\ \midrule
baichuan2           & 袋鼠的唾液真是神奇，夏天必备降温神器！                                      & Kangaroo's saliva is amazing, a must-have cooling tool for summer!                                                                                                         & 小男孩的探险之旅，仿佛回到了原始时代！期待下一集！              & The little boy's adventure journey, as if he returned to the primitive era! Looking forward to the next episode!             \\ \midrule
baichuan2sft        & 袋鼠的唾液真是神奇，不仅能清洁还能降温！这迷惑行为让我大开眼界了                         & Kangaroo's saliva is amazing, it can not only clean but also cool down! This baffling behavior really opened my eyes                                                       & 哇，宝宝的好奇心都被激发出来了，好想和爸爸妈妈一起去探险！          & Wow, the baby's curiosity has been aroused, I really want to go on an adventure with mom and dad!                            \\ \midrule
baichuan2sft-rl     & 袋鼠的唾液不仅可以用来清洁身体，还可以用来降温，这让我想到了人类的口水，它不仅可以用来清洁口腔，还可以用来降温。 & Kangaroo's saliva can not only be used to clean the body but also to cool down. This reminds me of human saliva, which can not only clean the mouth but also cool it down. & 原始人使用石器工具，小男孩拿着一根棍子，这让我想到了人类进化史上的一个阶段。 & Primitive people used stone tools, the little boy holding a stick, this reminds me of a stage in human evolutionary history. \\ \midrule
\texttt{ComHeat}             & 袋鼠：我吐口水给你降降温                                             & Kangaroo: I'll drool to cool you down                                                                                                                                      & 哇，这个'原始人房子'是不是藏着小帅哥的秘密宝藏？这个孩子也太可爱了！    & Wow, does this 'primitive person's house' hide the secret treasure of the little handsome guy? This kid is too adorable!     \\ \midrule
Human               & 如果我和袋鼠打一场拳击赛，我有没有可能赢                                     & If I were to box with a kangaroo, would I have a chance to win?                                                                                                            & 动物不怕天真可爱的孩子！尤其像我大侄儿这样的！                & Animals are not afraid of innocent and cute children, especially ones like my grandnephew!                                   \\ \bottomrule
\end{tabular}}
\caption{\small Some hot-comments derived from several themes generated by the proposed \texttt{ComHeat} in comparison to other baselines of Chinese video comment generation.}
\label{tab:case-more5}
\end{table*}
\end{CJK}

\begin{CJK}{UTF8}{gkai}
\begin{table*}[htbp]
\centering
\resizebox{\textwidth}{!}{%
\begin{tabular} {p{2cm}|p{3cm}p{5cm}|p{4cm}p{4cm}}
\toprule
Video&\multicolumn{2}{c|}{\raisebox{-0.22\height}{\includegraphics[width=80pt]{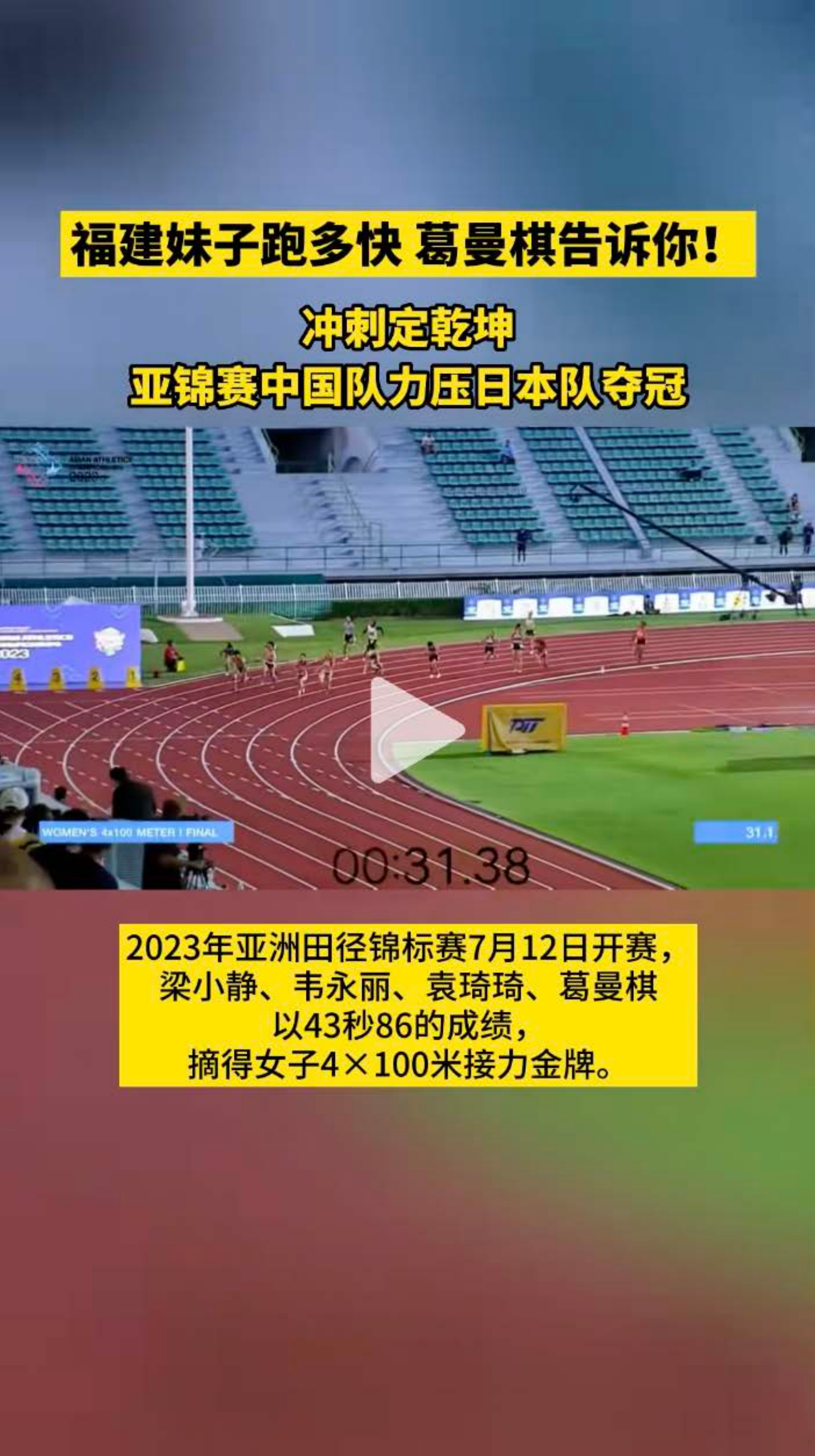}}}	&\multicolumn{2}{c}{\raisebox{-0.22\height}{\includegraphics[width=80pt]{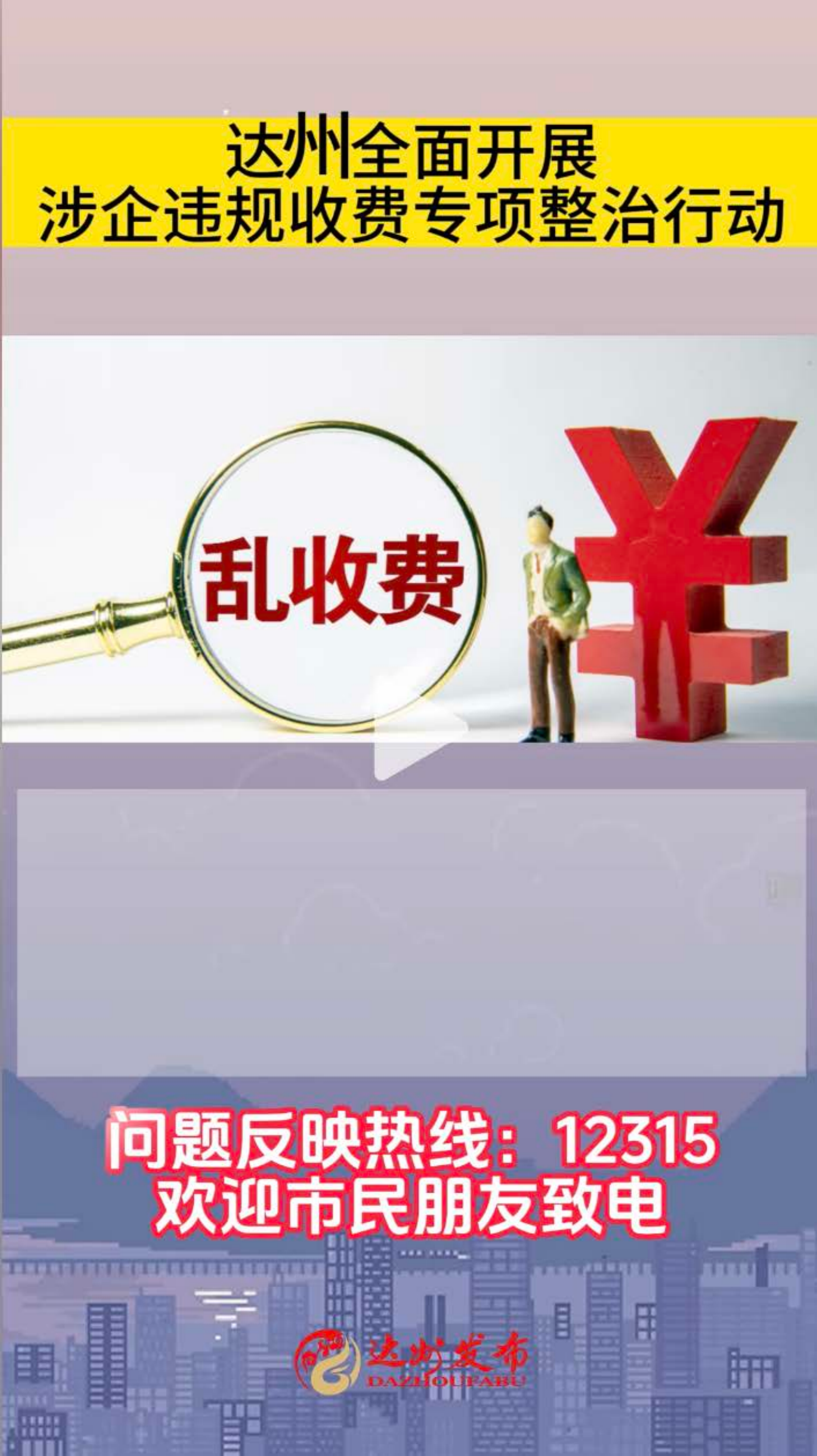}}}\\ \midrule
Theme               & 体育                                               & Sports                                                                                                                                                                     & 时政社会                              & Current Affairs and Society                                                                                                          \\ \midrule
Title               & 祝贺！福建“短跑女飞人” 葛曼棋，助力中国女子接力队4×100米接力夺冠！\#接力赛  \#金牌 & Congratulations! Fujian's "Sprint Flying Woman" Ge Manqi, helps the Chinese women's relay team win the 4x100m relay championship! \#RelayRace \#GoldMedal                  & 达州全面开展涉企违规收费专项整治行动\#达州 \#市场 \#收费  & Dazhou comprehensively launches a special rectification campaign on irregular enterprise-related fees. \#Dazhou \#Market \#Fees      \\ \midrule
Unified Transformer & 福建“短跑女飞人” 葛曼棋，助力中国女子接力队4×100米接力夺冠！               & Fujian's "Sprint Flying Woman" Ge Manqi, helps the Chinese women's relay team win the 4x100m relay championship! \#RelayRace \#GoldMedal                                   & 达州全面开展涉企违规收费专项整治行动，点赞支持！          & Dazhou is comprehensively launching a special rectification campaign on irregular enterprise-related fees. Thumbs up for support!    \\ \midrule
MML-CG              & 葛曼棋，你真棒！                                         & Ge Manqi, you're amazing!                                                                                                                                                  & 涉企违规收费专项整治行动开始了！                  & The special rectification campaign on irregular enterprise-related fees has begun!                                                   \\ \midrule
KLVCG               & 葛曼棋，你的速度真的很快！                                    & Ge Manqi, your speed is really fast!                                                                                                                                       & 为达州的涉企违规收费专项整治行动点赞！               & Thumbs up for Dazhou's special rectification campaign on irregular enterprise-related fees!                                          \\ \midrule
baichuan2           & 我怀疑她是不是福建的                                       & I wonder if she's from Fujian                                                                                                                                              & 这个专项整治行动真是太棒了！为企业减轻负担，促进经济发展。点赞！  & This special rectification campaign is fantastic! It reduces the burden on enterprises and promotes economic development. Thumbs up! \\ \midrule
baichuan2sft        & 葛曼棋，福建的骄傲                                        & Ge Manqi, the pride of Fujian                                                                                                                                              & 让我们一起关注市场经济中的收费行为，共同维护公平竞争的市场环境。  & Let's pay attention to the charging behavior in the market economy and jointly maintain a fair competitive market environment.       \\ \midrule
baichuan2sft-rl     & 葛曼棋的速度和孙悟空的奔跑速度一样快！为中国田径队赢得亚锦赛首枚金牌，真是太棒了！        & Ge Manqi's speed is as fast as Sun Wukong's running speed! Winning the first gold medal for the Chinese track and field team in the Asian Championships is really awesome! & 达州的出租车，打表的少，不打表的多                 & In Dazhou, fewer taxis use the meter, and many don't.                                                                                \\ \midrule
\texttt{ComHeat}             & 葛曼棋，你的速度真是飞一般！为中国女子接力队加油，金牌实至名归！辛苦了！      & Ge Manqi, your speed is truly flying! Cheering for the Chinese women's relay team, the gold medal is well-deserved! Great job!                                             & 我开了一个小超市，每个月要交500元的卫生费，可以帮忙查一下合理吗 & I opened a small supermarket and have to pay 500 yuan in sanitation fees every month. Can you help check if it's reasonable?         \\ \midrule
Human               & 反兴奋剂中心辛苦了                                        & The anti-doping center has worked hard                                                                                                                                     & 还是查一下学校的收费吧，打过12345还不是一样。         & Better check the school fees. Even after calling 12345, it's still the same.                                                         \\ \bottomrule
\end{tabular}}
\caption{\small Some hot-comments derived from several themes generated by the proposed \texttt{ComHeat} in comparison to other baselines of Chinese video comment generation.}
\label{tab:case-more6}
\end{table*}
\end{CJK}

\begin{CJK}{UTF8}{gkai}
\begin{table*}[htbp]
\centering
\resizebox{\textwidth}{!}{%
\begin{tabular} {p{2cm}|p{3cm}p{5cm}|p{3cm}p{6cm}}
\toprule
Video&\multicolumn{2}{c|}{\raisebox{-0.22\height}{\includegraphics[width=80pt]{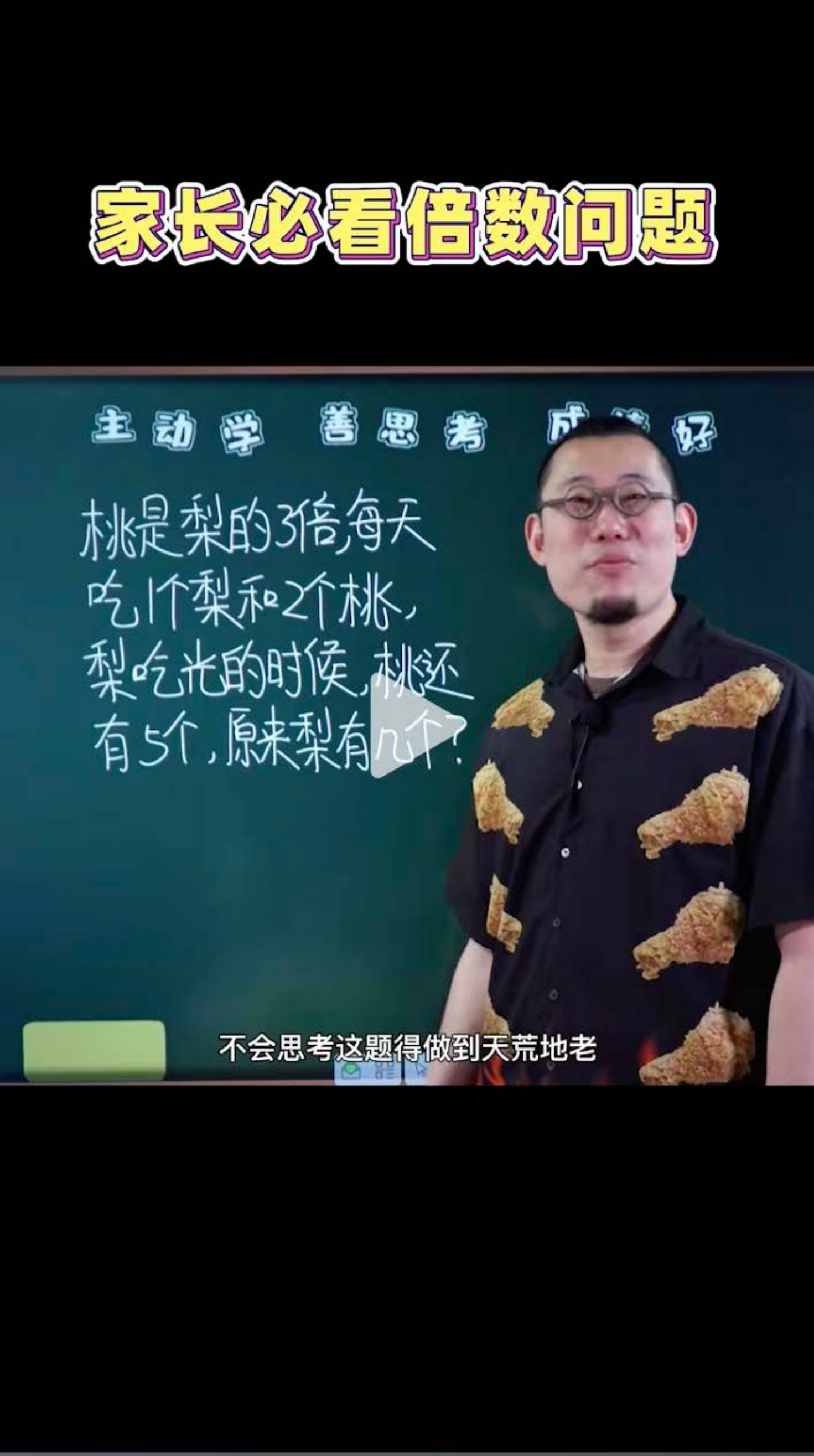}}}	&\multicolumn{2}{c}{\raisebox{-0.22\height}{\includegraphics[width=80pt]{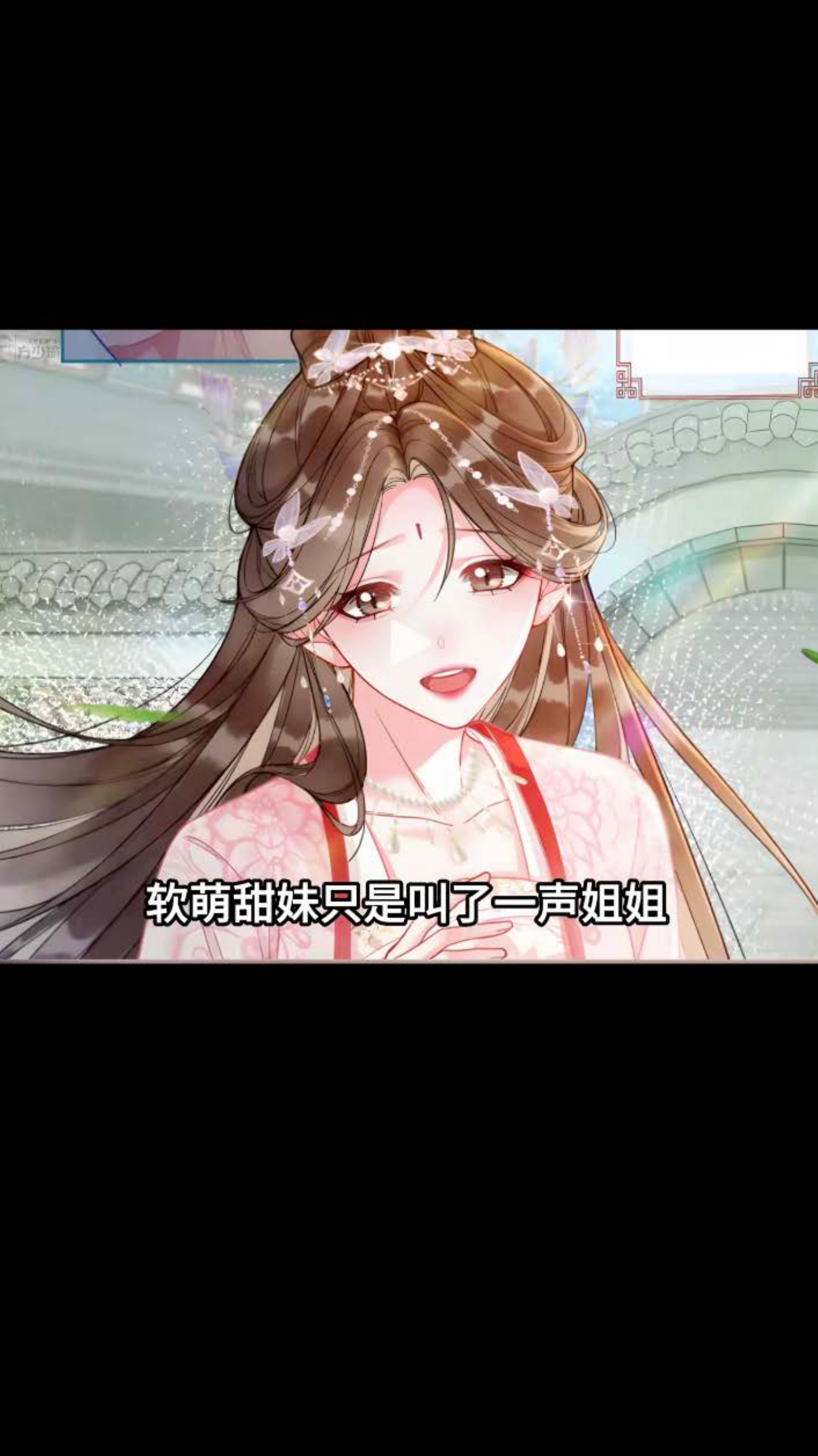}}}\\ \midrule
Theme               & 校园教育                                          & School Education                                                                                                                                                                              & 二次元                                               & 2D World                                                                                                                                                                                            \\ \midrule
Title               & 这题的突破口是找倍数，背后是转化思维，你学会了吗？\#思维训练 \#学习方法 \#知识分享 & The key to this problem is to find the multiple, and the underlying principle is the transformation of thinking. Have you learned it? \#ThinkingTraining \#LearningMethods \#KnowledgeSharing & 1姜幼安教训丫鬟立威，墨扶白动心了\#快看的漫画 \#二次元 \#漫画解说充能计划 \#漫画推荐  & "I Finally Became the Group's Favorite" Jiang You'an teaches the maid Li Wei a lesson, and Mo Fubai is touched. \#QuickReadComics \#2DWorld \#ComicExplanationEmpowermentPlan \#ComicRecommendation \\ \midrule
Unified Transformer & 这题的突破口是找倍数，背后是转化思维，你学会了吗？\#思维训练 \#学习方法 \#知识分享 & The key to this problem is to find the multiple, and the underlying principle is the transformation of thinking. Have you learned it? \#ThinkingTraining \#LearningMethods \#KnowledgeSharing & 我终于成为团宠                                           & I finally became the group's favorite.                                                                                                                                                              \\ \midrule
MML-CG              & 这题的突破口是找倍数，背后是转化思维，你学会了吗？                     & The key to this problem is to find the multiple, and the underlying principle is the transformation of thinking. Have you learned it? \#ThinkingTraining \#LearningMethods \#KnowledgeSharing & 我终于成为团宠，姜幼安教训丫鬟立威，墨扶白动心了，快看的漫画，二次元，漫画解说充能计划，漫画推荐。 & I finally became the group's favorite. Jiang You'an teaches the maid Li Wei a lesson, and Mo Fubai is touched. QuickReadComics, 2D World, Comic Explanation Empowerment Plan, Comic Recommendation. \\ \midrule
KLVCG               & 吃个梨和个桃。有5个，原来梨有几个？                            & Ate some pears and peaches. There are 5 in total, how many pears were there originally?                                                                                                       & 这个漫画真好看啊！                                         & This comic is really good!                                                                                                                                                                          \\ \midrule
baichuan2           & 这题的突破口是找倍数，背后是转化思维，你学会了吗？                     & The key to this problem is to find the multiple, and the underlying principle is the transformation of thinking. Have you learned it?                                                         & 哇，姜幼安果然是个勇敢的姑娘                                    & Wow, Jiang You'an is indeed a brave girl.                                                                                                                                                           \\ \midrule
baichuan2sft        & 老师，我儿子今年初一，数学成绩很差，怎么办？                        & Teacher, my son is in the first year of junior high this year, and his math grades are poor. What should I do?                                                                                & 姜幼安的做法真是大快人心！                                     & Jiang You'an's actions are really satisfying!                                                                                                                                                       \\ \midrule
baichuan2sft-rl     & 原来这么简单，我竟然到现在才看懂。感谢你的分享                       & Turns out it's so simple. I only just understood it. Thank you for sharing.                                                                                                                   & 哇，这个姜幼安真的好霸气，教训丫鬟的时候一点都不手软                        & Wow, this Jiang You'an is really domineering, she didn't hold back at all when teaching the maid a lesson.                                                                                          \\ \midrule
\texttt{ComHeat}             & 这个解题思路真的很清晰，以前总觉得这种题目很难，现在终于明白了！感谢分享！         & This problem-solving approach is really clear. I always thought such problems were difficult, but now I finally understand! Thanks for sharing!                                               & 真的解气！江幼安终于翻身了！大赞！                                 & So satisfying! Jiang You'an finally turned things around! Big thumbs up!                                                                                                                            \\ \midrule
Human               & 我想知道这种题在现实生活中什么场景下可以用到                        & I want to know in what real-life scenarios this kind of problem can be applied.                                                                                                               & 快打开你的视频赞赏，我送不了花花自觉点打开，让我送花花                       & Quickly open your video to receive rewards. If I can't send flowers, please open it yourself, let me send flowers.                                                                                  \\ \bottomrule
\end{tabular}}
\caption{\small Some hot-comments derived from several themes generated by the proposed \texttt{ComHeat} in comparison to other baselines of Chinese video comment generation.}
\label{tab:case-more7}
\end{table*}
\end{CJK}

\begin{CJK}{UTF8}{gkai}
\begin{table*}[htbp]
\centering
\resizebox{\textwidth}{!}{%
\begin{tabular} {p{2cm}|p{3cm}p{5cm}|p{4cm}p{4cm}}
\toprule
Video&\multicolumn{2}{c|}{\raisebox{-0.22\height}{\includegraphics[width=80pt]{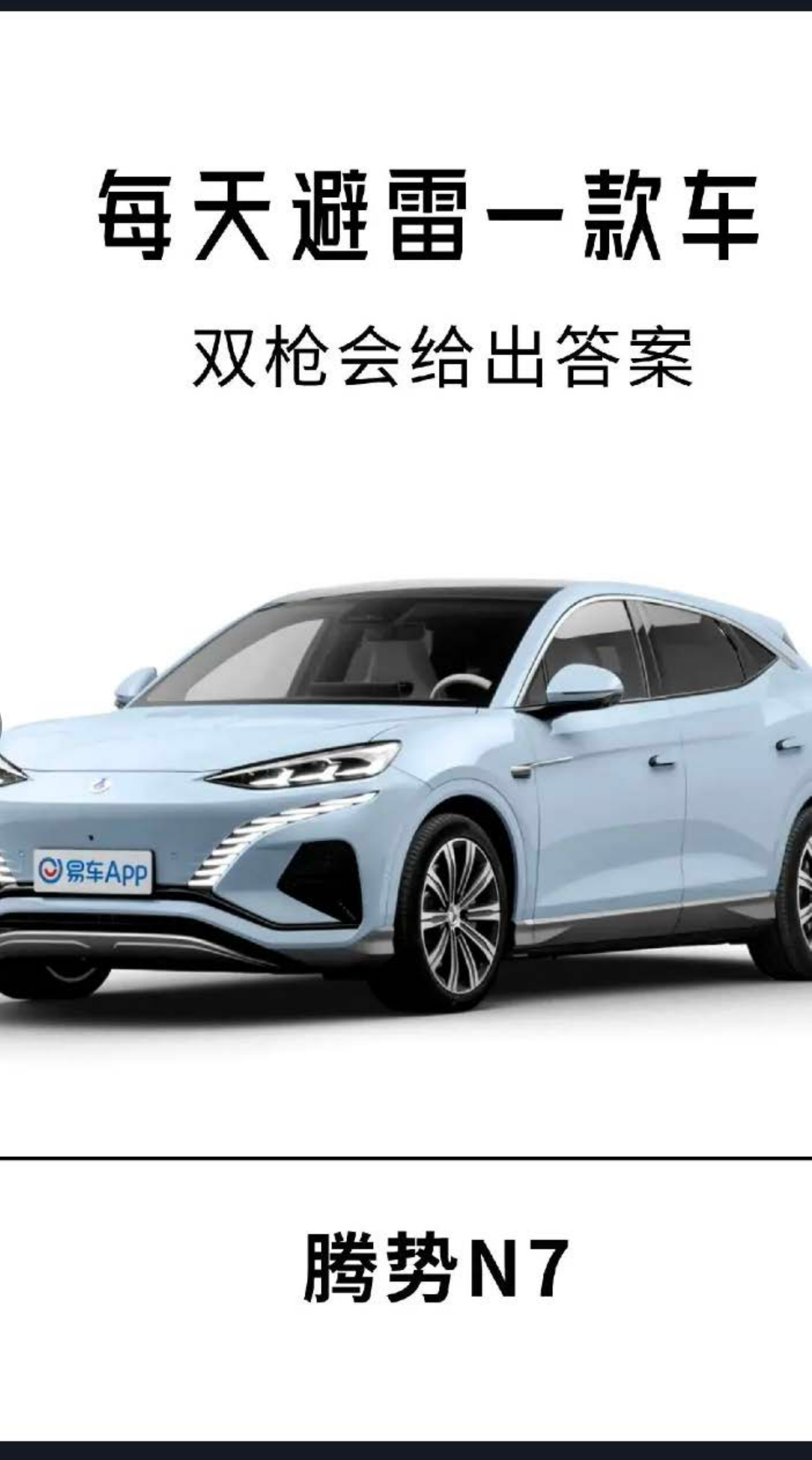}}}	&\multicolumn{2}{c}{\raisebox{-0.22\height}{\includegraphics[width=80pt]{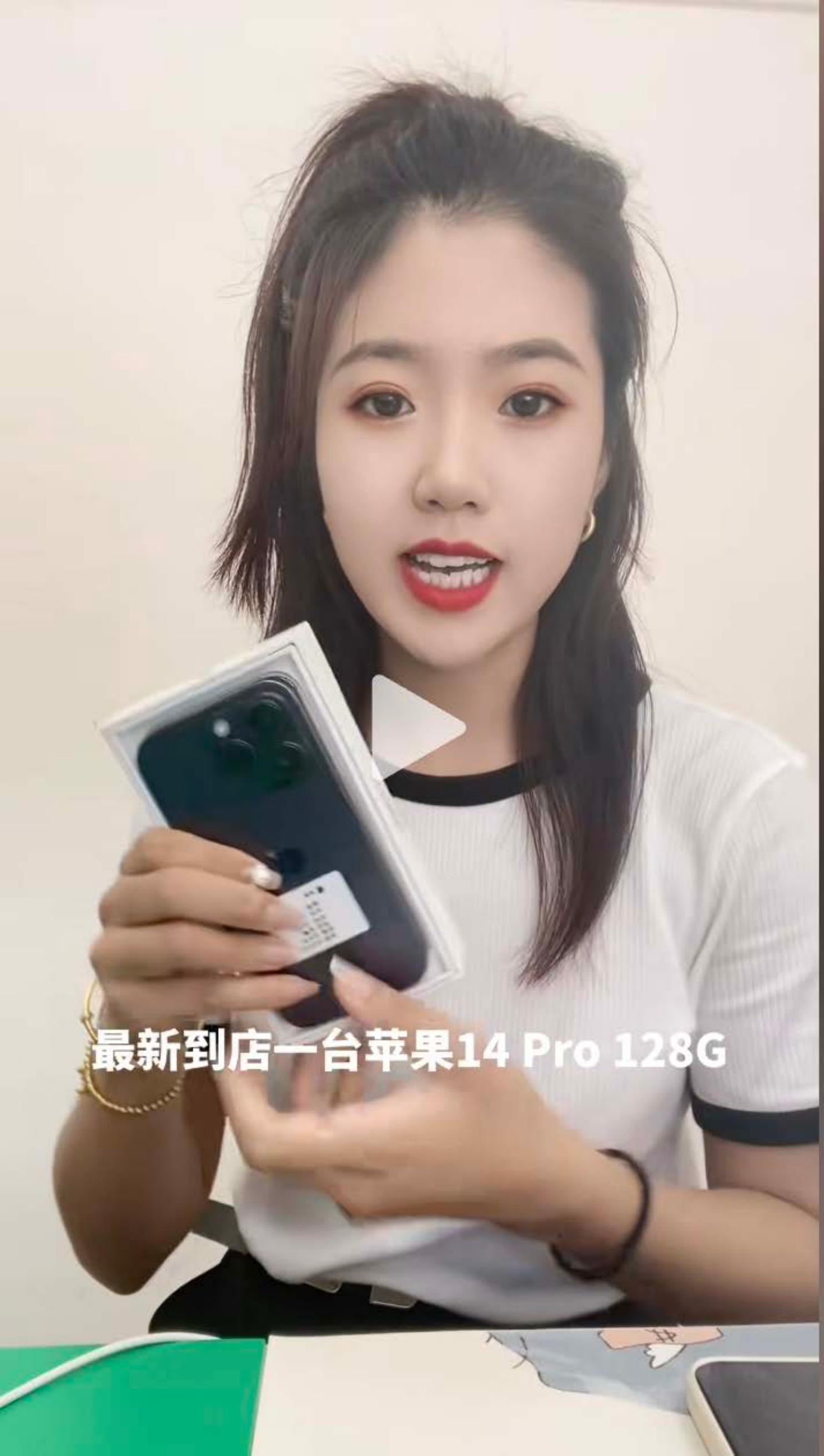}}} \\ \midrule
Theme               & 汽车                                                           & Car                                                                                                                                                                                                                       & 科技                                                        & Technology                                                                                                                                                       \\ \midrule
Title               & 腾势N7上市，价格怎么样？\#腾势 \#腾势N7上市 \#三口之家大五座SUV怎么选 \#新能源汽车 \#Lucky试驾 & Tengshi N7 is on the market, how is the price? \#Tengshi \#TengshiN7Launch \#HowToChooseA5-seaterSUVForAFamilyOfThree \#NewEnergyVehicle \#LuckyTestDrive                                                                 & \#苹果14pro \#精品二手手机 \#宝鸡同城 \#数码科技 \#手机                     & \#Apple14pro \#PremiumUsedPhone \#BaojiLocal \#DigitalTech \#MobilePhone                                                                                         \\ \midrule
Unified Transformer & 腾势N7上市，价格怎么样？                                                & Tengshi N7 is on the market, how is the price?                                                                                                                                                                            & 苹果14pro是个精品二手手机                                           & Apple 14pro is a premium used phone.                                                                                                                             \\ \midrule
MML-CG              & 腾势N7值得一试！                                                    & Tengshi N7 is worth a try!                                                                                                                                                                                                & 手机已经成为了我们生活中不可或缺的一部分，它们不仅帮助我们进行社交互动，还能满足我们的购物和工作需求。       & Mobile phones have become an indispensable part of our lives. They not only help us with social interactions but also meet our shopping and work needs.          \\ \midrule
KLVCG               & 腾势N7是新能源汽车品牌。                                                & Tengshi N7 is a new energy vehicle brand.                                                                                                                                                                                 & iPhone14pro国行双卡。256G电池100\%在保。想入手iPhone14pro的友友们你们的福利来了！。 & iPhone14pro has dual SIM cards for the Chinese version. 256G with a 100\% battery under warranty. For those who want to get iPhone14pro, here's a treat for you! \\ \midrule
baichuan2           & 30万买个比亚迪。                                                    & 300,000 yuan for a BYD.                                                                                                                                                                                                   & 灵动岛新玩法，我也要试试！                                             & New gameplay in Agile Island, I want to try it too!                                                                                                              \\ \midrule
baichuan2sft        & 腾势N7的价格有点高啊，不过它的驾驶辅助系统还是挺不错的。                                & The price of Tengshi N7 is a bit high, but its driving assistance system is pretty good.                                                                                                                                  & 灵动岛新玩法，我也要试试！                                             & New gameplay in Agile Island, I want to try it too!                                                                                                              \\ \midrule
baichuan2sft-rl     & 腾势N7的价格还不错，但是我觉得它的外观设计有点普通，不知道大家怎么看？                         & The price of Tengshi N7 is not bad, but I think its exterior design is a bit ordinary. I wonder what everyone thinks?                                                                                                     & “灵动车岛”变“灵动车壳”，这iPhone14Pro成了手机界的“变形金刚”！                   & "Agile Car Island" has turned into "Agile Car Shell", making this iPhone14Pro the "Transformers" of the mobile world!                                            \\ \midrule
\texttt{ComHeat}             & N7的外观设计确实很独特，但个人觉得前脸的车灯那么大，有点过于夸张，希望实车能看到更多细节。期待试驾体验$\sim$   & The exterior design of N7 is indeed unique, but I personally think the front lights are a bit too large and exaggerated. I hope to see more details on the actual car. Looking forward to the test drive experience$\sim$ & 14pro256G 6000多？这么多钱吗？我买的14p 256g 才5800                   & 14pro256G for over 6000? Is it that expensive? I bought my 14p 256g for only 5800.                                                                               \\ \midrule
Human               & 很少车 车灯那么大                                                    & Few cars have such large headlights.                                                                                                                                                                                      & 这样多少钱？                                                    & How much is this?                                                                                                                                                \\ \bottomrule
\end{tabular}}
\caption{\small Some hot-comments derived from several themes generated by the proposed \texttt{ComHeat} in comparison to other baselines of Chinese video comment generation.}
\label{tab:case-more8}
\end{table*}
\end{CJK}

\begin{CJK}{UTF8}{gkai}
\begin{table*}[htbp]
\centering
\resizebox{\textwidth}{!}{%
\begin{tabular} {p{2cm}|p{5cm}p{5cm}|p{4cm}p{4cm}}
\toprule
Video&\multicolumn{2}{c|}{\raisebox{-0.22\height}{\includegraphics[width=80pt]{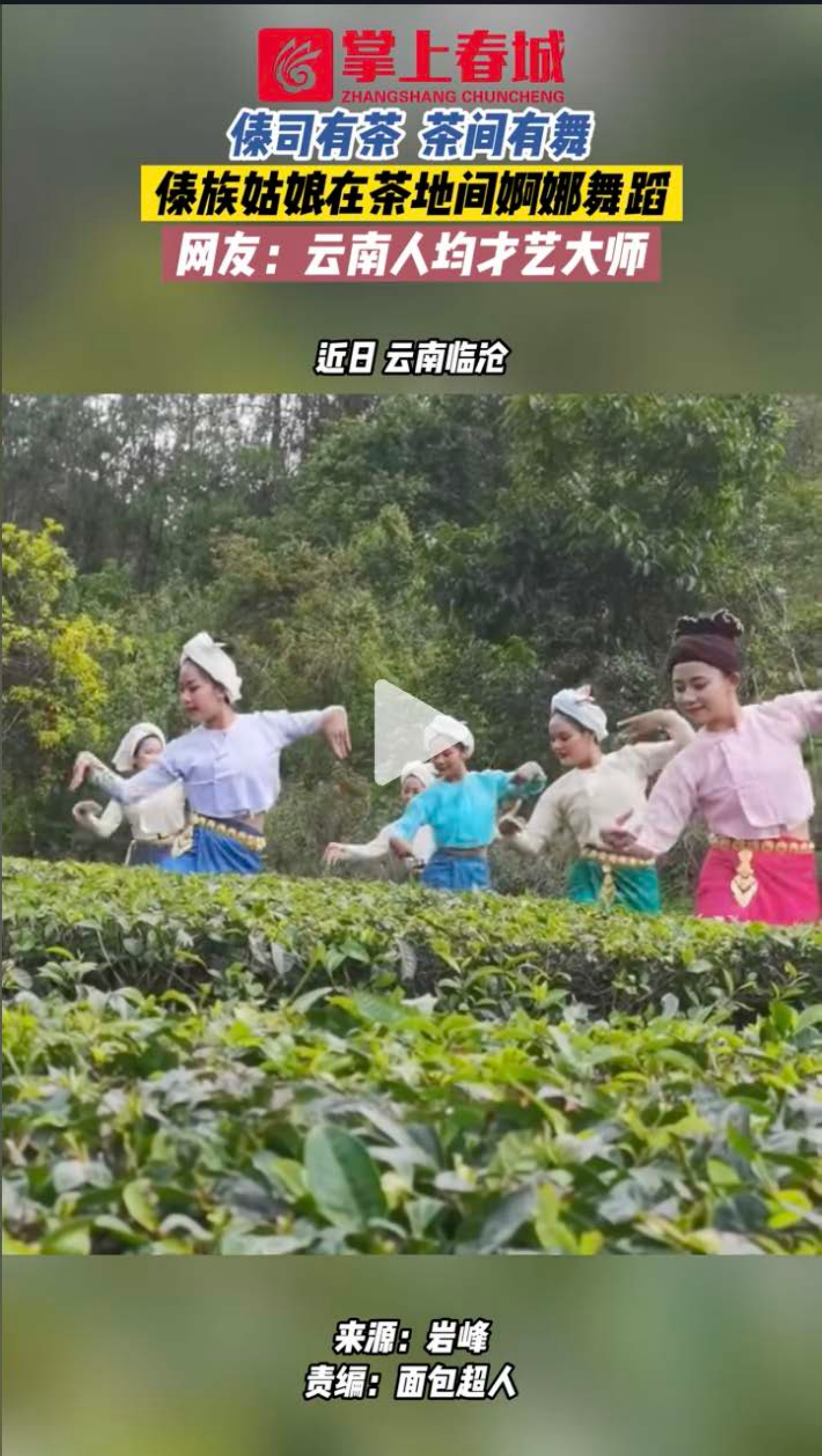}}}	&\multicolumn{2}{c}{\raisebox{-0.22\height}{\includegraphics[width=80pt]{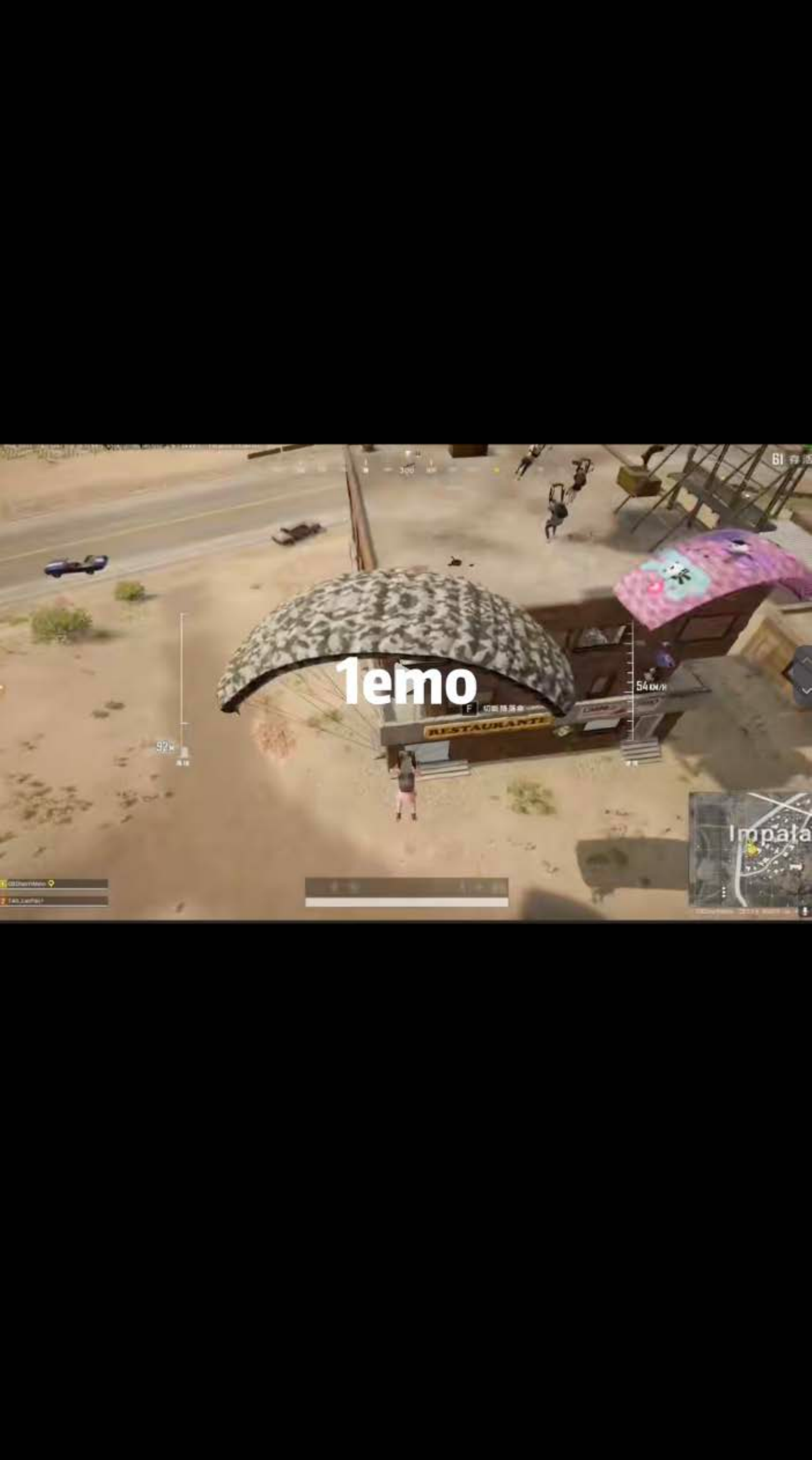}}}\\ \midrule
Theme               & 舞蹈                                                                   & Dance                                                                                                                                                                                                         & 游戏            & Game                                                                                                            \\ \midrule
Title               & 傣司有茶 茶间有舞，傣族姑娘在茶地间婀娜舞蹈\#2023年云南昆明网络文化节 \#傣族舞 \#少数民族风情 \#云南 （来源：@岩峰 ） & There's tea in Dai houses, and dance amidst the tea. Dai girls gracefully dance among the tea fields. \#2023YunnanKunmingInternetCultureFestival \#DaiDance \#EthnicMinorityStyle \#Yunnan (Source: @Yanfeng) & 生命只有一次，你要活得畅快淋漓\#吃鸡             & Life only happens once, live it to the fullest \#PUBG                                                                                                               \\ \midrule
Unified Transformer & 傣司有茶 茶间有舞，傣族姑娘在茶地间婀娜舞蹈                                               & There's tea in Dai houses, and dance amidst the tea. Dai girls gracefully dance among the tea fields.                                                                                                         & 生命只有一次，你要活得畅快淋漓！                & Life only happens once, live it to the fullest!                                                                                                                     \\ \midrule
MML-CG              & 傣族舞真是有少数民族风情啊                                                        & The Dai dance truly embodies the ethnic minority style.                                                                                                                                                       & 生命只有一次，你要活得畅快淋漓！                & Life only happens once, live it to the fullest!                                                                                                                     \\ \midrule
KLVCG               & 傣族姑娘在茶地间的婀娜舞蹈，感受云南昆明的网络文化节魅力！                                        & The graceful dance of Dai girls in the tea fields, feel the charm of the Yunnan Kunming Internet Culture Festival!                                                                                            & 吃鸡游戏 让生活畅快淋漓！                   & The PUBG game makes life exhilarating!                                                                                                                              \\ \midrule
baichuan2           & 傣族姑娘们的舞蹈太美了，仿佛让人置身于美丽的云南茶园之中，感受到了浓厚的民族风情。这个视频真是太棒了！                  & The dance of the Dai girls is so beautiful, it's as if you're in the beautiful tea gardens of Yunnan, feeling the rich ethnic style. This video is just amazing!                                              & 生命只有一次，你要活得畅快淋漓！一起加入这场刺激的游戏大战吧！ & Life only happens once, live it to the fullest! Join this thrilling game battle together!                                                                           \\ \midrule
baichuan2sft        & 点赞分享，让更多人欣赏这美丽的民族风情！                                                 & Like and share, let more people appreciate this beautiful ethnic style!                                                                                                                                       & 我玩这个游戏的时候，你们还在穿开裆裤              & When I was playing this game, you guys were still wearing split-crotch pants (a way to say "When I started this game, you were still very young or inexperienced"). \\ \midrule
baichuan2sft-rl     & 云南的少数民族风情真是太美了！这个视频让我想起了《云南印象》这首歌曲，它描绘了云南的美丽风景和少数民族的文化。              & The ethnic minority style of Yunnan is truly beautiful! This video reminds me of the song "Impression of Yunnan", which depicts the beautiful scenery of Yunnan and the culture of its ethnic minorities.     & 玩家在游戏中活出了畅快淋漓的人生，每一枪都充满了激情！     & Players live life to the fullest in the game, every shot is full of passion!                                                                                        \\ \midrule
\texttt{ComHeat}             & 我去过西双版纳，那里的姑娘都是这样美！                                                  & I've been to Xishuangbanna, and the girls there are all so beautiful!                                                                                                                                         & 生命只有一次，你要活得畅快淋漓！吃鸡这游戏真的不错！      & Life only happens once, live it to the fullest! PUBG is really a good game!                                                                                         \\ \midrule
Human               & 各位，来了都是朋友，随便坐，我去弄点吃的我们慢慢看                                            & Everyone, if you're here, you're a friend. Please have a seat. I'll go get some food and we can watch leisurely.                                                                                              & 这个游戏是什么？看起来是不错的                 & What is this game? It looks pretty good.                                                                                                                            \\ \bottomrule
\end{tabular}}
\caption{\small Some hot-comments derived from several themes generated by the proposed \texttt{ComHeat} in comparison to other baselines of Chinese video comment generation.}
\label{tab:case-more9}
\end{table*}
\end{CJK}

\begin{CJK}{UTF8}{gkai}
\begin{table*}[htbp]
\centering
\resizebox{\textwidth}{!}{%
\begin{tabular} {p{2cm}|p{4cm}p{4cm}|p{4cm}p{6cm}}
\toprule
Video&\multicolumn{2}{c|}{\raisebox{-0.22\height}{\includegraphics[width=80pt]{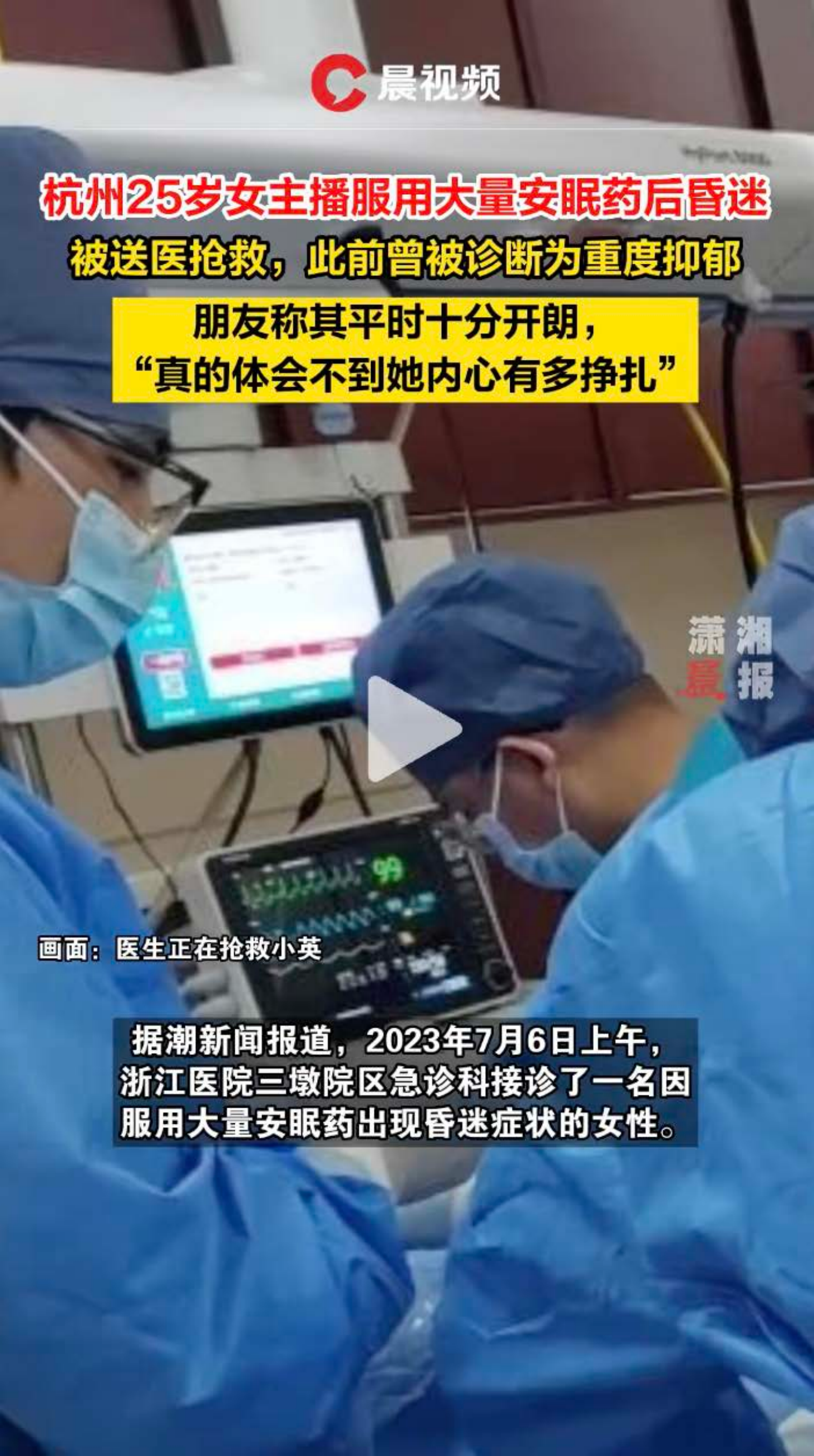}}}	&\multicolumn{2}{c}{\raisebox{-0.22\height}{\includegraphics[width=80pt]{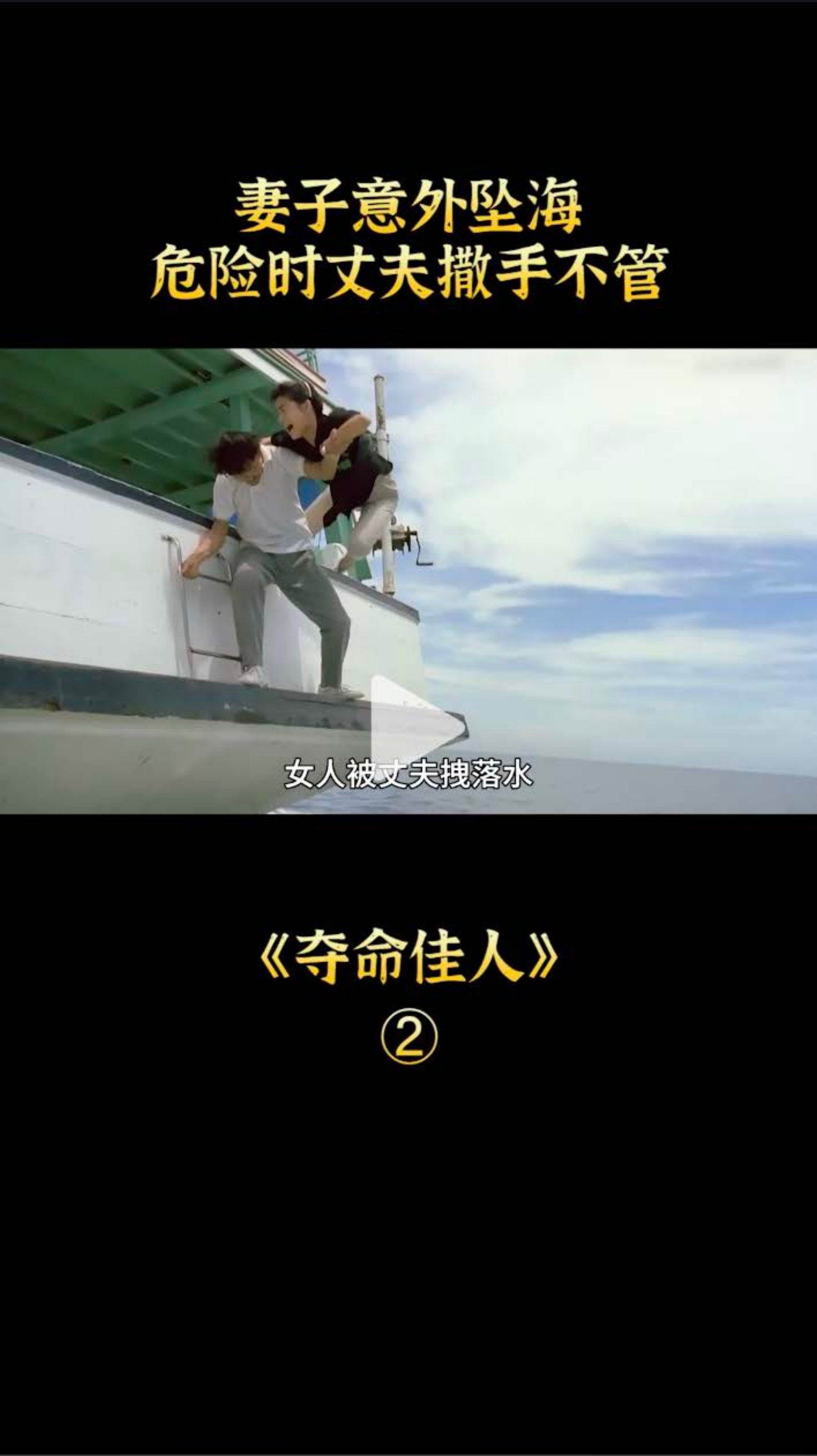}}}\\ \midrule
Theme               & 医疗健康                                                      & Medical Health                                                                                                                                                                                                                                 & 影视                                                           & Film and Television                                                                                                                                                                                                         \\ \midrule
Title               & \#杭州25岁女主播服用大量安眠药后昏迷 被诊断为微笑型抑郁症。朋友称其十分开朗，“真体会不到她内心有多挣扎”   & \#A 25-year-old female anchor in Hangzhou fell into a coma after taking a large amount of sleeping pills and was diagnosed with smiling depression. Friends say she's very cheerful, "You really can't feel how much she's struggling inside." & 妻子意外坠海 危险时丈夫撒手不管 \#人性 \#剧情 \#宅家dou剧场                         & Wife accidentally falls into the sea, husband lets go in a dangerous situation \#HumanNature \#Plot \#StayHomeDouTheater                                                                                                    \\ \midrule
Unified Transformer & 真体会不到她内心有多挣扎                                              & You really can't feel how much she's struggling inside.                                                                                                                                                                                        & 妻子的遭遇真是太可怜了                                                  & The wife's experience is really pitiful.                                                                                                                                                                                    \\ \midrule
MML-CG              & 让我们更加珍惜生活中的每一刻。为她加油                                       & Let's cherish every moment in life even more. Rooting for her.                                                                                                                                                                                 & 妻子意外坠海，危险时丈夫居然撒手不管                                           & The wife accidentally falls into the sea, and the husband shockingly lets go at the dangerous moment.                                                                                                                       \\ \midrule
KLVCG               & 希望她能早日康复！                                                 & Hope she recovers soon!                                                                                                                                                                                                                        & 这个视频真是看得人心惊胆战                                                & This video is really heart-stopping.                                                                                                                                                                                        \\ \midrule
baichuan2           & 抑郁症患者，真的需要更多的关注和关爱。                                       & Depression patients really need more attention and care.                                                                                                                                                                                       & 妻子坠海，丈夫撒手不管，这人性啊，太可怕了。                                       & The wife falls into the sea, and the husband lets go. This human nature is too horrifying.                                                                                                                                  \\ \midrule
baichuan2sft        & 抑郁症真的太可怕了，希望小英早日康复。                                       & Depression is really scary. Hope Xiao Ying recovers soon.                                                                                                                                                                                      & 这个丈夫真是冷血无情，竟然眼睁睁地看着妻子被急浪带走，事后还能装作无事一样。真是人性泯灭！                & This husband is so cold-blooded, watching his wife being carried away by the rushing waves, and afterwards can act as if nothing happened. Truly, human nature has eroded!                                                  \\ \midrule
baichuan2sft-rl     & 笑容背后的挣扎，如花般坚强。愿每个努力生活的人都能找到心灵的慰藉。                         & The struggle behind the smile is as resilient as a flower. May everyone who strives in life find solace for their souls.                                                                                                                       & 这丈夫真是冷血无情，竟然眼睁睁地看着妻子被急浪带走，事后还谎称是意外坠海。这种人不配活在世上，希望法律能严惩他。     & This husband is so cold-hearted, watching his wife being taken by the waves and later falsely claiming it was an accidental fall into the sea. Such a person doesn't deserve to live, I hope the law severely punishes him. \\ \midrule
\texttt{ComHeat}             & 我也有过这样的经历，看似光鲜亮丽，乐观开朗，但是内心的挣扎和压力真得很大。希望小英能早日康复，勇敢面对生活的挑战。 & I've also been through this. On the surface, I appear radiant and optimistic, but the inner struggle and pressure are indeed immense. I hope Xiao Ying recovers soon and bravely faces the challenges of life.                                 & 这个丈夫真是太可怕了，为了自己的利益竟然不惜谋害妻子，最后还能全身而退，真是让人气愤！我老公要是这样对我，我肯定把他杀了 & This husband is really terrifying, harming his wife for his own benefit and then getting away with it. It's infuriating! If my husband treated me like this, I'd surely kill him.                                           \\ \midrule
Human               & 我去年差点走出来，也是靠安眠药才睡得着了平时性格在别人眼中都是开郎的但一个人静下来时那种苦只有自己懂        & Last year, I almost made it through, and I could only sleep with the help of sleeping pills. Normally, others see me as cheerful, but when I'm alone, only I understand the pain I feel.                                                       & 感谢我老公，只是绿了我，没有杀了我                                            & I'm grateful to my husband, he just cheated on me, he didn't kill me.                                                                                                                                                       \\ \bottomrule
\end{tabular}}
\caption{\small Some hot-comments derived from several themes generated by the proposed \texttt{ComHeat} in comparison to other baselines of Chinese video comment generation.}
\label{tab:case-more10}
\end{table*}
\end{CJK}

\end{document}